\documentclass{article}
\usepackage{arxiv}
\usepackage{amsmath, amsthm, amsfonts, amssymb}
\usepackage[utf8]{inputenc} 
\usepackage[T1]{fontenc}    
\usepackage{hyperref}       
\usepackage{url}            
\usepackage{booktabs}       
\usepackage{amsfonts}       
\usepackage{nicefrac}       
\usepackage{microtype}      
\usepackage{cleveref}       
\usepackage{lipsum}         
\usepackage{graphicx}
\usepackage[numbers]{natbib}
\usepackage{doi}
\usepackage{bm}
\usepackage{empheq}
\usepackage{mathrsfs}
\usepackage[noend]{algpseudocode}
\usepackage{algorithmicx,algorithm}
\usepackage{graphicx}
\usepackage{wrapfig}
\usepackage{fancyhdr}
\usepackage{mathtools}
\usepackage{xcolor,color}
\usepackage{esint}
\usepackage{enumerate}
\usepackage{bbm}
\numberwithin{equation}{section}
\newtheorem{theorem}{Theorem}[section]

\newtheorem{lemma}[theorem]{Lemma}
\usepackage{mathrsfs}
\usepackage{appendix}
\usepackage{bm}
\usepackage{booktabs}
\usepackage{multirow}
\usepackage{subfigure}

\usepackage{ulem}

\DeclareMathOperator*{\argmin}{arg\,min}

\title{Adaptive deep density approximation for fractional Fokker-Planck equations}

\author{
	{Li Zeng} \thanks{ LSEC, Institute of Computational Mathematics and Scientific/Engineering
		Computing, AMSS, Chinese Academy of Sciences, Beijing, China. Email: zengli@lsec.cc.ac.cn.}\\
	\And
		{Xiaoliang Wan} \thanks {Department of Mathematics and Center for Computation and Technology, Louisiana State University, Baton Rouge 70803, USA. Email: xlwan@lsu.edu.}\\
	 \And
	 {Tao Zhou} \thanks{ LSEC, Institute of Computational Mathematics and Scientific/Engineering
	 	Computing, AMSS, Chinese Academy of Sciences, Beijing, China. Email: tzhou@lsec.cc.ac.cn.} \\
}

\begin{document}
	\maketitle
	
\begin{abstract}
	In this work, we propose adaptive deep learning approaches based on normalizing flows for solving fractional Fokker-Planck equations (FPEs). The solution of a FPE is a probability density function (PDF). Traditional mesh-based methods are ineffective because of the unbounded computation domain, a large number of dimensions and the nonlocal fractional operator. To this end, we represent the solution with an explicit PDF model induced by a flow-based deep generative model, simplified KRnet, which constructs a transport map from a simple distribution to the target distribution. We consider two methods to approximate the fractional Laplacian. One method is the Monte Carlo approximation. The other method is to construct an auxiliary model with Gaussian radial basis functions (GRBFs) to approximate the solution such that we may take advantage of the fact that the fractional Laplacian of a Gaussian is known analytically. Based on these two different ways for the approximation of the fractional Laplacian, we propose two models, MCNF and GRBFNF, to approximate stationary FPEs and MCTNF to approximate time-dependent FPEs. To further improve the accuracy, we refine the training set and the approximate solution alternately. A variety of numerical examples is presented to demonstrate the effectiveness of our adaptive deep density approaches.	

\end{abstract}


\section{Introduction}

The fractional Fokker-Planck equations (FPEs) describe the time evolution of the probability density function of particles driven by Levy noise as well as Gaussian noise. Compared to integer-order FPEs whose associated stochastic differential equations (SDEs) are only driven by Gaussian noise, the fractional FPEs have a much wider range of applications in physics, biology, and other fields \cite{shlesinger2014levy, ditlevsen1999observation, elowitz2002stochastic} since more than one kind of noise are often needed to simulate complex systems in practice. However, it is very challenging to approximate the fractional FP equations due to the following four obstacles:
\begin{itemize}
	\item [(i)]The solution is a probability density function requiring vanishing boundary, normality and non-negative conditions.
	\item [(ii)]The computational domain may be unbounded.
	\item [(iii)]The fractional Laplacian operator is nonlocal.
	\item [(iv)]The problem may have a large number of dimensions.
\end{itemize}
Traditional methods such as finite difference method, finite element method, spectral method as well as path integral method\cite{gao2016fokker, ayi2021structure, deng2009finite,  ZHANG2018302, XU201941} have been applied to approximate fractional FPEs. Most of these methods are limited to problems of dimension one or two because the mesh-based discretization of high-dimensional problems induces unaffordable computational cost. On the other hand, simulating the SDEs associated with the FPEs \cite{Zan2020} needs a large number of sample paths. Thus more efficient methods are still needed to approximate the fractional FPEs.

Recently, deep learning techniques have shown strong vitality in solving PDEs, e.g. deep Galerkin method \cite{sirignano2018dgm}, deep Ritz method \cite{weinan2018deep} and physics-informed neural networks (PINNs) \cite{raissi2019physics}. These techniques have gained encouraging performance in many applications \cite{raissi2020hidden,brunton2020machine, han2018solving,zang2020weak, yang2021b,qin2021deep,zhang2018deep,iten2020discovering,meng2020composite}. Meanwhile, many deep  generative models such as generative adversarial networks (GANs) \cite{goodfellow2014generative}, variational autoencoder (VAE) \cite{kingma2013auto} and normalizing flow (NF) \cite{papamakarios2019normalizing,rezende2015variational} have been successfully applied to learn forward and inverse SDEs \cite{chen2021solving,zhu2019physics,yang2019adversarial,liu2020neural}. For instance, a physics-informed generative adversarial model was proposed in \cite{yang2020physics} to tackle high-dimensional SDEs. In \cite{guo2021normalizing}, a normalizing field flow was developed to build surrogate models for uncertainty quantification problems. The key issue of these methods is to convert the PDE problem into an optimization problem constrained to physical laws where the loss function is discretized by random training points. The training points here refer to space-time collocation points where the equations are enforced through optimization. The choice of training points will significantly affect the final numerical accuracy especially for unbounded problems. An adaptive sampling procedure was proposed in \cite{tang2021adaptive, feng2021solving} to solve integer-order FPEs, where the training set is updated by the current approximate solution which will be subsequently improved by the new training set.  We will employ a similar adaptive procedure to deal with fractional FPEs.

To alleviate the difficulties induced by the constraints of a probability density function (PDF) we consider an explicit PDF model given by the normalizing flow.
A normalizing flow constructs an invertible mapping from a simple distribution to the target distribution and results in an explicit PDF through the change of variable. We represent the solution of the FPE via a normalizing flow.  In particular, we employ KRnet \cite{tang2020deep}, which has been successfully applied to estimate high-dimensional density function and to approximate integer-order FPEs \cite{tang2021adaptive, feng2021solving}.

Since KRnet yields a PDF explicitly, the first difficulty is avoided naturally. What's more, as a generative model, KRnet can generate exact random samples efficiently, which resolves the second obstacle because the commonly used uniform samples cannot be applied to an unbounded domain and are not effective for a large truncated domain. Using KRnet, we may update the training points by new samples from the current KRnet which automatically generates more samples in the region of high density. It is well known that automatic differentiation brings great convenience to the approximation of PDEs. However, it only works for the computation of integer-order derivatives. An effective method is needed to tackle the fractional derivatives. Several approaches have been developed to discretize the fractional derivatives when the PDE solution is modeled by neural networks. For example, a finite difference method is applied in \cite{fPINN}, and a directly Monte Carlo sampling approach was proposed in \cite{guo2022monte}. In \cite{burkardt2021unified}, Gaussian radial basis functions (GRBFs) are used to represent the solution of fractional PDEs based on the fact that the fractional Laplacian of GRBFs can be derived analytically. In this work, we will employ the Monte Carlo sampling approach or  auxiliary GRBFs to deal with the fractional Laplacian operator in the nonlocal FPEs. 

Integrating the PDF model from KRnet, automatic differentiation for integer-order derivatives and Monte Carlo sampling/GRBFs approach for fractional Laplacian, we have developed two effective deep learning techniques to address the approximation of nonlocal FPEs without requiring any labeled data. Following are the main features of our approach:
\begin{itemize}
	\item Our approach is based on the explicit PDF model given by KRnet, which satisfies naturally all the constraints of a PDF. This is different from work \cite{zhang2020statistical} which handles the constraints via adding penalty terms to the loss function.
	\item Our approach is an extension to the previous work \cite{tang2021adaptive, feng2021solving} where only FPEs with integer-order derivatives are investigated. We have paid particular attention to how to improve both the accuracy and efficiency when the fractional derivatives are involved.
	\item Being a machine learning scheme, the proposed approach is mesh-free and can be easily applied to high dimensional problems.
\end{itemize}

The remainder of this paper is structured as follows. In Section \ref{Problem_setup}, we present a brief description of the fractional FPEs. Section \ref{Stationary_FP} provides an adaptive density approximation scheme for stationary fractional FPEs. In Section \ref{Time_FP}, we generalize the approach to deal with time-dependent fractional FPEs. We demonstrate the effectiveness and efficiency of our adaptive sampling approach with several numerical experiments in Section \ref{Numerical_ex} followed by some concluding remarks in Section \ref{Conclusion}.
\section{Problem setup}\label{Problem_setup}
The main aim of this work is to solve the fractional FPEs. We first give a brief introduction to the fractional FPEs.
\subsection{Fractional Fokker-Planck equations}
Consider the state variable $\bm{X}_t$ modeled by the following stochastic differential equation
\begin{equation}
\mathrm{d}\bm {X} _{t}={\boldsymbol {\mu }}(\bm {X} _{t},t)\,\mathrm{d}t+{\boldsymbol {\sigma }}(\bm {X} _{t},t)\,\mathrm{d}\bm {W} _{t} + \mathrm{d}\bm{L}^{\alpha}_t,
\label{sde}
\end{equation}
where $\bm {X} _{t}$  and $\bm{\mu}(\bm {X} _{t},t)$ are $d$-dimensional random vectors,  $\bm{\sigma}(\bm{X}_t, t)$ is a $d\times M$ matrix, $\bm{W}_t$ is an $M$-dimensional standard Wiener process and $\bm{L}^{\alpha}_t$ is a $\alpha$-stable Levy motion with $\alpha\in(0,2)$. The probability density function (PDF) $p(\bm{x} ,t)$ for $\bm {X} _{t}$ satisfies the time-dependent FPE:
\begin{equation}
{\frac {\partial p}{\partial t}}=\mathcal{L}p-(-\Delta)^{\alpha/2}p,
\label{ffp_eq}
\end{equation}
where
\begin{equation}
\mathcal{L}p=-\nabla\cdot(p\bm{\mu})+\frac{1}{2}\nabla\cdot\nabla\cdot(\bm{\sigma}\bm{\sigma}^{\rm T}p),
\end{equation}
is induced by the drift and the diffusion,
and the following nonlocal Laplacian operator
\begin{equation}
(-\Delta)^{\alpha/2}p=C_{d,\alpha}{~\rm{P.V.}}\int_{\mathbb{R}^d\setminus\{0\}}\frac{p(\bm{x})-p(\bm{y})}{|\bm{x}-\bm{y}| _2^{d+\alpha}}{\rm d}\bm{y},
\label{nonlocal_eq}
\end{equation}
is induced by the Levy motion, where $|\cdot|_2$ indicates the $\ell_2$ norm of a vector and $\rm{P.V.}$ denotes the principle value of the integral and $C_{d,\alpha}$ is a constant given by
\begin{equation}
C_{d,\alpha}=\frac{2^{\alpha-1}\alpha\Gamma(\frac{\alpha+d}{2})}{\pi^{d/2}\Gamma(1-\alpha/2)},
\label{C_a_d}
\end{equation}
with $\Gamma(\cdot)$ being the gamma function.

In general, equation \eqref{ffp_eq} is defined on $\mathbb{R}^d$ with the following boundary condition
\begin{equation}\label{BCs}
p(\bm{x})\to 0 \quad \mbox{as} \quad | \bm{x}|_2 \to \infty.
\end{equation}
Furthermore, the solution as a probability density function should be conservative and non-negative, i.e.,
\begin{equation}\label{density_constraint}
\int_{\mathbb{R}^d}p(\bm{x},t)\mathrm{d}\bm{x}\equiv 1, \quad \mbox{and} \quad p(\bm{x},t)\geq 0.
\end{equation}
In this work, we first address the numerical approximation of equation \eqref{ffp_eq} when $\partial_tp=0$, i.e.,
\begin{equation}\label{stationart_eq}
(\mathcal{L}-(-\Delta)^{\alpha/2})p=0,
\end{equation}
and then  consider the time-dependent FPE, i.e., $\partial_tp\neq0$.

\section{MCNF and GRBFNF for stationary fractional FPE}\label{Stationary_FP}

\subsection{A bird's-eye view of proposed approaches}
As it is mentioned in Introduction, we resort to deep generative modeling to construct an explicit PDF model on $\mathbb{R}^d$ to remove all the constraints of a PDF, which also alleviates the curse of dimensionality. Depending on how to approximate the fractional Laplacian operator, we will develop two approaches to solve the fractional FPE (see Table \ref{table1}). In MCNF we approximate the fractional Laplacian by the Monte Carlo method while in GRBNF we introduce an auxiliary model to represent the approximate solution with Gaussian radial basis functions such that we may take advantage of the fact that the fractional Laplacian of a Gaussian is known explicitly. 
As for the time-dependent fractional FPEs, temporal KRnet is considered as in \cite{feng2021solving}, see Section \ref{Time_FP} for the definition of MCTNF.

\begin{table}[!htbp]
	\centering
	\begingroup
	\setlength{\tabcolsep}{10pt} 
	\renewcommand{\arraystretch}{1.5} 
	\begin{tabular}{c|c}
		\toprule
		Notations & methods\\
		\hline
		{GRBFNF} & Normalizing flow + Gaussian radial basis function\\
		\hline
		{MCNF} & Normalizing flow + Monte Carlo sampling\\
		\hline
		{MCTNF} & Temporal normalizing flow + Monte Carlo sampling\\
		\bottomrule
	\end{tabular}
	\endgroup
	\vskip 4pt
	\caption{NF indicates how to obtain a solution model. GRBF and MC indicate how to deal with the fractional Laplacian operator.}
	\label{table1}
\end{table}
\subsubsection{{\bf{MCNF}}}
Assume that the unknown PDF $p(\bm{x})$ is modeled by KRnet as $p_{\text{KRnet},\bm{\theta}}$ which will be specified in Section \ref{p_KR}. 
We adopt the idea of physics-informed neural network to deal with equation \eqref{stationart_eq}, where the overall residuals of equation \eqref{stationart_eq} on some prescribed collocation points in the computation domain will be minimized. For the given training data $S=\{\bm{x}^i\}_{i=1}^{N_S}$, we define the following loss function,
\begin{equation}
\label{Loss_MC}
\begin{aligned}
{L}(p_{\text{KRnet}, \bm{\theta}}) \coloneqq \frac{1}{N_S}\sum_{i=1}^{N_S}\vert R_{\bm{\theta} }(\bm{x}^i)\vert ^2, \quad
\end{aligned}
\end{equation}
where $R_{\bm{\theta}}(\bm{x})$ is the residual
\begin{equation}
R_{\bm{\theta}}(\bm{x})  \coloneqq (\mathcal{L}-(-\Delta)^{\alpha/2})p_{\text{KRnet},\bm{\theta}}(\bm{x}).
\end{equation}

The optimal parameters $\bm{\theta}^*$ is given by the following optimization problem
\begin{equation}
\bm{\theta}^*=\argmin_{\bm{\theta}}L(p_{\text{KRnet}, \bm{\theta}}).
\label{opti_theta}
\end{equation}

The stochastic approximation proposed in \cite{guo2022monte} is used to compute the fractional Laplacian of $p_{\text{KRnet},\bm{\theta}}$, which will be specify in Section \ref{MC_laplacian}. Another key component of our approach is the adaptive improvement of $p_{\text{KRnet},\bm{\theta}}$ (see Section 3.5), where the training set $S$ is updated by samples from the current optimal model $p_{\text{KRnet},\bm{\theta}^*}$ that will be subsequently improved by the new training set. When the convergence is reached, we expect that the samples in $S$ are distributed in terms of the exact solution $p(\bm{x})$.

\subsubsection{{\bf{GRBFNF}}}
We rewrite equation \eqref{stationart_eq} as
\begin{equation}
\left\{\begin{split}
\mathcal{L}p_{\text{KRnet},\bm{\theta}}(\bm{x})&=(-\Delta)^{\alpha/2}p_{\text{GRBF},\tilde{\bm{\theta}}}(\bm{x}),\\
p_{\text{KRnet},\bm{\theta}}(\bm{x})&=p_{\text{GRBF},\tilde{\bm{\theta}}}(\bm{x}),
\end{split}\right.
\label{loss_rbf}
\end{equation}
where $p_{\text{KRnet},\bm{\theta}}(\bm{x})$ is the same as the model used for MCNF and $ p_{\text{GRBF},\tilde{\bm{\theta}}}(\bm{x})$ is an auxiliary model for $p(\bm{x})$ (see Section \ref{p_RBF}). In other words,
\[
p(\bm{x})\approx p_{\text{KRnet}\bm{\theta}}(\bm{x}), \quad p(\bm{x})\approx p_{\text{GRBF},\tilde{\bm{\theta}}}(\bm{x}).
\]
For a set $S=\{\bm{x}^{i}\}^{N_S}_{i=1}$ of collocations points on the computation domain, we consider the following optimization problem:
\begin{equation}
(\bm{\theta}^*,\tilde{\bm{\theta}}^*)=\argmin_{\bm{\theta},\tilde{\bm{\theta}}}\tilde{L}(p_{\text{KRnet},\bm{\theta}},p_{\text{GRBF},\tilde{\bm{\theta}}}),
\end{equation}
where the tuple $(\bm{\theta}^*,\tilde{\bm{\theta}}^*)$ is the minimizer of the loss function defined as
\begin{equation}
\begin{aligned}\label{RBF_loss_eq}
\tilde{L}(p_{\text{KRnet},\bm{\theta}},p_{\text{GRBF},\tilde{\bm{\theta}}}) = &\frac{1}{
	N_S}\sum_{i=1}^{N_S}\left(\mathcal{L}p_{\text{KRnet},\bm{\theta}}({\bm{x}^{i}})-(-\Delta)^{{\alpha}/2}p_{\text{GRBF},\tilde{\bm{\theta}}}(\bm{x}^{i})\right)^2\\
&+\frac{\beta_m}{N_S}\sum_{i=1}^{N_S}\left(p_{\text{KRnet},\bm{\theta}}({\bm{x}^{i}})-p_{{\text{GRBF},\tilde{\bm{\theta}}}}(\bm{x}^{i})\right)^2,
\end{aligned}
\end{equation}
with $0 <\beta_m<\infty$ being a penalty parameter. The main difference of GRBFNF from MCNF is the introduction of the auxiliary model $p_{\text{GRBF},\tilde{\bm{\theta}}}(\bm{x})$, which will be mainly used to simplify the computation of the fractional Laplacian. More specifically, $p_{\text{GRBF},\tilde{\bm{\theta}}}(\bm{x})$ is a linear combination of the Gaussian radial basis functions with centers $\tilde{\bm{x}}_i\in S_{\text{center}}$, which corresponds to a neural network with one hidden layer. The fractional Laplacian of $p_{\text{GRBF},\tilde{\bm{\theta}}}(\bm{x})$ can be computed efficiently because the fractional Laplacian of a standard Gaussian is known analytically.  

\subsection{The density model $p_{\text{KRnet}, {\mathbf{\theta}}}$}\label{p_KR}
The constraints specified in equations \eqref{BCs} and \eqref{density_constraint} on $p(\bm{x})$ bring essential difficulties to mesh-based numerical schemes for the approximation of the fractional FPEs. To this end, we employ KRnet, a certain type of normalizing flow, to build an effective approximator for FPEs \cite{feng2021solving,tang2021adaptive}.

Normalizing flows seek an invertible mapping that corresponds to a transport map between a specified distribution and an arbitrary one.
Let $\bm{Z}\in\mathbb{R}^d$ be a simple reference random variable with a known PDF $p_{\bm{Z}},$ e.g., Gaussian. Let $f:\bm{x}\to \bm{z}$ be  an invertible mapping defined by a normalizing flow.
Then the PDF of $\bm{X}=f^{-1}(\bm{Z})$ is given by the change of variables, i.e.,
\begin{equation}
p_{\bm{X}}(\bm{x})=p_{\bm{Z}}(f(\bm{x}))\bigg|\det \nabla_{\bm{x}} f(\bm{x})\bigg|,
\label{variable_formula}
\end{equation}
where $\nabla_{\bm{x}} f(\bm{x})$ is the Jacobian matrix. Given observations of $\bm{X},$ the unknown invertible mapping can be learned via the maximum likelihood estimations.

To construct a complex bijection $f$, a general idea is to stack a sequence of simple bijections, each of which is a shallow neural network, in other words, the overall mapping is a deep neural network. Namely, the mapping $f(\cdot)$ can be written in a composite form:
\begin{equation}
\bm{z}=f(\bm{x})=f_{[L]}\circ f_{[L-1]}\circ \cdots \circ f_{[1]}(\bm{x}).
\label{com_no_time}
\end{equation}
Its inverse and Jacobian determinants are given as
\begin{eqnarray}
\bm{x} = f^{-1}(\bm{z})=f_{[1]}^{-1}\circ \cdots \circ f_{[L-1]}^{-1}\circ f^{-1}_{[L]}(\bm{z}),\\
\vert \det \nabla_{\bm{x}} f(\cdot)\vert = \prod_{i=1}^L\vert \det \nabla _{\bm{x}_{[i-1]}}f_{[i]}(\cdot)\vert,
\end{eqnarray}
where $\bm{x}_{[i-1]}$ indicates the immediate variables with $\bm{x}_{[0]}=\bm{x}, \bm{x}_{[L]}=\bm{z}.$ Many variants of $f$ have been proposed to enhance the expressive power and alleviate the computational cost of Jacobian determinants at the same time \cite{kingma2018glow, dinh2014nice, dinh2016density}. Among them, a successful example is KRnet \cite{dinh2016density}. We here employ a simplified KRnet, which includes affine coupling layers with an invertible block-triangle structure and actnorm layers. 

\subsubsection{Actnorm layer: scale and bias layer}
We adopt the Actnorm layer $L_{\text{Actn},[i]}$ with data dependent initialization proposed by  Kingma and Dhariwal \cite{kingma2018glow}:
\begin{equation}
\label{actnorm_layer}
\bm{y}_{[i]}=\bm{a}_i \odot \bm{x}_{[i]} + \bm{b}_i,
\end{equation}
where $\bm{a}_i$ and $\bm{b}_i$ are trainable parameters. When data are available, the parameters $\bm{b}_i$ and $\bm{a}_i$ can be initialized by the statistical mean and standard deviation respectively from data.  Otherwise, we may simply initialize $\bm{b}_i$ and $\bm{a}_i$ as $\bm{b}_i=\bm{0}$ and $\bm{a}_i=\bm{1}_d$, where $\bm{1}_{d}$ denotes a $d$-dimensional vector whose components are all 1. After initialization, the scale and bias are treated as regular trainable parameters that are
independent of the data.
The inverse can be easily obtained via
\begin{equation}
\bm{x}_{[i]} = (\bm{y}_{[i]}-\bm{b}_i)/\bm{a}_i,
\end{equation}
where the division here is operated on each corresponding component. 
\subsubsection{Affine coupling layer}\label{sec:a_c_l}
Let $\bm{x}_{[i]}=(\bm{x}_{[i],1}, \bm{x}_{[i],2})$ be a partition with $\bm{x}_{[i],1}\in \mathbb{R}^m$ and $\bm{x}_{[i],2}\in \mathbb{R}^{d-m}$. An affine coupling layer $L_{\text{Aff},[i]}(\cdot)$ is defined as
\begin{equation}
\label{affine_layer}
\begin{aligned}
& \bm{x}_{[i], 1}=\bm{x}_{[i-1],1},\\
& \bm{x}_{[i], 2}=\bm{x}_{[i-1],2}\odot \big(\bm{1}_{d-m}+\beta\tanh (\bm{s}_i(\bm{x}_{[i-1],1}))\big) + e^{\bm{\zeta}_i}\odot\tanh(\bm{q}_i(\bm{x}_{[i-1],1})),
\end{aligned}
\end{equation}
where $|\beta|< 1$ is a user-specified parameter (a commonly used choice is $\beta=0.6$),  $\bm{s}_i,\,\bm{q}_i:\mathbb{R}^{m}\to \mathbb{R}^{d-m}$ are scaling and translation depending only on $\bm{x}_{[i-1], 1}$, and $\bm{\zeta}_i\in \mathbb{R}^{d-m}$ is a trainable variable. Notice that the inverse can be easily computed via:
\begin{equation}
\begin{aligned}
& \bm{x}_{[i-1], 1}=\bm{x}_{[i],1},\\
& \bm{x}_{[i-1], 2}=(\bm{x}_{[i],2} - e^{\bm{\zeta}_i}\odot\tanh(\bm{q}_i(\bm{x}_{[i],1}))) \odot \big(\bm{1}_{d-m}+\beta\tanh (\bm{s}_i(\bm{x}_{[i],1}))\big)^{-1}.
\end{aligned}
\end{equation}
The Jacobian of $\bm{x}_{[i]}(\cdot)$ is given by
\begin{equation}
\nabla _{\bm{x}_{[i-1]}}\bm{x}_{[i]}(\cdot) = \left[\begin{array}{cc}
\bm{I}&\bm{0}\\
\nabla _{\bm{x}_{[i-1],1}} \bm{x}_{[i],2}& \mathrm{diag}(\bm{1}_{d-m}+\alpha\tanh (\bm{s}_i(\bm{x}_{[i-1],1})))
\end{array}\right].
\end{equation}
Furthermore, we can model $\bm{s}_{i},\bm{b}_{i}$ via neural networks
\begin{equation}
(\bm{s}_{i},\bm{q}_{i}) =\mathrm{NN}_{[i]}(\bm{x}_{[i-1],1}).
\label{NN_i}
\end{equation}
Note that $L_{\text{Aff},[i]}(\cdot)$ only changes $\bm{x}_{[i-1],2}$, implying that in the next affine coupling layer we should exchange the positions of $\bm{x}_{[i],1}$ and $\bm{x}_{[i],2}$ to ensure that each component of $\bm{x}_{[i]}$ will be updated.

Based on the actnorm layer and affine coupling layer, our simplified KRnet can be represented by
\begin{align}
&\bm{z}=f_{\text{KRnet}}(\bm{x})=f_{[L]}\circ f_{[L-1]}\circ \cdots \circ f_{[1]}(\bm{x}),\\
&f_{[i]}=L_{\text{Aff},[i]}\circ  L_{\text{Actn},[i]},\quad i=1,\dots,L,
\end{align}
where $L_{\text{Aff},[i]}$ is an affine coupling layer defined by \eqref{affine_layer} and $L_{\text{Actn},[i]}$ is an Actnorm layer defined by \eqref{actnorm_layer}.
\subsection{Stochastic approximation of the fractional operators}\label{MC_laplacian}
To compute the fractional Laplacian of the $p_{\text{KRnet},\bm{\theta}}(x)$ with $\alpha\in (0,2)$, we apply the stochastic approximation proposed in \cite{guo2022monte}.

\begin{lemma}\cite{guo2022monte}\label{lem:mc_laplacian}
	Given a function $u$,  
	its fractional Laplacian can be decomposed over a neighborhood
	$B_{r_0}(\bm{x})=\{\bm{y}\mid| \bm{y}-\bm{x}|_2\le r_0\}$ around $\bm{x}$ and its complement as
	\begin{equation}\label{eqn:Flaplacian1}
	(-\Delta)^{\alpha/2}u(\bm{x}) = C_{d,\alpha}\bigg (\int_{\bm{y}\in B_{r_{0}}(\bm{x})}\frac{u(\bm{x})-u(\bm{y})}{|\bm{x}-\bm{y}|_{2}^{d+\alpha}}\mathrm{d}\bm{y} + \int_{\bm{y}\notin B_{r_{0}}(\bm{x})}\frac{u(\bm{x})-u(\bm{y})}{|\bm{x}-\bm{y}| _{2}^{d+\alpha}}\mathrm{d}\bm{y}\bigg ).
	\end{equation}
	which, if exists, takes the form
	\begin{equation}\label{eqn:Flaplacianmc}
	\begin{aligned}
	\left(-\Delta\right)^{\alpha/2}u(\bm{x}) & =  C_{d,\alpha}\frac{\left|S^{d-1}\right|r_{0}^{2-\alpha}}{2\left(2-\alpha\right)}\mathbb{E}_{\bm{\xi}\sim \rm{U}(S^{d-1}),r_1\sim f_{\rm{I}}(r)}\left[\frac{2u(\bm{x})-u(\bm{x}-r_1\bm{\xi})-u(\bm{x}+r_1\bm{\xi})}{r_1^{2}}\right]\\
	&   +C_{d,\alpha}\frac{\left|S^{d-1}\right|r_{0}^{-\alpha}}{2\alpha}\mathbb{E}_{\bm{\eta}\sim \rm{U}(S^{d-1}),r_2\sim f_{\rm{O}}(r)}\big[2u(\bm{x})-u(\bm{x}-r_2\bm{\eta})-u(\bm{x}+r_2\bm{\eta})\big].
	\end{aligned}
	\end{equation}
	where $\bm{\xi}$ and $\bm{\eta}$ are uniformly distributed on the the unit $(d-1)$-sphere $S^{d-1}$,
	$|S^{d-1}|$ denotes the surface area of $S^{d-1}$,
	\[
	f_{\rm{I}}(r)=\frac{2-\alpha}{r_{0}^{2-\alpha}}r^{1-\alpha}\cdot1_{r\in[0,r_{0}]},\quad
	f_{\rm{O}}(r)=\alpha r_{0}^{\alpha}r^{-1-\alpha}1_{r\in[r_{0},\infty)},
	\]
	$1_{\Omega}$ is a characteristic function and $r_1$ and $r_2$ can be sampled as
	\begin{equation}\label{eqn:rI_sample}
	r_1/r_0 \sim \mathrm{Beta}(2-\alpha,1), \quad r_0/r_2 \sim \mathrm{Beta}(\alpha,1).
	\end{equation}
\end{lemma}

Notice that the first expectation in equation \eqref{eqn:Flaplacianmc} may suffer the round-off error and give rise to numerical instability for an extremely small $r$. Therefore, the following approximation is considered in practice
\begin{equation}
\label{eqn:E}
\mathbb{E}_{\bm{\xi}\sim \rm{U}(S^{d-1}),r_1\sim f_{\rm{I}}(r)}\left[\frac{2u(\bm{x})-u(\bm{x}-r_1\bm{\xi})-u(\bm{x}+r_1\bm{\xi})}{r_1^{2}}\right]
\approx\mathbb{E}_{\bm{\xi}\sim\rm{U}(S^{d-1}),r_1\sim f_{I}(r)}\left[\frac{2u(\bm{x})-u(\bm{x}-r_{\epsilon}\bm{\xi})-u(\bm{x}+r_{\epsilon}\bm{\xi})}{r_{\epsilon}^{2}}\right],
\end{equation}
with $r_{\epsilon}=\max\{\epsilon,r_1\}$, where $\epsilon>0$ is a small positive number.

Combining the stochastic approximation for the fractional Laplacian operator and the physics-informed neural network \eqref{Loss_MC}, along with the automatic differentiation for the integer-order derivative, we obtain the finial approximation for $L(p_{\text{KRnet},\bm{\theta}})$ as follows
\begin{equation}
\begin{split}
\label{num_MC_frac}
L(p_{\text{KRnet},\bm{\theta}})\approx&\hat{L}(p_{\text{KRnet},\bm{\theta}};r_{\epsilon}, r_0)\\
=&\frac{1}{N_S}\sum_{i=1}^{N_S}\left|-\nabla\cdot(\bm{\mu}p_{\text{KRnet},\bm{\theta}})(\bm{x}^i)+\frac{1}{2}\nabla\cdot\nabla\cdot(\bm{\sigma}\bm{\sigma}^{\rm T}p_{\text{KRnet},\bm{\theta}})(\bm{x}^i)\right.\\
 &-C_{d,\alpha}\frac{\left|S^{d-1}\right|r_{0}^{2-\alpha}}{2\left(2-\alpha\right)}\mathbb{E}_{\bm{\xi}\sim \rm{U}(S^{d-1}),r_1\sim f_{\rm{I}}(r)}\left[\frac{2p_{\text{KRnet},\bm{\theta}}(\bm{x}^i)-p_{\text{KRnet},\bm{\theta}}(\bm{x}^i-r_{
		\epsilon}\bm{\xi})-p_{\text{KRnet},\bm{\theta}}(\bm{x}^i+r_{\epsilon}\bm{\xi})}{r_{\epsilon}^{2}}\right]\\
  &-C_{d,\alpha}\left.\frac{\left|S^{d-1}\right|r_{0}^{-\alpha}}{2\alpha}\mathbb{E}_{\bm{\eta}\sim \rm{U}(S^{d-1}),r_2\sim f_{\rm{O}}(r)}\big[2p_{\text{KRnet},\bm{\theta}}(\bm{x}^i)-p_{\text{KRnet},\bm{\theta}}(\bm{x}^i-r_2\bm{\eta})-p_{\text{KRnet},\bm{\theta}}(\bm{x}^i+r_2\bm{\eta})\big]\right|^2.
\end{split}
\end{equation}

\begin{figure}[H]
	\centering
	\subfigure[$\alpha=1.5$]{\includegraphics[scale=0.4]{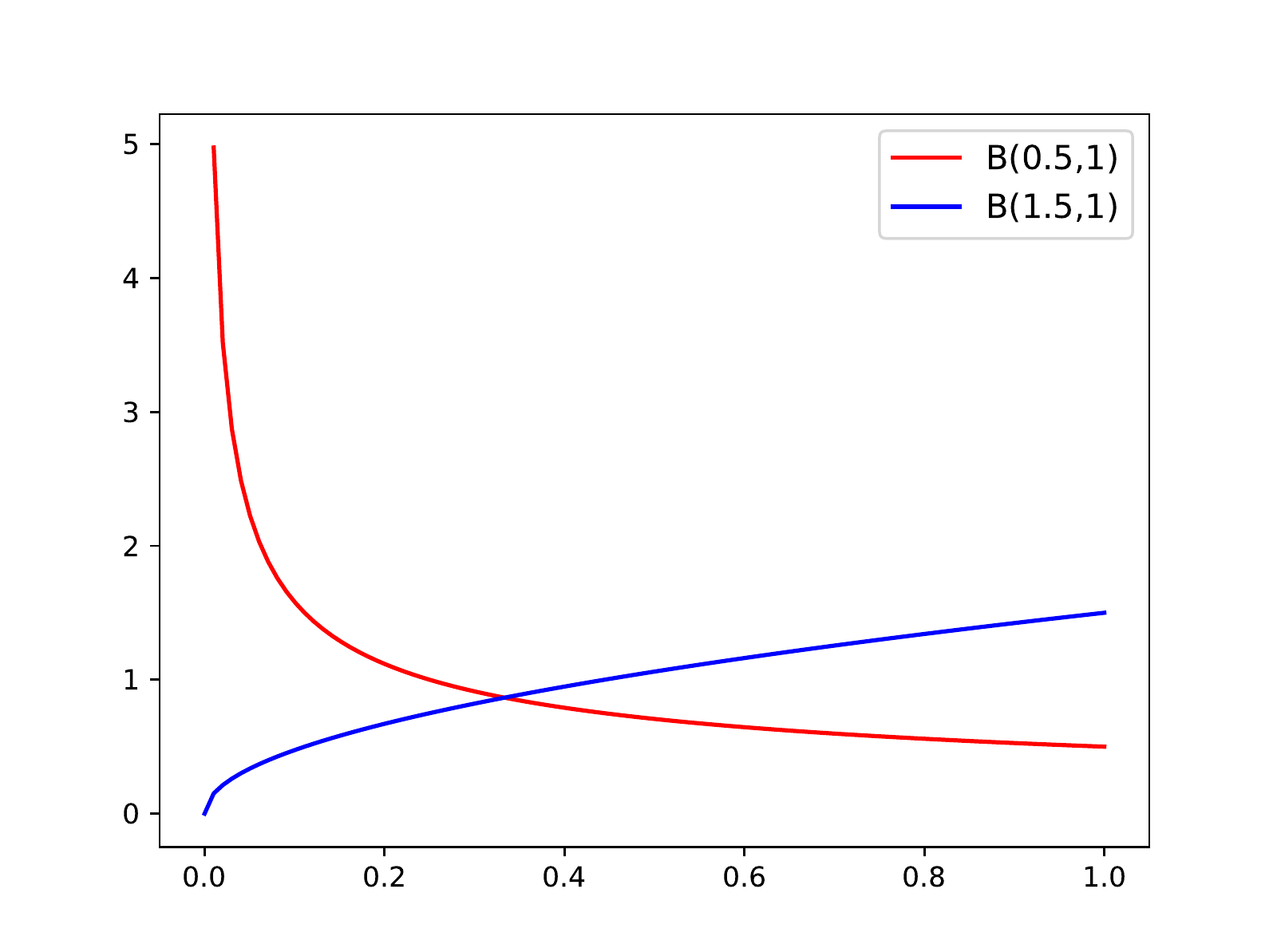}}
	\subfigure[$\alpha=1.8$]{\includegraphics[scale=0.4]{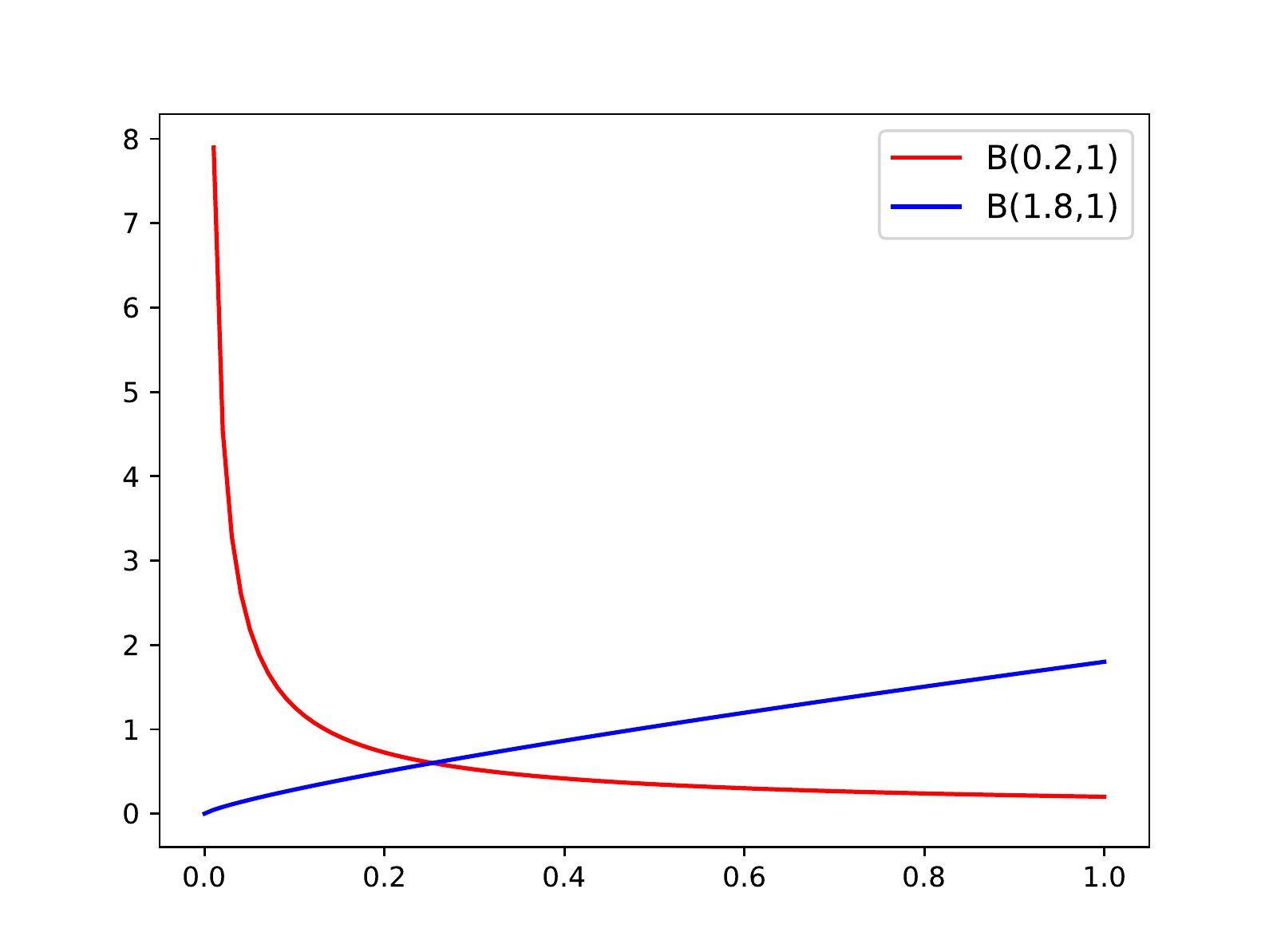}}
	\caption{Beta distribution}
	\label{Beta_fig}
\end{figure}

In Lemma \ref{lem:mc_laplacian} we need samples from Beta distributions Beta$(2-\alpha, 1)$ and Beta$(\alpha, 1)$. It is well known that Beta$(a, 1)$ becomes concentrated on origin as $a$ goes to zero, see Fig. \ref{Beta_fig}. Thus when $\alpha$ increases, the samples of $r_1$ in equation \eqref{eqn:E} may concentrate on the area close to zero, which indicates a bigger $r_{\epsilon}$ is needed to guarantee numerical stability.

\subsection{The auxiliary density model $p_{\text{GRBF},\tilde{\bm{\theta}}}$}\label{p_RBF}

The definition of the auxiliary density model $p_{\text{GRBF},\tilde{\bm{\theta}}}$ is based on the following lemma \cite{burkardt2021unified}:\\
\begin{lemma}\label{gauss_laplace_eq}
	Let $u$ be a Gaussian function of the form $u(\bm{x}) = exp(-{\sigma}^{-2}|\bm{x}-\bm{x_0}|_2^2)$ for $\bm{x,x}_0 \in \mathbb{R}^d$. Then the fractional Laplacian of $u$ is analytically given as
	\begin{equation}
	(-\Delta)^{\frac{\alpha}{2}}u(\bm{x})=c_{\alpha ,d}{|\sigma|^{-\alpha}}{_1F_1}\left(\frac{d+\alpha}{2};\frac{d}{2};-{\sigma}^{-2}|\bm{x}-\bm{x_0}|_2^2\right) \text{ for } \bm{x} \in \mathbb{R}^d,\alpha \geq 0,
	\end{equation}
	where ${_1F_1}$ denotes the confluent hypergeometric function, and
	$$	c_{\alpha,d}=\frac{2^{\alpha}\Gamma\left(\frac{d+\alpha}{2}\right)}{\Gamma\left(\frac{d}{2}\right)}.$$
\end{lemma}

For a set $S_{\text{center}}=\left\{\tilde{\bm{x}}_i\right\}^{M}_{i=1}$, we let
\begin{equation}
p_{\text{GRBF},\tilde{\bm{\theta}}}(\bm{x})=\sum_{i=1}^{M}w_i\mathcal{N}(\tilde{\bm{x}}_i,\sigma^2_i\mathbf{I})(\bm{x}),
\label{rhoRBF}
\end{equation}
where $0\leq\omega_i\leq 1$ such that $\sum_{i=1}^{M}\omega_i=1$, $\sigma_i$ is the bandwidth at $\tilde{\bm{x}}_i$, and $\mathcal{N}$ denotes the Normal distribution,
\begin{equation}
\mathcal{N}(\tilde{\bm{x}}_i, \sigma^2_i\mathbf{I})(\bm{x})=(2\pi)^{-d/2}\sigma_i^{-d}\exp\left(-\frac{|\bm{x}-\tilde{\bm{x}}_i|^2_2}{2\sigma_i^2}\right).
\end{equation}
Here both $\omega_i$ and $\sigma_i$ can be trainable parameters, which are included in $\tilde{\bm{\theta}}$. Using Lemma \ref{gauss_laplace_eq}, we obtain that
\begin{equation}
\begin{aligned}
(-\Delta)^{\alpha/2}p_{\text{GRBF},\tilde{\bm{\theta}}}(\bm{x}) &=\sum_{i=1}^{M}w_i(-\Delta)^{\alpha/2}\mathcal{N}(\tilde{\bm{x}}_i,\sigma^2_i\mathbf{I})(\bm{x})\\
&=c_{\alpha,d}\pi^{-d/2}2^{-\frac{d+\alpha}{2}}\sum_{i=1}^{M}w_i{|\sigma_i|^{-(d+\alpha)}}{_1F_1}\left(\frac{d+\alpha}{2};\frac{d}{2};-\frac{|\bm{x}-\tilde{\bm{x}}_i|_2^2}{2{\sigma_i}^{2}}\right).
\end{aligned}
\end{equation}

Consequently, the loss function \eqref{RBF_loss_eq} can be rewrite by
\begin{equation}
\label{num_RBF_frac}
\begin{split}
\tilde{L}(p_{\text{KRnet},\bm{\theta}},p_{\text{GRBF},\tilde{\bm{\theta}}}) &= \frac{1}{
	N}\sum_{i=1}^{N}\left(\mathcal{L}p_{\text{KRnet},\bm{\theta}}({\bm{x}^{i}})-c_{\alpha,d}\pi^{-\frac{d}{2}}2^{-\frac{d+\alpha}{2}}\sum_{j=1}^{M}w_j{|\sigma_j|^{-(d+\alpha)}}{_1F_1}\left(\frac{d+\alpha}{2};\frac{d}{2};-\frac{|\bm{x}^i-\tilde{\bm{x}}_j|_2^2}{2{\sigma_j}^{2}}\right)\right)^2\\
&\quad+\frac{\beta_m}{N}\sum_{i=1}^{N}\left(p_{\text{KRnet},\bm{\theta}}({\bm{x}^{i}})-\sum_{j=1}^{M}w_j(-\Delta)^{\alpha/2}\mathcal{N}(\tilde{\bm{x}}_j,\sigma^2_j\mathbf{I})(\bm{x}^{i})\right)^2,
\end{split}
\end{equation}

where the integer-order derivatives in operator $\mathcal{L}$  can be conducted via automatic differentiation.

It is seen that the factional Laplacian of $p_{\text{GRBF},\tilde{\bm{\theta}}}$ is determined by the confluent hypergeometric function $_1F_1(\cdot)$. If we allow $\sigma_i$ to be a trainable parameter, we need the derivative of $_1F_1$ which is
\begin{equation}
{\frac{\mathrm{d}}{\mathrm{d}x}}{_1F_1}\left(\frac{d+\alpha}{2};\frac{d}{2};x\right)={\frac{d+\alpha}{d}}{_1F_1}\left(\frac{d+\alpha}{2}+1;\frac{d}{2}+1;x\right).
\label{_1F_1_dx}
\end{equation}

In general it is computationally expensive to evaluate the confluent hypergeometric function. Fortunately, only the one-dimensional hypergeometric function is needed. We then use piecewise Chebyshev polynomials to approximate the one-dimensional confluent hypergeometric function up to a desired accuracy, which can be done once for all at the preprocessing stage.

\subsection{An adaptive strategy for the training process}\label{adapt_S}
\subsubsection{Where do we need adaptivity}
We pay particular attention to two components of the algorithm that are closely related to adaptivity: one is the training set $S$ and the other one is the auxiliary model $p_{\text{GRBF},\tilde{\bm{\theta}}}$. In MCNF, we only consider adaptivity for the training set $S$ while in GRBFNF we address the adaptivity for both $S$ and the model $p_{\text{GRBF},\tilde{\bm{\theta}}}$.

If the modeling capability of $p_{\text{KRnet},\bm{\theta}}$ is sufficient, the training set $S$ determines the accuracy of $p_{\text{KRnet},\bm{\theta}^*}$ because it defines the loss function for both MCNF and GRBFNF. For a fixed domain, the collocation points in $S$ are often sampled from a uniform distribution, which is obviously not optimal especially for a high-dimensional problem. Note that without any prior knowledge it is not straightforward to define a properly truncated domain to generate samples for $S$.  For $S$ with uniform samples, the loss function \eqref{Loss_MC} of MCNF can be regarded as a Monte Carlo approximation of the $L_2$ norm of the residual  in terms of a Lebesgue measure on the computation domain. The accuracy of such a Monte Carlo approximation depends on the number of samples and the variance of residual $R_{\bm{\theta}}(\bm{x})$. One way to reduce the variance is to choose collocation points in terms of another measure instead of the Lebesgue measure such that the residual $R_{\bm{\theta}}(\bm{x})$ is more uniform in terms of $\bm{x}$. For example, the loss function \eqref{Loss_MC} can be regarded as
\begin{equation}\label{Loss_MC_rho}
{L}(p_{\text{KRnet}, \bm{\theta}}) \coloneqq \frac{1}{N_S}\sum_{i=1}^{N_S}\vert R_{\bm{\theta} }(\bm{x}^i)\vert ^2\approx\int_{\mathbb{R}^d}R^2_{\bm{\theta}}(\bm{x})\rho(\bm{x})d\bm{x},
\end{equation}
where $\bm{x}^{i}$ are samples from a PDF $\rho(\bm{x})$ with $\rho(\bm{x})>0$ for any $\bm{x}\in\mathbb{R}^d$. A straightforward choice for the PDF $\rho(\bm{x})$ is the solution $p(\bm{x})$ because the residual is large more likely in the region of high probability density. If more samples are selected in the region of high density and less samples in the region of low density, the residual $R_{\bm{\theta}}(\bm{x})$ would be more evenly distributed such that the Monte Carlo approximation of the integral of $R^2_{\bm{\theta}}(\bm{x})$ in equation \eqref{Loss_MC_rho} would have a smaller statistical error. By minimizing a better approximation of the integral of $R_{\bm{\theta}}^2(\bm{x})$, a better $\bm{\theta}^*$ would be obtained.  Since $p(\bm{x})$ is unknown, we may sample its approximation $p_{\text{KRnet},\bm{\theta}^*}$ to form a new training set $S$. This suggests an adaptive solver for $p_{\text{KRnet},\bm{\theta}}$, where we update $S$ and $p_{\text{KRnet},\bm{\theta}^*}$ alternately. 

The auxiliary model $p_{\text{GRBF}, \tilde{\bm{\theta}}}$ as  an alternative representation of $p_{\text{KRnet},\bm{\theta}}$ can be regarded as a kernel density estimator (KDE) since $p_{\text{KRnet}, \bm{\theta}}$ is a PDF. Given a set of samples $\{\bm{x}_i\}$, a general adaptive
multivariate KDE takes the form \cite{terrell1992variable},
\begin{equation}
\label{KDE_eq}
\hat{p}(\bm{x})=\frac{1}{N}\sum_{i=1}^N{K_{\bm{H}_i}}(\bm{x}-\bm{x}_i),
\end{equation}
where $H_i$ is the bandwidth matrix and
$K_{\bm{H}_i}=|\bm{H}_i|^{\rm{-1}}{K(\bm{H}_i^{\rm{-1}}\bm{x})}$
rescales a kernel function $K(\bm{x})$. Due to Lemma \ref{gauss_laplace_eq}, we choose $\bm{K}(\bm{x})$ as a standard
multivariate Gaussian and $\bm{H}_i = h_i\bm{I}$ with $h_i$ being the bandwidth shared by all dimensions. An optimal bandwidth can be estimated either analytically or statistically. The main difference between $p_{\text{GRBF},\tilde{\bm{\theta}}}$ and a kernel density estimator is that the points in $S_{\text{center}}$ may not be samples from the probability density function to be approximated. This is why $p_{\text{GRBF},\tilde{\bm{\theta}}}$ in equation \eqref{rhoRBF} has variable coefficients $w_i$ while the KDE in equation \eqref{KDE_eq} has a constant coefficient $\frac{1}{N}$ . Since $p_{\text{GRBF},\tilde{\bm{\theta}}^*}\approx p_{\text{KRnet},\bm{\theta}^*}$, we expect that $S_{\text{center}}$ has a data distribution that is consistent with $p_{\text{KRnet},\bm{\theta}^*}$. When $p_{\text{KRnet},\bm{\theta}^*}$ is updated adaptively, the set $S_{\text{center}}$ should be updated accordingly for a more effective representation of $p_{\text{GRBF},\tilde{\bm{\theta}}^*}$. As $N\to\infty$, the KDE is simply the Monte Carlo simulation. However, for a GRBF approximation with a relatively small number of basis functions,  varying $w_i$ rather than the constant $\frac{1}{N}$ yield a better performance. Once a new $S_{\text{center}}$ is specified, a straightforward idea to update the parameters of GRBFs is to project $p_{\text{KRnet},\bm{\theta}^*}$ onto the new space spanned by the Gaussian radial basis functions with updated centers.

\subsubsection{Adaptivity of MCNF}
We propose the following adaptive sampling strategy to update the training set $S$. The initial collocation points in $S$ are drawn from a uniform distribution in an area determined by our prior knowledge of $p(\bm{x})$. Then we solve the optimization problem (3.24) via the Adam optimizer to obtain optimal $\bm{\theta}^*$, which corresponds to a NF mapping $f_{\bm{\theta}^{*,0}}$ and a PDF $p_{\text{KRnet},\bm{\theta}^{*,0}}(\bm{x})$. We subsequently update $S$ using samples from $p_{\text{KRnet},\bm{\theta}^{*,0}}(\bm{x})$. To be precise, we sample the latent Gaussian random variable $\bm{Z}$, and use the samples of $\bm{X}=\big({f_{\bm{\theta}^{*,0}}}\big)^{\rm{-1}}(\bm{Z})$ to form the new training set $S_1$. With $S_1$, we start a new round of training to update $p_{\text{KRnet},\bm{\theta}^{*,0}}(\bm{x})$. We repeat this procedure until the maximum iteration number is reached. Such a strategy can be concluded as follows.

\begin{enumerate}
	\item[1.] Generate an initial training set with samples uniformly distributed in $\Omega_0\subset\mathbb{R}^d$:
	$$S_0=\{\bm{x}^{i,0}\}_{i=1}^{N_S}\subset \Omega_0, \quad \bm{x}^{i,0}\sim {\mathrm{Uniform}}\;\;  \Omega_0.$$
	\item[2.] Train the KRnet by minimizing the loss function \eqref{num_MC_frac} with training data $S_0$ and hyper-parameter $r_{\epsilon}, r_0$ to obtain 
	$\bm{\theta}^{*,0}$.
	$$	\bm{\theta}^{*,0}=\arg\min_{\bm{\theta}}\hat{L}(p_{\text{KRnet},\bm{\theta}};r_{\epsilon}, r_0).$$
	\item[3.] Generate samples from $p_{\text{KRnet},\bm{\theta}^{*,0}}(\cdot)$ to get a new training set $S_1=\{\bm{x}^{i,1}\}_{i=1}^{N_S},$ and set $S_0=S_1$. Notice that $\bm{x}^{i,1}$ can be obtained by transforming the prior Gaussian samples via the inverse temporal normalizing flow,
	\begin{align*}
	\bm{z}^{i,1}\sim \mathcal{N}(\bm{0},\bm{I}), \quad \bm{x}^{i,1}=\big(f_{\bm{\theta}^{*,0}}\big)^{\rm{-1}}(\bm{z}^{i,1}).
	\end{align*}
	
	\item[4.] Repeat steps 2-3 for $N_{\mathrm{adaptive}}$ times to get a convergent approximation.
\end{enumerate}

\begin{figure}[h]
	\centering
	\includegraphics[scale=0.15]{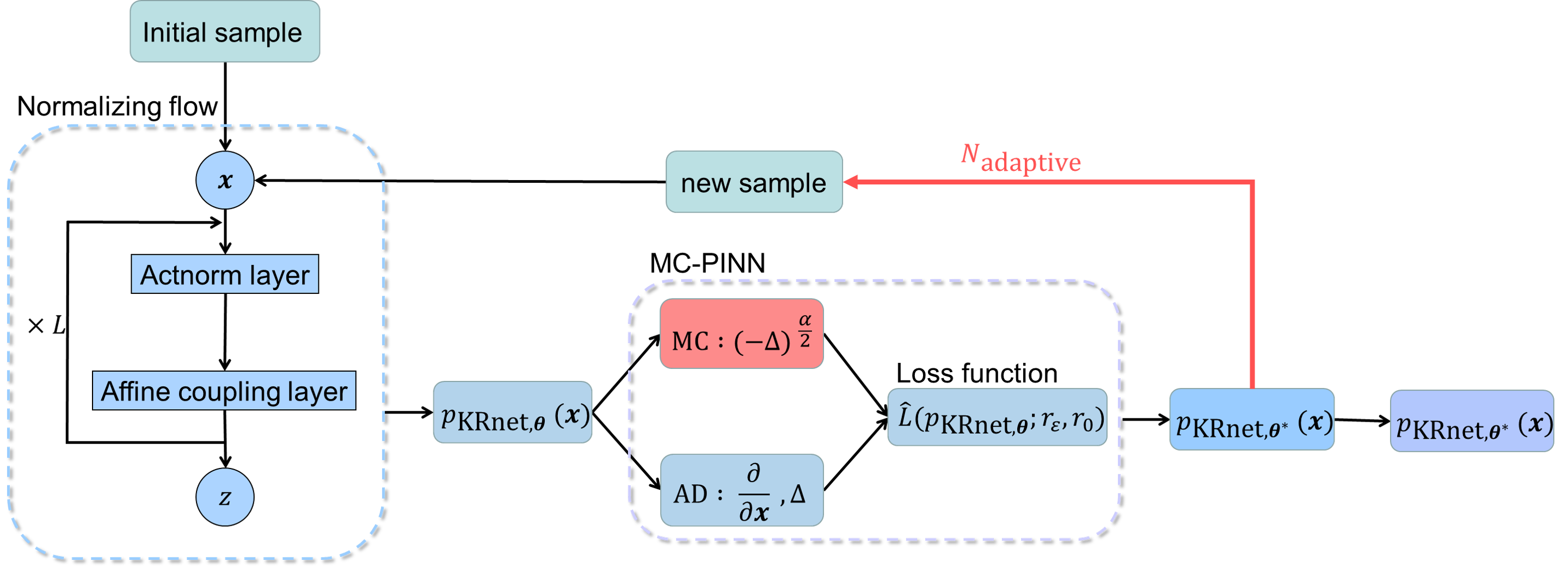}
	\caption{Flow chart of MCNF.}
	\label{flow_chart_MCNF}
\end{figure}

The algorithm for MCNF is given in Algorithm \ref{alg:1} and the flow chart is given in figure \ref{flow_chart_MCNF}. Mini batches are used to accelerate the training process. Since the initial training points are uniformly distributed, we only expect that $p_{\text{KRnet},\bm{\theta}^{*,0}}(\bm{x})$ could capture the main behavior of the exact solution $p(\bm{x})$, which implies that a relatively small number of epochs is enough. As the convergence is being established by the adaptive procedure, we expect that $p_{\text{KRnet},\bm{\theta}^{*,i}}(\bm{x})$ could capture more details of $p(\bm{x})$ as the iteration number $i$ increases, which implies that the number of epochs may increase accordingly. We introduce a hyper-parameter $\gamma$ in Algorithm \ref{alg:1} to represent the growth rate of epoch number for each adaptivity iteration.

\begin{algorithm}
	\caption{MCNF}
	\label{alg:1}
	\begin{algorithmic}
		\State \textbf{Input:} maximum epoch number $N_e$, maximum iteration number $N_{\mathrm{adaptive}},$ fractional order $\alpha$, hyper-parameter $r_{\epsilon}, r_0, \gamma$, initial training data $S=\{\bm{x}^i\}_{i=1}^{N_S},$ tolerance $\epsilon_1,\epsilon_2$;
		\State $L_{old} = 0$;
		\For{$k=1,\cdots,N_{\mathrm{adaptive}}$}
		\For{$j=1,\cdots,N_e$}
		\State Divide $S$ into $m$ batch $\{S^{ib}\}_{ib=1}^m$ randomly;
		\For{$ib = 1,\cdots,m$}
		\State Compute the loss function \eqref{num_MC_frac} ${\hat{L}^{ib}}(p_{\text{KRnet},\bm{\theta}};r_{\epsilon}, r_0)$ for mini-batch data $S^{ib}$ and order $\alpha$;
		\State Update $\bm{\theta}$ by using the Adam optimizer;
		\EndFor
		${L_{new}}=\frac{1}{m}\sum_{ib=1}^{m}\hat{L}^{ib}(p_{\text{KRnet},\bm{\theta}};r_{\epsilon}, r_0);$
		\If{${L_{new}}<\epsilon_1$ or $|{L_{old}-L_{new}}|<\epsilon_2$}
		\State \textbf{Break};
		\Else
		\State{${L_{old} = L_{new}}$};
		\EndIf
		\EndFor
		\State $N_e = \gamma * N_e$;
		\State Sample from $p_{\text{KRnet},\bm{\theta}}(\cdot)$ and update training set $S$;
		\EndFor
		\State \textbf{Output:} The predicted solution $p_{\text{KRnet},\bm{\theta}}(\bm{x})$.
	\end{algorithmic}
\end{algorithm}

\subsubsection{Adaptivity of GRBFNF}

Compared to MCNF, we need to address the adaptivity for both $S$ and the auxiliary model $p_{\text{GRBF},\tilde{\bm{\theta}}}$. 
The training set $S$ follows the same adaptive procedure as in the MCNF.  We here focus on the adaptivity for the auxiliary model. Depending on the prior knowledge, the initial center set $S_{\text{center}}$ will be formed by uniform samples in a certain area. After $p_{\text{KRnet},\bm{\theta}^{*,0}}$ is obtained, $S_{\text{center}}$ will be updated by samples from $p_{\text{KRnet},\bm{\theta}^{*,0}}$. To continue the training process with the updated $S$ and $S_{\text{center}}$, we need to reinitialize the weights $\{w_i\}$ and the bandwidths $\{\sigma_i\}$ for GRBFs of the auxiliary model, which will be done by solving a least-square problem.

\noindent\textbf{Reinitialization of the GRBFs.} Once $p_{\text{KRnet},\bm{\theta}^{*,k}}$ is obtained for the $k$-th adaptivity iteration, we sample it to update the training set from $S_{k}$ to $S_{k+1}$ and GBRF centers from $S_{\text{center},k}$ to $S_{\text{center},k+1}$. The new auxiliary model is defined as
\begin{equation}
p_{\text{GRBF},\tilde{\bm{\theta}}^{k+1}}(\bm{x})=\sum_{i=1}^Mw_iN(\tilde{\bm{x}}_i^{k+1}, \sigma^2_i\bm{I}_d)(\bm{x}), \quad \tilde{\bm{x}}_i^{k+1}\in S_{\text{center},k+1}.
\end{equation}
and initialized as
\begin{equation}\label{project_prob}
\{w_{\text{new},i}, \sigma_{\text{new},i}\}_{i=1}^M=\argmin_{w_i,\sigma_i} Loss=\argmin_{w_i,\sigma_i}\frac{1}{N_{S_{k+1}}}\sum_{j=1}^{N_{S_{k+1}}}(p_{\text{GRBF},\tilde{\bm{\theta}}^{k+1}}(\bm{x}^j)-p_{\text{KRnet},\bm{\theta}^{*,k}}(\bm{x}^j))^2,
\end{equation}
where the Adam optimizer is used to solve the above optimization problem. After initialization both the weights $\{w_i\}$ and the bandwidths $\{\sigma_i\}$ are trainable.

\begin{figure}[H]
	\centering
	\includegraphics[scale=0.145]{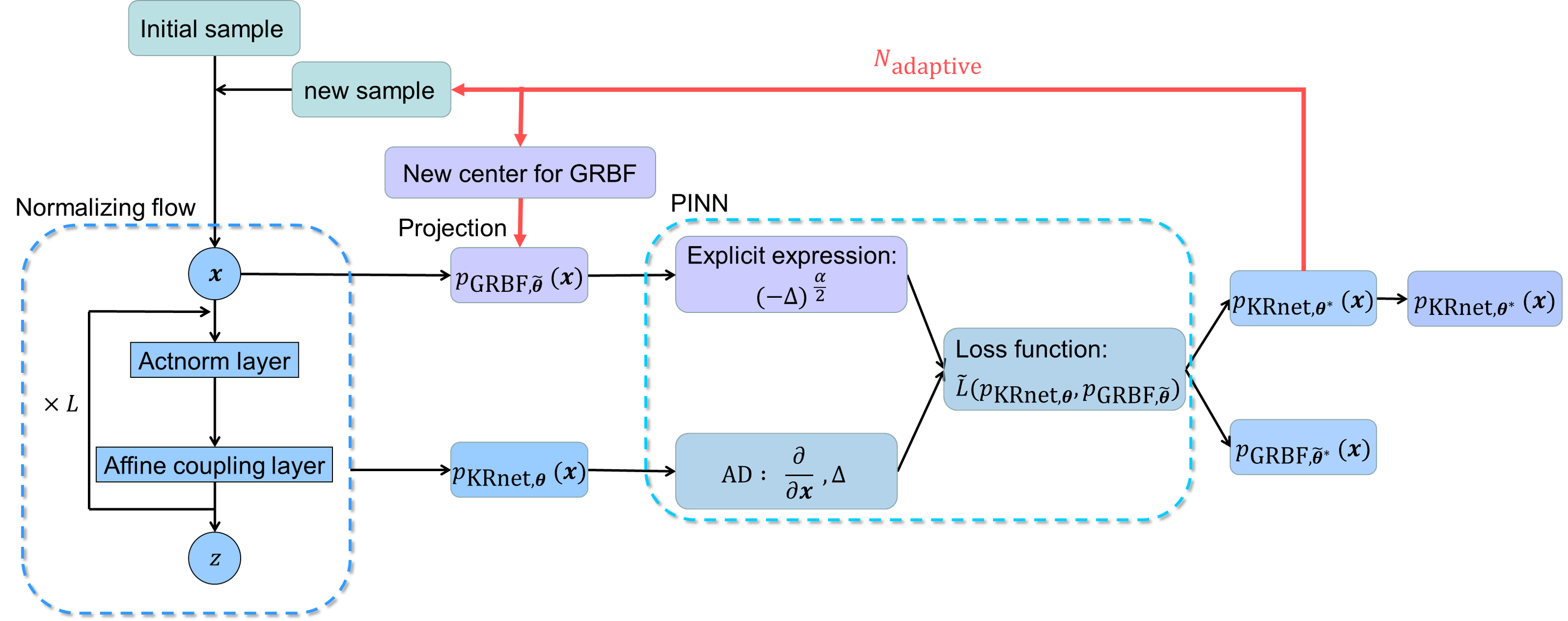}
	\caption{Flow chart of GRBFNF.}
	\label{flow_chart_RBF}
\end{figure}

Such a strategy can be concluded as follows.
\begin{enumerate}
	\item[1.] Generate initial training sets $S$ and $S_{\text{center}}$ with samples uniformly distributed in a certain physical domain:
	\begin{align*}
	S_0=\{\bm{x}^{i,0}\}_{i=1}^{N_S}\subset \Omega_0,\quad \bm{x}^{i,0}\sim {\mathrm{Uniform}}\;\;  \Omega_0,\\
	S_{\text{center},0}=\{\tilde{\bm{x}}^{0}_i\}_{i=1}^{M}\subset \Omega_0, \quad \tilde{\bm{x}}^{0}_i\sim {\mathrm{Uniform}}\;\;  \Omega_0.
	\end{align*}
	
	\item[2.] Train the KRnet by minimizing the loss function \eqref{num_MC_frac} with training data $S_0$ to obtain $\bm{\theta}^{*,0}$ and $\tilde{\bm{\theta}}^{*,0}$, i.e.,
	\begin{align*}	\{\bm{\theta}^{*,0},\tilde{\bm{\theta}}^{*,0}\}=\arg\min_{\bm{\theta},\tilde{\bm{\theta}}}\tilde{L}(p_{\text{KRnet},\bm{\theta}},p_{\text{GRBF},\tilde{\bm{\theta}}}).
	\end{align*}
	\item[3.] Generate samples with $p_{\text{KRnet},\bm{\theta}^{*,0}}$ to get a new training set $S_1=\{\bm{x}^{i,1}\}_{i=1}^{N_r},$ and center set $S_{\text{center},1}=\{\tilde{\bm{x}}^{1}_i\}$. Notice that $\bm{x}^{i,1}$ and $\tilde{\bm{x}}^{1}_j$ can be obtained by transforming the prior Gaussian samples via the inverse temporal normalizing flow.
	\begin{align*}
	\bm{z}^{i,1}\sim \mathcal{N}(\bm{0},\bm{I}), \quad \bm{x}^{i,1}=\big(f_{\bm{\theta}^{*,0}}\big)^{\rm{-1}}(\bm{z}^{i,1}),\\
	\tilde{\bm{z}}^{1}_j\sim \mathcal{N}(\bm{0},\bm{I}), \quad \tilde{\bm{x}}^{1}_j=\big(f_{\bm{\theta}^{*,0}}\big)^{\rm{-1}}(\tilde{\bm{z}}^{1}_j).
	\end{align*}
	\item[4.] Project $p_{\text{KRnet},\bm{\theta}^{*,0}}$ onto the new GRBF space by solving problem \eqref{project_prob}.  Set $S_0=S_1, S_{\text{center},0}=S_{\text{center},1}$.
	\item[5.] Repeat steps 2-3 for $N_{\mathrm{adaptive}}$ times to get a convergent approximation.
\end{enumerate}
The algorithm for GRBFNF is summarized in Algorithm \ref{alg:2} and a flow chart is given in figure \ref{flow_chart_RBF}.
\begin{algorithm}
	\caption{GRBFNF}
	\label{alg:2}
	\begin{algorithmic}
		\State \textbf{Input:} maximum epoch number $N_e$, maximum iteration number $N_{\mathrm{adaptive}},$ fractional order $\alpha$,  hyper parameter $\gamma$, initial training data $S=\{\bm{x}^{i}\}_{i=1}^{N},$ center set $S_{\text{center}}=\{\tilde{\bm{x}}_{i}\}_{i=1}^M$, tolerance $\epsilon_1,\epsilon_2$;
		\State $L_{old} = 0$;
		\For{$k=1,\cdots,N_{\mathrm{adaptive}}$}
		\For{$j=1,\cdots,N_e$}
		\State Divide $S$ into $m$ batch $\{S^{ib}\}_{ib=1}^m$ randomly;
		\For{$ib = 1,\cdots,m$}
		\State Compute the loss function \eqref{num_RBF_frac} $\tilde{L}^{ib}(p_{\text{KRnet},\bm{\theta}},p_{\text{GRBF},\tilde{\bm{\theta}}})$ for mini-batch data $S^{ib}$ and fractional order $\alpha$;
		\State Update $\bm{\theta},\tilde{\bm{\theta}}$ by using the Adam optimizer;
		\EndFor
		${L_{new}}=\frac{1}{m}\sum_{ib=1}^{m}{\tilde{L}}^{ib}(p_{\text{KRnet},\bm{\theta}},p_{\text{GRBF},\tilde{\bm{\theta}}});$
		\If{${L_{new}}<\epsilon_1$ or $|{L_{old}-L_{new}}|<\epsilon_2$}
		\State \textbf{Break};
		\Else
		\State{${L_{old} = L_{new}}$};
		\EndIf
		\EndFor
		\State $N_e = \gamma * N_e$;
		\State Sample from $p_{\text{KRnet},\bm{\theta}}(\cdot)$ and update training set $S$, $S_{\text{center}}$;
		\State Update $p_{\text{GRBF},\tilde{\bm{\theta}}}$ by solving optimization problem \eqref{project_prob}.
		\EndFor
		\State \textbf{Output:} The predicted solution $p_{\text{KRnet},\bm{\theta}}(\bm{x})$.
	\end{algorithmic}
\end{algorithm}

\section{MCTNF for time-dependent fractional FPEs}\label{Time_FP}

The procedure is overall similar to the stationary case if we can address the time-dependent problems on a space-time domain. Considering that the update of GRBF centers cannot be straightforwardly generalized to the space-time domain, we only generalize MCNF for time-dependent FPEs in this work.

Given training sets $S_t=\{(x^i,t^i)\}_{i=1}^{N_t}$ and $S_{\text{ic}}=\{(\bm{x}^i_{\text{ic}},p_0(\bm{x}_{\text{ic}}^i))\}_{i=1}^{N_{\text{ic}}}$ with $(\bm{x},t)\in\mathbb{R}^d\times[0,T]$, we define the following loss function
\begin{equation}
\label{Loss_MC_t}
\begin{aligned}
{L_T}(p_{\text{KRnet}, \bm{\theta}}(\bm{x},t)) \coloneqq \frac{1}{N_t}\sum_{i=1}^{N_t}\big\vert R_{\bm{\theta}} (\bm{x}^i,t^i)\big\vert ^2+\frac{\beta_D}{N_{\text{ic}}}\sum_{i=1}^{N_{\text{ic}}}\big|p_{\text{KRnet},\bm{\theta}}(\bm{x}^i_{\text{ic}},0)-p_0(\bm{x}^i_{\text{ic}})\big|^2, \quad
\end{aligned}
\end{equation}
where $p_0(\cdot)$ is an initial distribution, $\beta_D$ is  a weight parameter to balance the governing equation loss and the initial condition loss, and the residual ${R}_{\bm{\theta}}(\bm{x},t)$ is defined as
\begin{equation}
{R}_{\bm{\theta}}(\bm{x},t)  \coloneqq (\partial_t-\mathcal{L}+(-\Delta)^{\alpha/2})p_{\text{KRnet},\bm{\theta}}(\bm{x}, t).
\end{equation}
The optimal parameter $\bm{\theta}^*$ can be obtained via solving the following optimization problem
\begin{equation}
\bm{\theta}^*=\argmin_{\bm{\theta}}L_T(p_{\text{KRnet}, \bm{\theta}}).
\label{opti_theta_t}
\end{equation}
Note that $p_{\text{KRnet}, \bm{\theta}}(\bm{x},t)$ depends on both $\bm{x}$ and $t$, meaning that the corresponding KRnet is a time-independent normalizing flow.

\subsection{Time-dependent density model }

The time-dependent PDF $p_{\text{KRnet},\mathbf{\theta}}(\bm{x}, t)$ can be regarded as a conditional PDF $p_{\text{KRnet},\mathbf{\theta}}(\bm{x}|t)$, which can be achieved by making the affine coupling layer time dependent.
Let $\bm{x}_{[i]}=(\bm{x}_{[i],1}, \bm{x}_{[i],2})$ be a partition with $\bm{x}_{[i],1}\in \mathbb{R}^m$ and $\bm{x}_{[i],2}\in \mathbb{R}^{d-m}$. We define a time-dependent coupling layer $T_{\text{Aff},[i]}(\cdot, t)$ as follows:
\begin{equation}
\label{affine_layer_t}
\begin{aligned}
& \bm{x}_{[i], 1}=\bm{x}_{[i-1],1},\\
& \bm{x}_{[i], 2}=\bm{x}_{[i-1],2}\odot \big(\bm{1}_{d-m}+\beta\tanh (\bm{s}_{i,t}(\bm{x}_{[i-1],1},t))\big) + e^{\bm{\zeta}_i}\odot\tanh(\bm{q}_{i,t}(\bm{x}_{[i-1],1},t)),
\end{aligned}
\end{equation}
where the only difference from the affine coupling layer defined in Section \ref{sec:a_c_l} is that $\bm{s}_{i,t}$ and $\bm{q}_{i,t}$ include $t$ as their inputs such that
\begin{equation}
(\bm{s}_{i,t},\bm{q}_{i,t}) =\mathrm{NN}_{[i],t}(\bm{x}_{[i-1],1},t).
\label{NN_i_t}
\end{equation}

Based on the Actnorm layer and time-dependent affine coupling layer, our simplified time-dependent KRnet can be represented by
\begin{align}
&\bm{z}=f_{\text{KRnet},\bm{\theta}}(\bm{x},t)=f_{[L]}\circ f_{[L-1]}\circ \cdots \circ f_{[1]}(\bm{x},t),\label{eqn:x_to_z_t}\\
&f_{[i]}=T_{\text{Aff},[i]}\circ  L_{\text{Actn},[i]},\quad i=1,\dots,L,
\end{align}
where $T_{\text{Aff},[i]}$ is an time-dependent affine coupling layer defined by equation \eqref{affine_layer_t} and $L_{\text{Actn},[i]}$ is an Actnorm layer defined by equation \eqref{actnorm_layer}. For any $t$, we obtain an explicit condition PDF from equation \eqref{eqn:x_to_z_t}
\begin{equation}\label{eqn:p_krnet_t}
p_{\text{KRnet},\mathbf{\theta}}(\bm{x}, t)=p_{\text{KRnet},\mathbf{\theta}}(\bm{x}|t)=p_{\bm{Z}}(f_{\text{KRnet},\bm{\theta}}(\bm{x},t))\left|\nabla_{\bm{x}}f_{\text{KRnet},\bm{\theta}}(\bm{x},t)\right|.
\end{equation}
Also note that for any $t$, $(-\Delta)^{\alpha/2}p_{\text{KRnet},\mathbf{\theta}}(\bm{x},t)$ can be approximated using the same procedure given in Section \ref{MC_laplacian}.

One commonly used strategy to enhance the effectiveness and robustness of the algorithm is to integrate some physical constraints explicitly into the algorithm. We here propose a simple modification for the affine coupling layer such that $p_{\text{KRnet},\mathbf{\theta}}(\bm{x}, t)$ may satisfy the initial condition exactly without introducing a penalty term in the loss function.

\textbf{Modified affine coupling layer.}
	To include the initial condition, we consider a modified affine coupling layer $T_{\text{Aff}',[i]}(\cdot, t)$ as follows
	\begin{equation*}
	\label{affine_layer_t_modi}
	\begin{aligned}
	& \bm{x}_{[i], 1}=\bm{x}_{[i-1],1},\\
	& \bm{x}_{[i], 2}=\bm{x}_{[i-1],2}\odot \big(\bm{1}_{d-m}+\beta\tanh (t\bm{s}_{i,t}(\bm{x}_{[i-1],1},t))\big) + e^{\bm{\zeta}_i}\odot\tanh(t\bm{q}_{i,t}(\bm{x}_{[i-1],1},t)).
	\end{aligned}
	\end{equation*}
	where $\bm{s}_{i,t}, \bm{q}_{i,t}$ are modeled by neural network \eqref{NN_i_t} and the only modification is the scaling of $\bm{s}_{i,t}$ and $\bm{q}_{i,t}$ with time $t$. Therefore $T_{\text{Aff}',[i]}$ is an identity when $t=0$.
	 Replacing $f_{[i]}$ with $T_{\text{Aff}',[i]}$  in equations \eqref{eqn:x_to_z_t} and \eqref{eqn:p_krnet_t} we obtain at $t=0$,
	\begin{equation}
		\bm{z}=f_{\text{KRnet},\bm{\theta}}(\bm{x},0)=\bm{x}\quad\text{or}\quad p_{\text{KRnet},\bm{\theta}}(\bm{x},0)=p_{\bm{Z}}(\bm{x}).
	\end{equation}
If we choose the prior $p_{\bm{Z}}(\bm{z})$ the same as the initial distribution $p_0(\bm{z})$, the initial condition is satisfied exactly. 

\subsection{Adaptive procedure of MCTNF}

We initialize $S_{t,0}=\{(\bm{x}^{i,0},t^{i,0})\}$ using uniform samples from a space-time domain $\Omega_0\times[0,T]$, where the volume $\Omega_0$ is finite, and specify $S_{\mathrm{ ic},0}=\{(\bm{x}^{i,0}_{\mathrm{ ic}}, p_0(\bm{x}^{i,0}_{\mathrm{ ic}}))\}$. Then we solve the optimization problem \eqref{opti_theta_t} to obtain optimal $\bm{\theta}^{*, 0}$. After that we update training points $S_{t,0}$ and $S_{\mathrm{ ic},0}$ from $p_{\text{KRnet},\bm{\theta}^{*,{0}}}(\bm{x}, t)$. To be precise, we sample temporal points $\{t^{i,1}\}$ from a uniform distribution on $(0, T]$. For each $t^{i,1}$, we sample a latent normal random variable $\bm{Z}$ to obtain a sample $\bm{x}^{i,1}$ of $\bm{X}=f^{\mathrm{-1}}_{\text{KRnet},\bm{\theta}^{*, 0}}(\bm{Z}, t^{i,1})$. We then form  $S_{t,1}=\{(\bm{x}^{i,1},t^{i,1})\}$. $S_{\mathrm{ ic},1}=\{\bm{x}^{i,1}_{\mathrm{ ic}}\}$ can be obtained via the same procedure by letting $t=0$, i.e. $\bm{x}^{i,1}_{\mathrm{ ic}}=f^{-1}_{\text{KRnet},\bm{\theta}^{*, 0}}(\bm{z}^i, 0)$. We then continue the training process with $S_{t,1}$ and $S_{\text{ic},1}$. The procedure is repeated after the second training is done. Such a strategy can be concluded as follows.
\begin{enumerate}
	\item[1.] Generate initial training sets using uniform samples on $\Omega_0\times(0,T]$ where $\Omega_0\in\mathbb{R}^d$ and $|\Omega_0|<\infty$:
	\begin{align*}
	&S_{t,0}=\{(\bm{x}^{i,0},t^{i,0})\}_{i=1}^{N_t}, \quad t^{i,0}\sim {\mathrm{Uniform}}(0, T], \quad \bm{x}^{i,0}\sim {\mathrm{Uniform}}\;\;  \Omega_0,\\
	&S_{\mathrm{ ic},0}=\{(\bm{x}_{\mathrm{ ic}}^{i,0}, p_0(\bm{x}_{\mathrm{ ic}}^{i,0}))\}, \quad \bm{x}_{\mathrm{ ic}}^{i,0}\sim \mathrm{Uniform} \;\; \Omega_{0}.
	\end{align*}
	\item[2.] Train the temporal KRnet by solving optimization problem \eqref{opti_theta_t} with training data $S_{t,0}, S_{\mathrm{ ic},0}$ to obtain optimal parameters $\bm{\theta}^{*,0}$:
	$$	\bm{\theta}^{*,0}=\argmin_{\bm{\theta}}{{L}}_{T}(p_{\text{KRnet},\bm{\theta}}(\bm{x},t)).$$
	
	\item[3.] Generate temporal samples from a uniform distribution on $(0,T]$ and spatial samples from $p_{\text{KRnet},\bm{\theta}^{*,0}}(\bm{x}|t)$ to obtain $S_{t,1}, S_{\mathrm{ ic},1}$.
	\begin{align*}
	&S_{t,1}=\{(\bm{x}^{i,1},t^{i,1})\}_{i=1}^{N_t}, \quad t^{i,1}\sim {\mathrm{ Uniform}}(0, T], \quad \bm{x}^{i,1}\sim p_{\text{KRnet},\bm{\theta}^{*,0}}(\bm{x}|t=t^{i,1}),\\
	&S_{\mathrm{ ic},1}=\{(\bm{x}_{\mathrm{ ic}}^{i,1}, p_0(\bm{x}_{\mathrm{ ic}}^{i,1}))\}, \quad \bm{x}_{\mathrm{ ic}}^{i,1}\sim p_{\text{KRnet},\bm{\theta}^{*,0}}(\bm{x}|t=0).
	\end{align*}
	Set $S_{t,0}=S_{t,1}$, $S_{\mathrm{ ic},0}=S_{\mathrm{ ic},1}$.
	\item[4.] Repeat steps 2-3 for $N_{\mathrm{adaptive}}$ times to get a convergent approximation.
\end{enumerate}
Our algorithm for solving time-dependent fractional FPEs is given in Algorithm \ref{alg:3}.

\begin{algorithm}[H]
	\caption{MCTNF}
	\label{alg:3}
	\begin{algorithmic}
		\State \textbf{Input:} maximum epoch number $N_e$, maximum iteration number $N_{\mathrm{adaptive}},$ fractional order $\alpha,$ hyper-parameter $ r_{\epsilon}, r_0, \beta_D$, initial training data $S_{t}=\{(\bm{x}^i,t^i)\}_{i=1}^{N_t}, S_{\mathrm{ic}}=\{(\bm{x}_{\mathrm{ic}}^i, p_0(\bm{x}_{\mathrm{ic}}^i))\}_{i=1}^{N_{\mathrm{ic}}}$, $C_T =\{t_r^i\}_{i=1}^{N_r}\cup \{0\}_{i=1}^{N_{\mathrm{ic}}}$, tolerance $\epsilon_1,\epsilon_2$;
		\State $L_{old} = 0$;
		\For{$k=1,\cdots,N_{\mathrm{adaptive}}$}
		\For{$j=1,\cdots,N_e$}
		\State Divide $S_t, S_{\mathrm{ic}}$ into $m$ batches $\{S^{\text{ib}}_t\}_{\text{ib}=1}^m, \{S^{\text{ib}}_{\mathrm{ ic}}\}_{\text{ib}=1}^m$ randomly;
		\For{$ib = 1,\cdots,m$}
		\State Compute the loss function ${L_T}(p_{\text{KRnet}, \bm{\theta}})$
		for mini-batch data $S_t^{\text{ib}}, S_{\text{ic}}^{\text{ic}}$ and fractional order $\alpha$;
		\State Update $\bm{\theta}_t$ by using the Adam optimizer;
		\EndFor
		${L_{new}}=\frac{1}{m}\sum\limits_{ib=1}^{m}{L_T}(p_{\text{KRnet}, \bm{\theta}})$;
		\If{${L_{new}}<\epsilon_1$ or $|{L_{old}-L_{new}}|<\epsilon_2$}
		\State \textbf{Break};
		\Else
		\State{${L_{old} = L_{new}}$};
		\EndIf
		\EndFor
		\State $N_e = \gamma * N_e$;
		\State Sample from $t\sim\text{Uniform}([0,T])$ and $p_{\text{KRnet}, \bm{\theta}_t}(\bm{x}|t)$ to update training sets $S_t, S_{\mathrm{ ic}}$;
		\EndFor
		\State \textbf{Output:} The predicted solution $p_{\text{KRnet},\bm{\theta}}(\bm{x},t)$.
	\end{algorithmic}
\end{algorithm}

\section{Numerical experiments}\label{Numerical_ex}
In this section, we present a series of comprehensive numerical tests to demonstrate the effectiveness of the proposed algorithms. To quantitatively evaluate the accuracy of the numerical solution $p_{\text{KRnet},\bm{\theta}}$, we shall consider both the relative $L_2$ error $\Vert p^* - p_{\text{KRnet},\bm{\theta}}\Vert_2 / \Vert p^*\Vert_2$ and the relative Kullback-Leibler (KL) divergence given by $$\frac{D_{\mathrm{KL}}(p^*||p_{\text{KRnet},\bm{\theta}})}{H(p^*)} = \frac{\mathbb{E}_{p^*}[\log(p^*/p_{\text{KRnet},\bm{\theta}})]}{-\mathbb{E}_{p^*}[\log p^*]},$$ where $\mathbb{E}$ denotes the expectation and $p^*$ the ground truth.
We approximate the above relative $L_2$ error by Monte Carlo integration, namely,
$$\frac{\big\Vert p^* - p_{\text{KRnet},\bm{\theta}}\big\Vert_2}{ \big\Vert p^*\big\Vert_2}=\frac{\left(\int (p^*(\bm{x})-p_{\text{KRnet},\bm{\theta}}(\bm{x}))^2\mathrm{d}\bm{x}\right)^{1/2}}{\left(\int (p^*(\bm{x}))^2\mathrm{d}\bm{x}\right)^{1/2}}\approx \frac{\left(\sum_{i=1}^N(p^*(\bm{x}_i)-p_{\text{KRnet},\bm{\theta}}(\bm{x}_i))^2\triangle \bm{x}_i\right)^{1/2}}{\left(\sum_{i=1}^Np^*(\bm{x}_i)^2\triangle \bm{x}_i\right)^{1/2}}.$$
Similarly, we also approximate the above relative KL divergence by Monte Carlo integration, i.e.,
$$\frac{D_{\mathrm{KL}}(p^*||p_{\text{KRnet},\bm{\theta}})}{H(p^*)}\approx \frac{\sum_{i=1}^{N_v}\big(\log(p^*(\bm{x}_i)-\log p_{\text{KRnet},\bm{\theta}}(\bm{x}_i))\big)}{-\sum_{i=1}^{N_v}\log p^*(\bm{x}_i)}.$$
Here $\bm{x}_i$ are drawn from the ground truth $p^*(\bm{x})$ and the amount of validation data is set to $N_v=10^6$. An uniform mesh is used to calculate {the} relative $L_2$ error with mesh size 0.04 along each spatial dimension. For time-dependent problems, we obtain the relative $L_2$ error and the relative KL divergence according to the aforementioned formulas for each given $t$.

We shall employ hyperbolic tangent  function (tanh) as the activation function. For each $i$, $\mathrm{NN}_{[i]}$ (see \ref{NN_i}) is a feed forward neural network with two hidden layers. We use a half-half partition $\bm{x}_{[i]}=(\bm{x}_{[i], 1}, \bm{x}_{[i],2})$, $\bm{x}_{[i], 1}\in\mathbb{R}^{\lfloor d/2\rfloor}$, $\bm{x}_{[i],2}\in\mathbb{R}^{d-\lfloor d/2 \rfloor}$ unless specified. We initialize all trainable parameters using Glorot initialization \cite{glorot2010understanding}. For the training procedure, we use the Adam optimizer \cite{kingma2014adam}. All numerical tests are implemented with Pytorch.

\subsection{FPE with only fractional Laplacian}
We start with a toy example with only the fractional Laplacian term. Consider the following 2D equation
\begin{equation}
\left\{
\begin{aligned}
&(-\Delta)^{\alpha/2}p(\bm{x})=f(\bm{x}), \quad \bm{x}\in\mathbb{R}^2,\\
&\int_{\mathbb{R}^2}p(\bm{x}){\rm d}\bm{x}=1, \quad p(\bm{x})\geq0,
\end{aligned}\right.
\end{equation}
where $f(\bm{x})=-\frac{1}{2\pi}B(2,\alpha)2^{-\frac{\alpha}{2}}\sigma^{-(2+\alpha)}{_1F_1}(\frac{2+\alpha}{2}; 1; -\frac{\left\|x-\mu\right\|^2_2}{2\sigma^2})$, $B(d, \alpha)=\frac{2^{\alpha}\Gamma((\alpha+d)/2)}{\Gamma(d/2)}.$ The true solution is $p(\bm{x})=\frac{1}{2\pi\sigma^2}\exp(-\frac{\left\|\bm{x}-\bm{\mu}\right\|^2_2}{2\sigma^2})$.
We take $\alpha=1, \bm{\mu}=(1,1), \sigma=2$.

For the NF, we take $8$ affine coupling layers with $32$ hidden neurons. The initial training set is generated via the uniform distributed points in $[0, 6]^2$. Note that $\mathbb{E}_p[1_{[0,6]^2}]\approx 0.5$, meaning that we have only used about $50\%$ information about the effective domain of the target $\bm{X}$, where $\mathbb{E}_p$ indicates the expectation with respect to $p(\bm{x})$ and $1_{\Omega}$ is an indicator function for $\Omega\subset\mathbb{R}^2$. The sample size is $5000$ and the batch size is $1024$. Both {MCNF} and {GRBFNF} are applied. For the {{MCNF}}, the number of Monte Carlo samples used to approximate fractional Laplacian is $100$, $r_0=4$, $r_{\epsilon}=0.01$.
The initial learning rate is $0.001$ with half decay each $100$ steps. For the {{GRBFNF}}, the number of basis functions is $100$ and the initial center points of basis function are generated from a uniform distribution on $[0, 6]^2$.
The learning rate is $0.01$ with half decay each $300$ steps.
\begin{figure}[H]
	\centering
	\subfigure[Different adaptive frequencies for MCNF.]{
		\includegraphics[height=3.5cm,width=5.5cm]{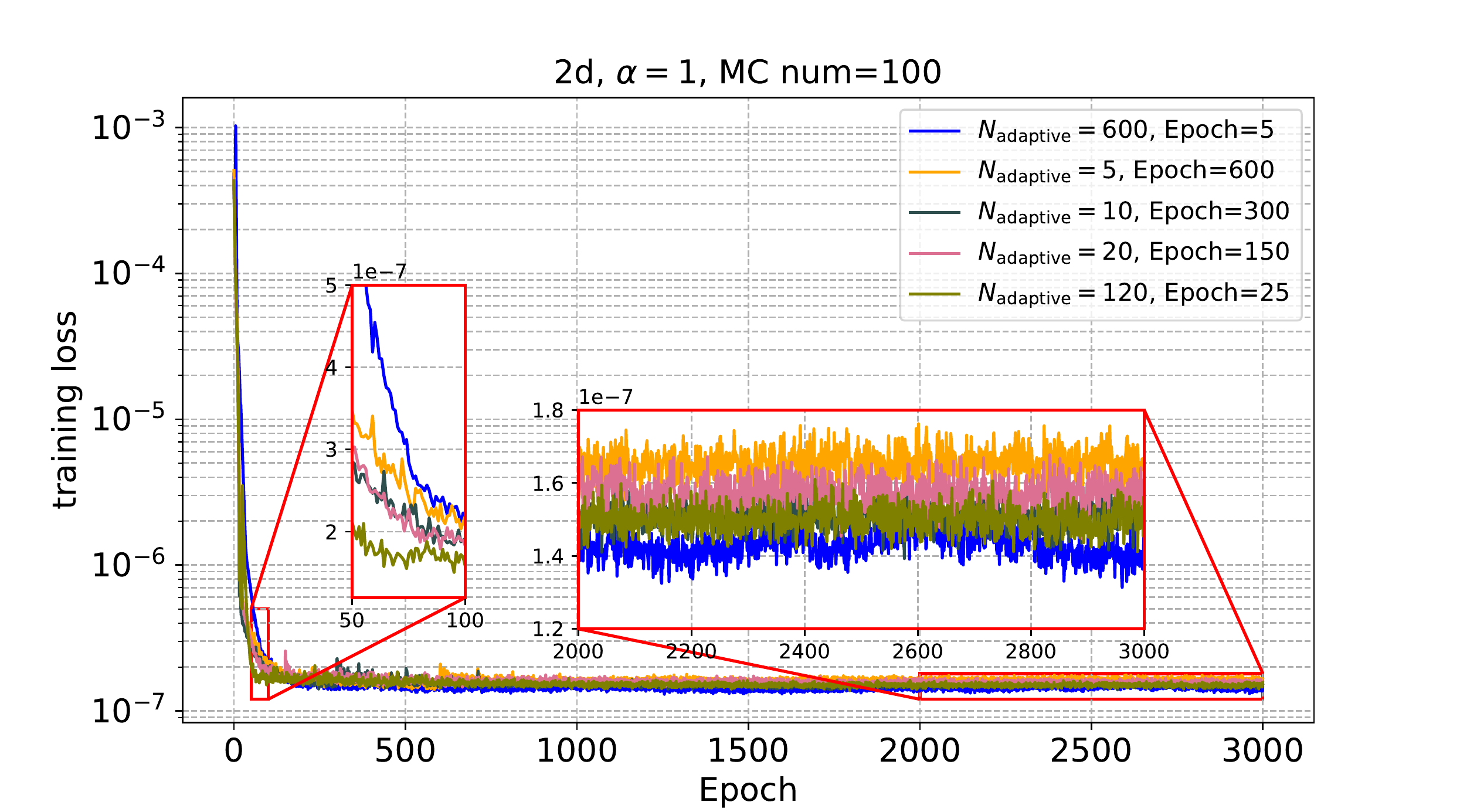}
		\includegraphics[height=3.5cm,width=5.5cm]{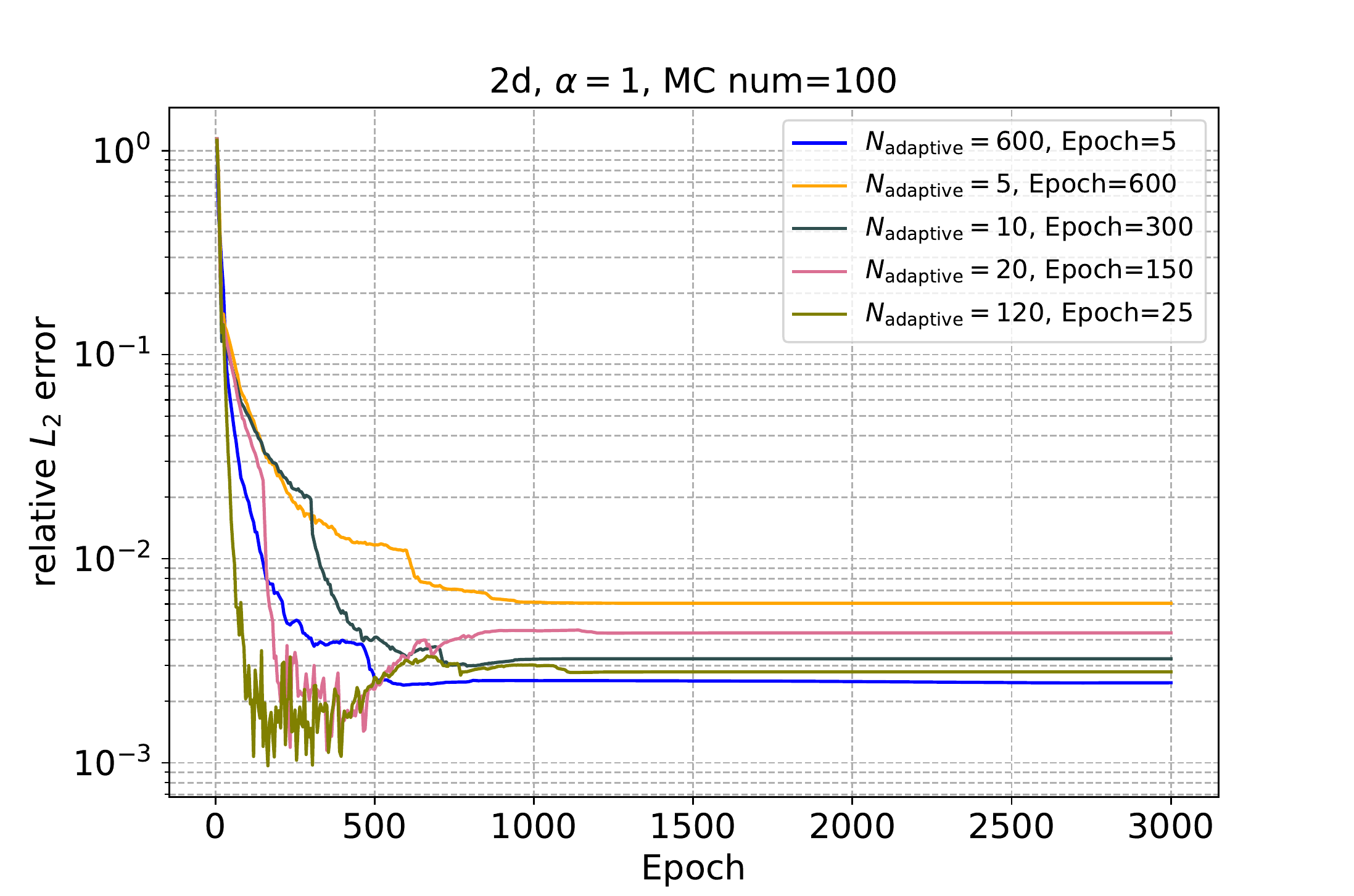}
		\includegraphics[height=3.5cm,width=5.5cm]{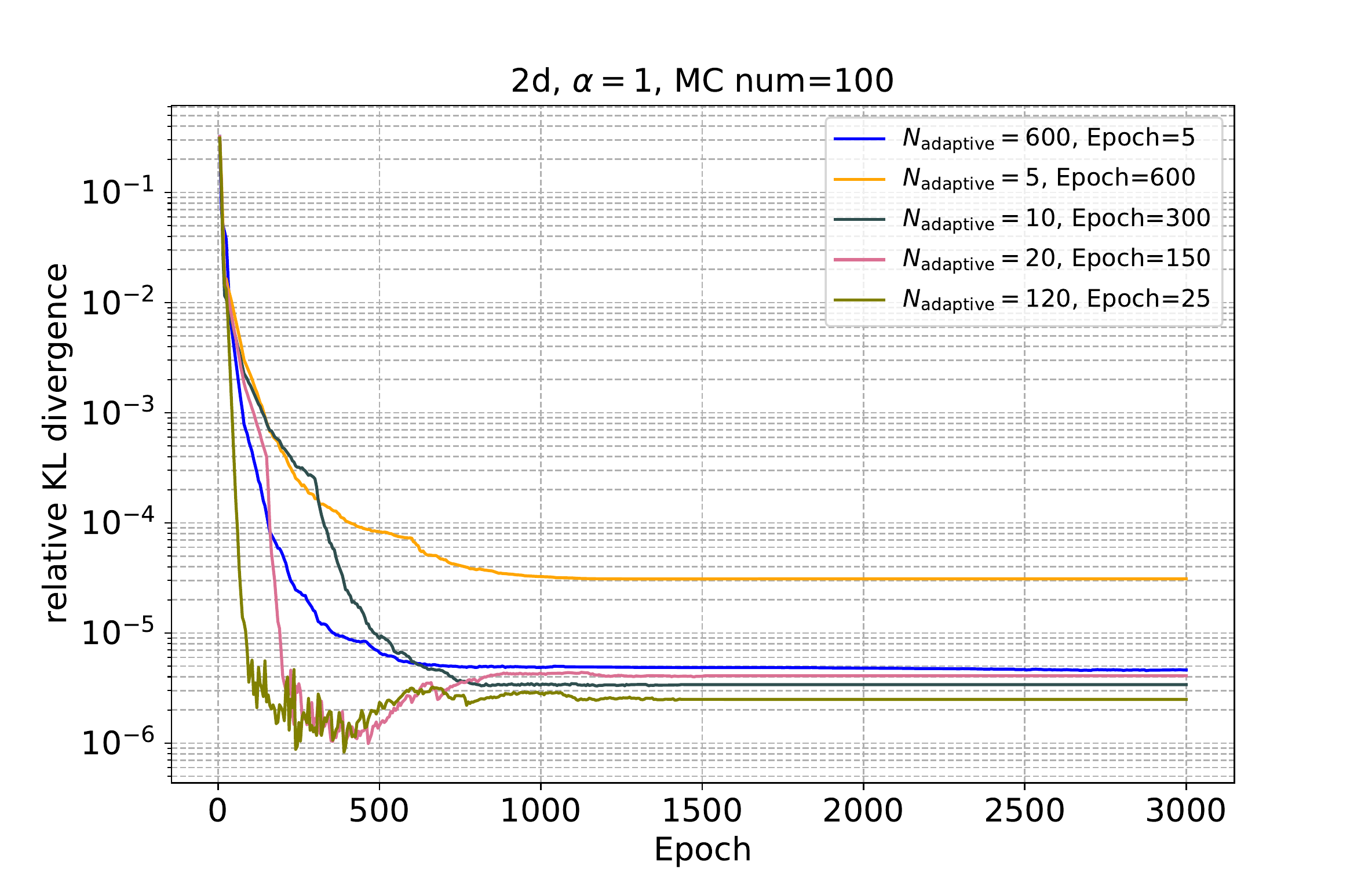}}
	\subfigure[Different adaptive frequencies for GRBFNF.]{
		\includegraphics[height=3.5cm,width=5.5cm]{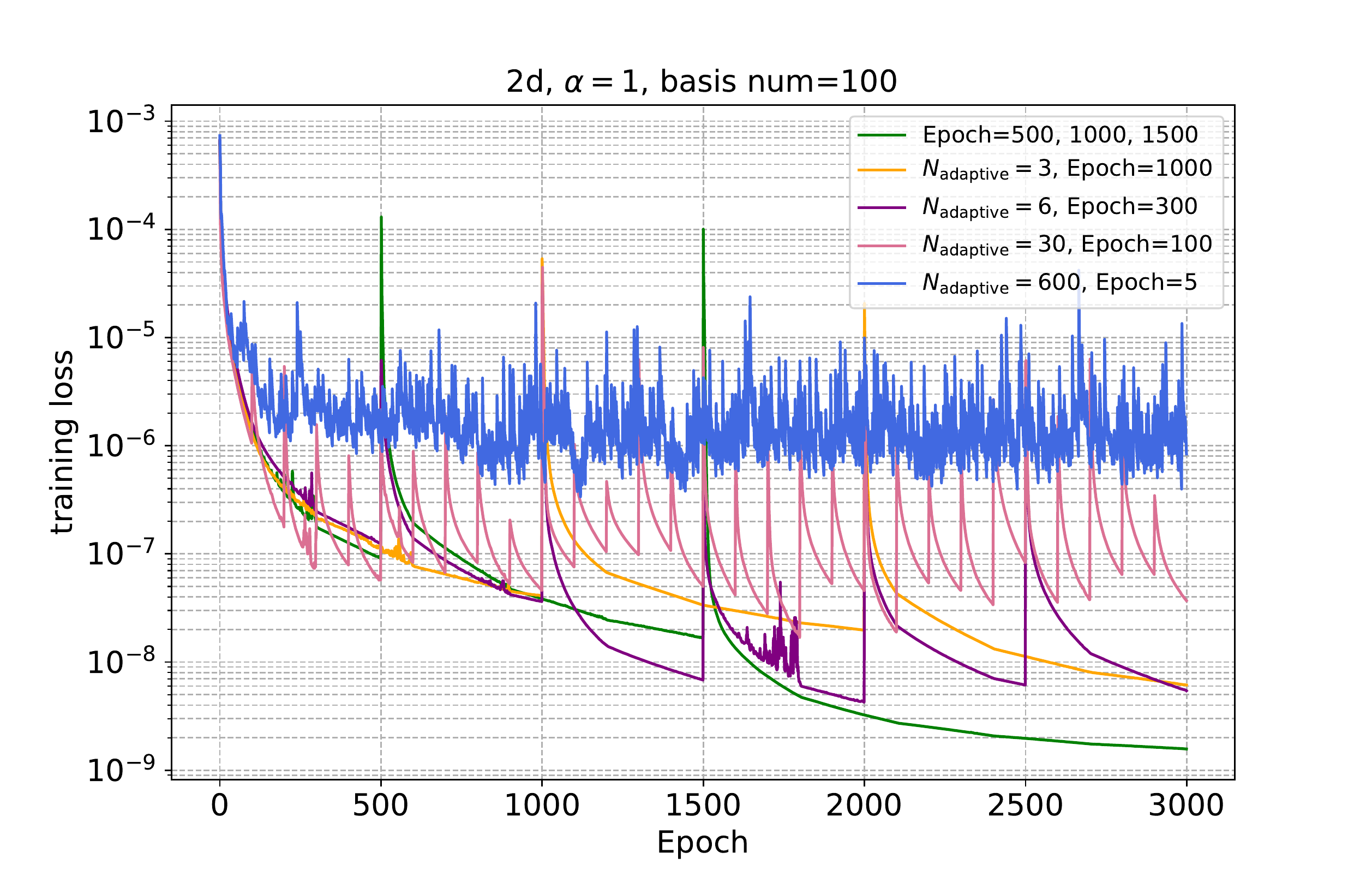}
		\includegraphics[height=3.5cm,width=5.5cm]{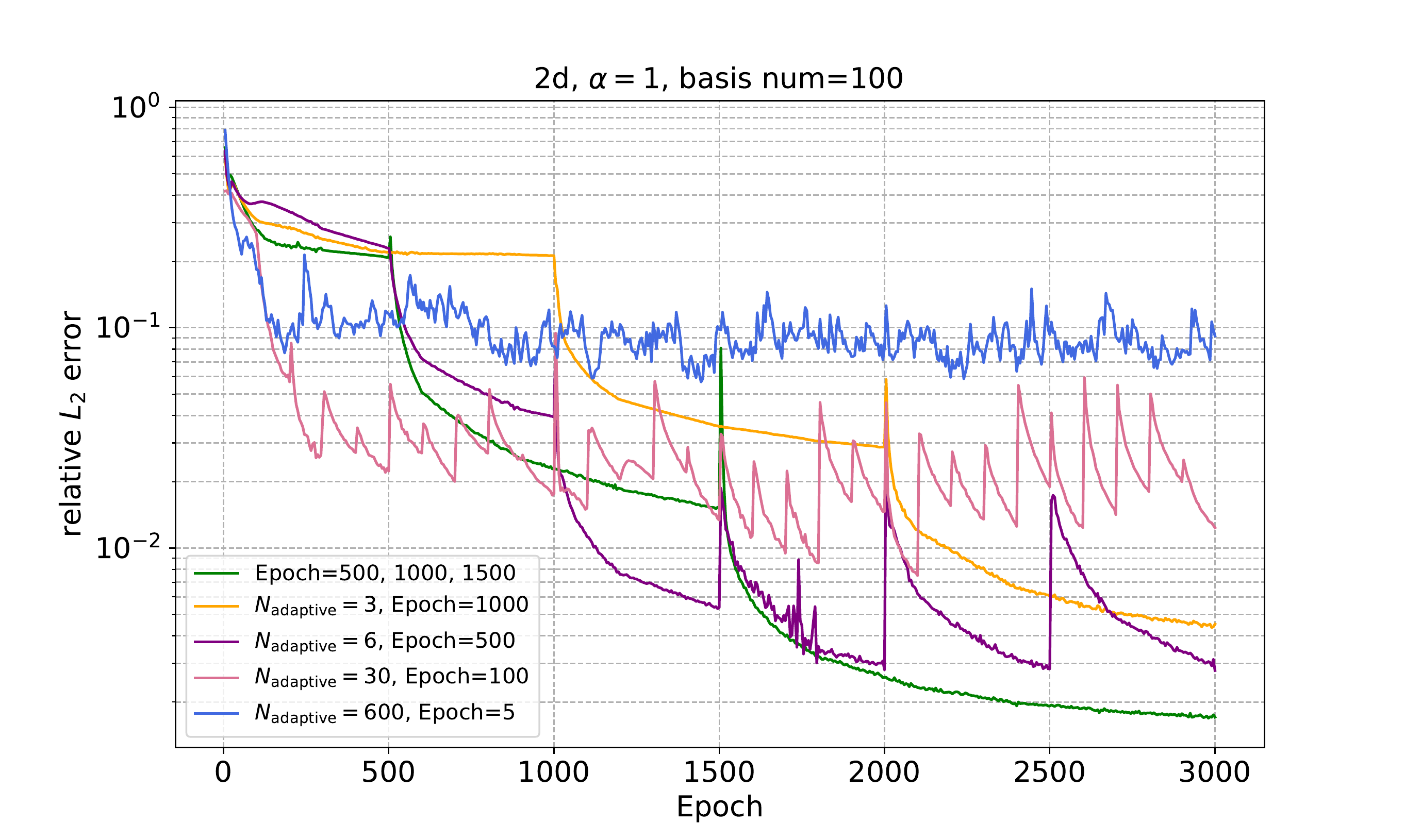}
		\includegraphics[height=3.5cm,width=5.5cm]{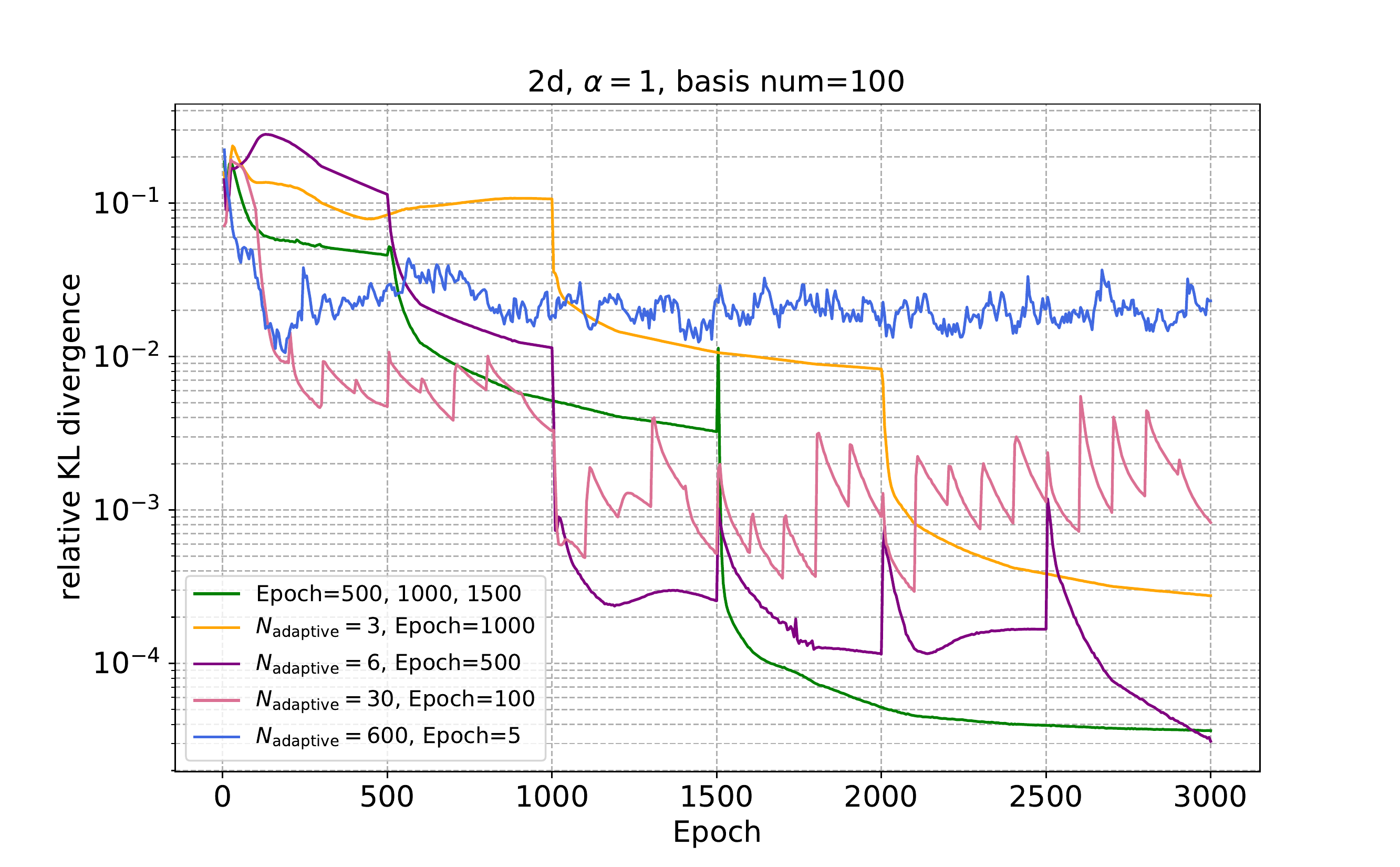}}
	\caption{FPE with only fractional Laplacian. Left: Training loss. Middle: The relative $L_2$ error. Right: The relative KL divergence.}
	\label{gauss_frac_N}
\end{figure}

We first discuss the training strategy of MCNF and GRBFNF by adjusting the adaptive frequency of training. We present the training loss, relative $L_2$ error and relative KL divergence for different adaptive frequencies in Fig. \ref{gauss_frac_N}. For the MCNF method, increasing the adaptive frequency leads to better results because the MC approximation of the fractional Laplacian is independent of the update of $S$. However, for the GRBFNF method, the adaptivity should not be activated until the current models is well trained, otherwise, the loss may be stuck in the transition period induced by the re-initialization of current models.
\begin{figure}[h!]
	\centering
	\begin{minipage}[b]{0.32\linewidth}
		\includegraphics[height=3.5cm,width=5.5cm]{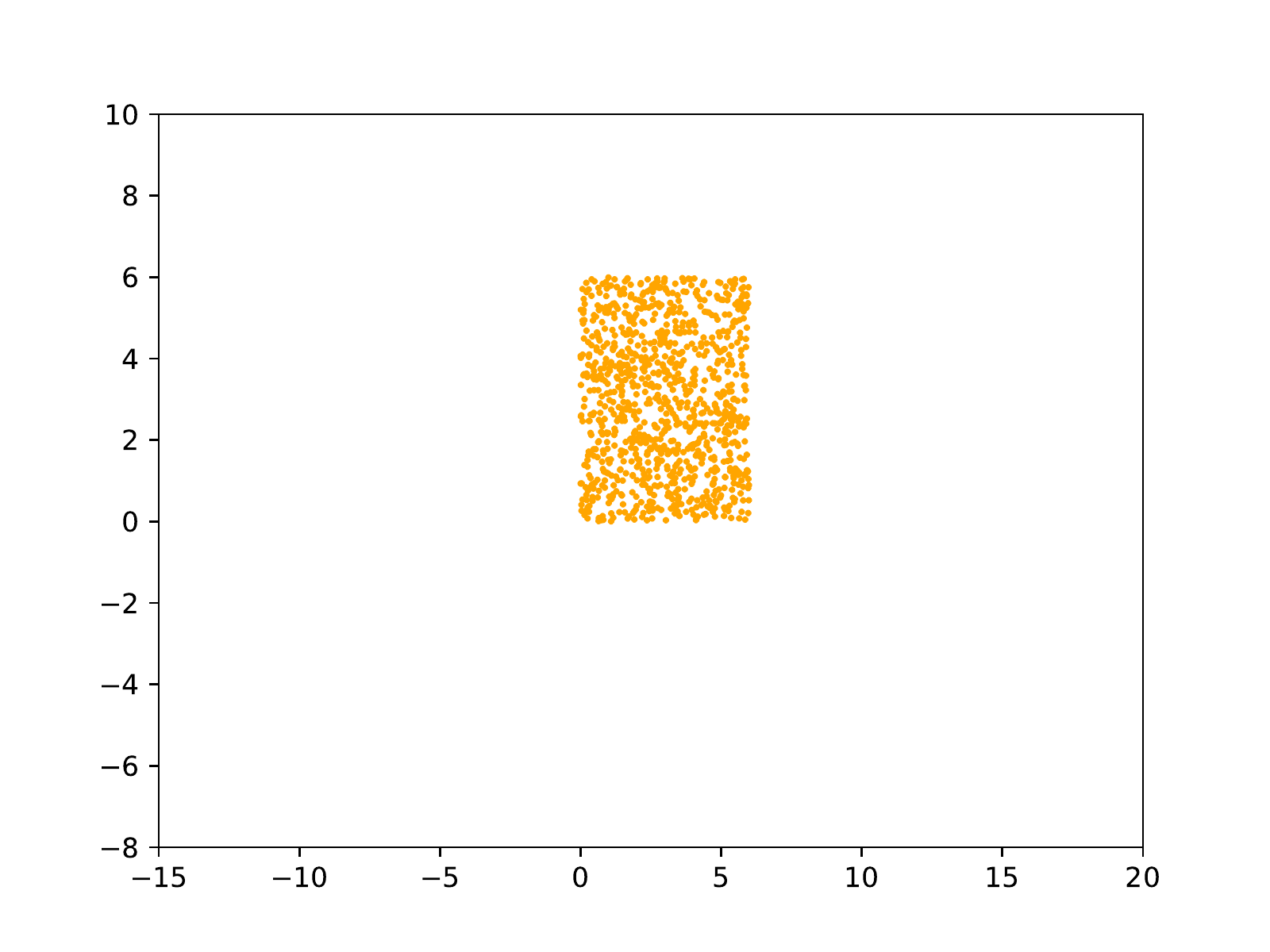}
	\end{minipage}
	\begin{minipage}[b]{0.32\linewidth}
		\includegraphics[height=3.5cm,width=5.5cm]{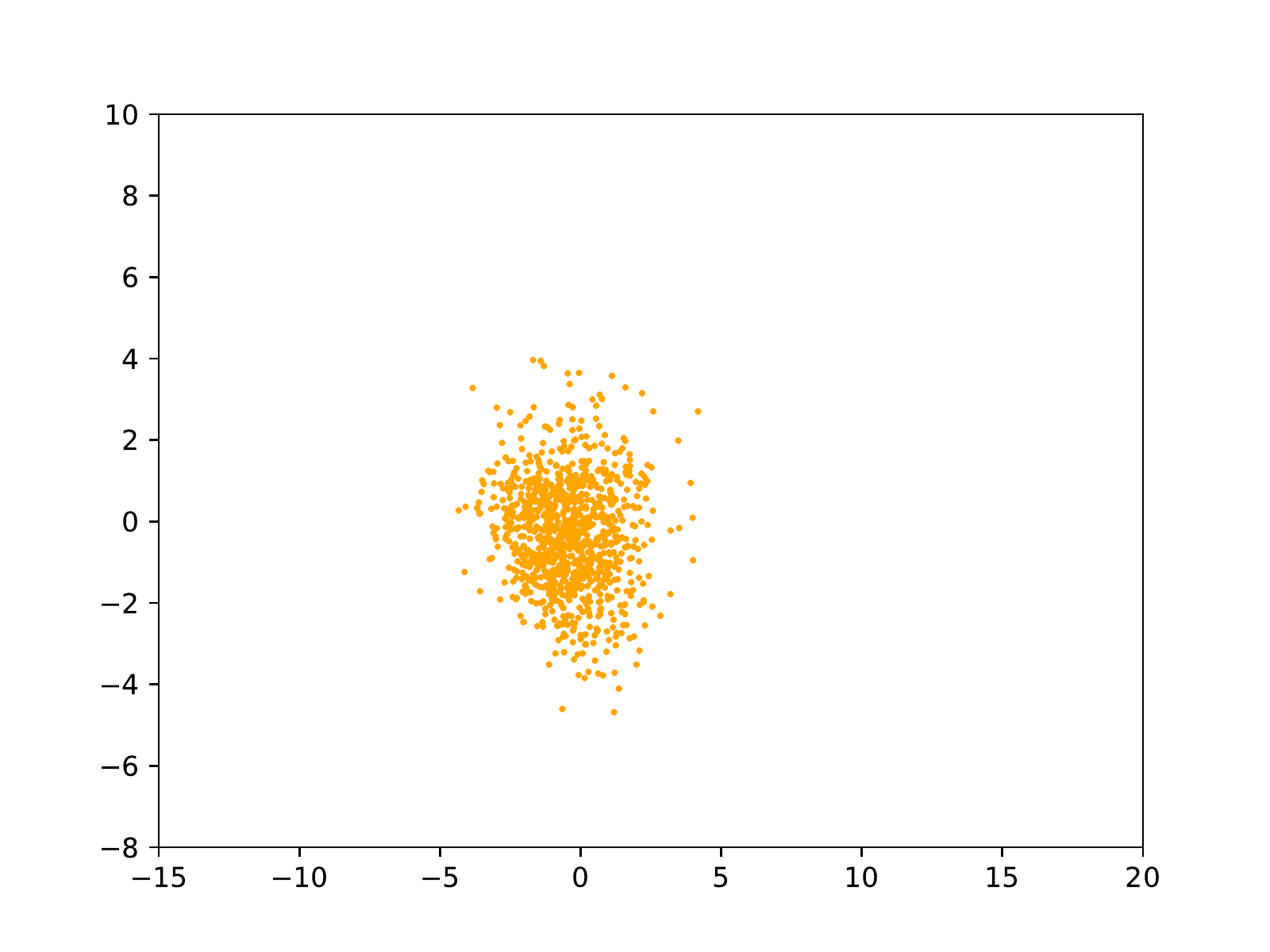}
	\end{minipage}
	\begin{minipage}[b]{0.32\linewidth}
		\includegraphics[height=3.5cm,width=5.5cm]{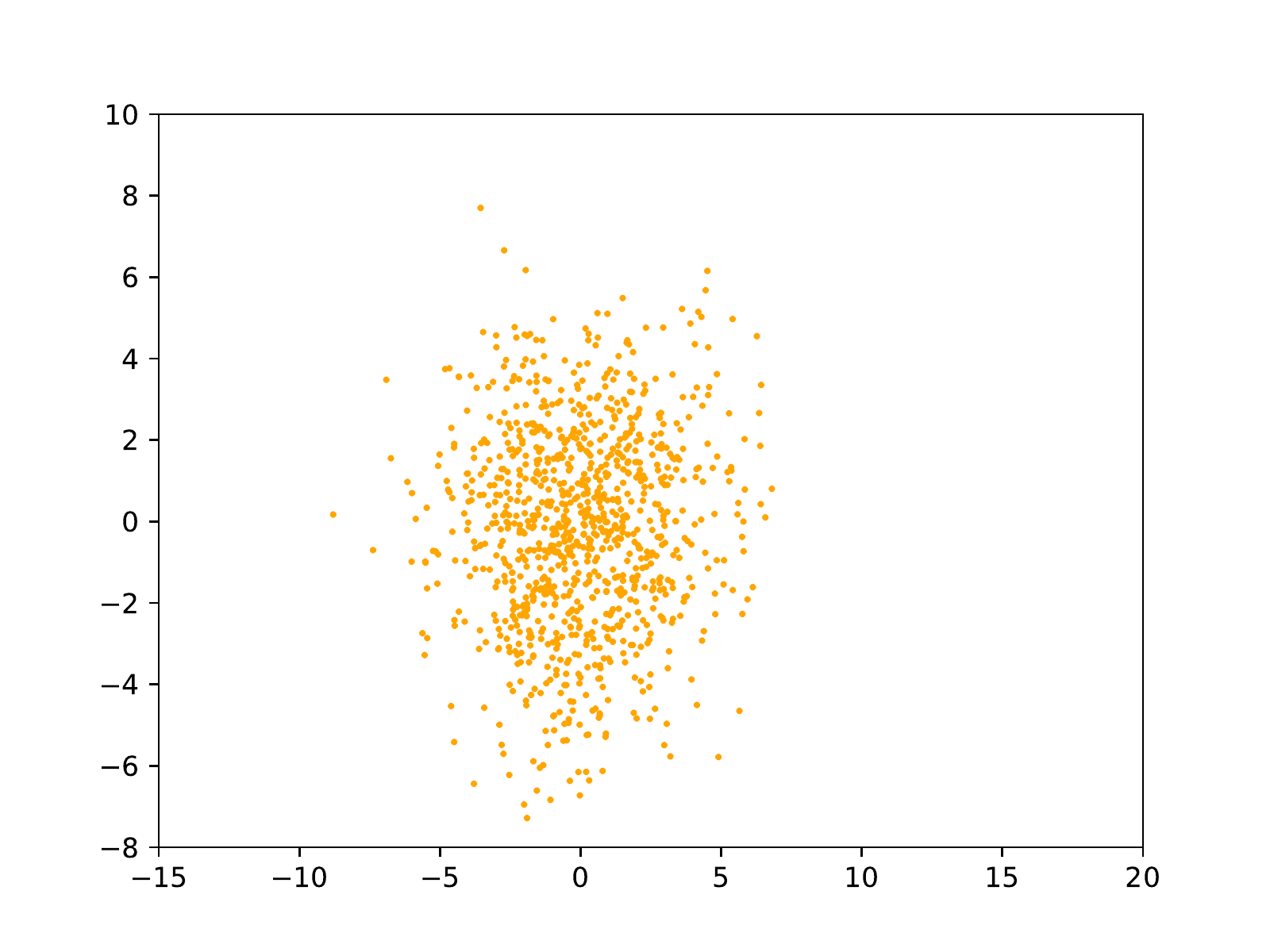}
	\end{minipage}
	
	\begin{minipage}[b]{0.32\linewidth}
		\includegraphics[height=3.5cm,width=5.5cm]{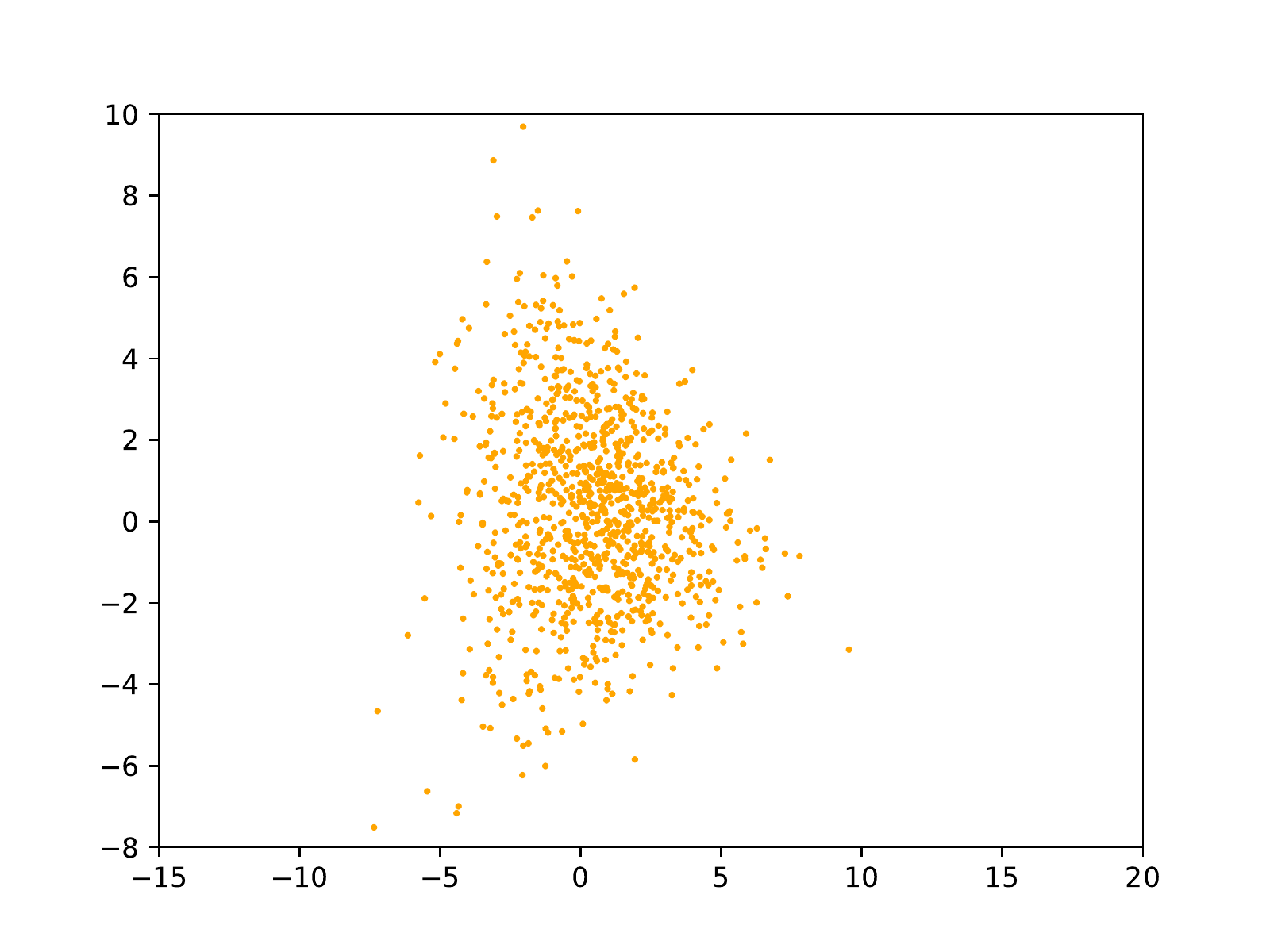}
	\end{minipage}
	\begin{minipage}[b]{0.32\linewidth}
		\includegraphics[height=3.5cm,width=5.5cm]{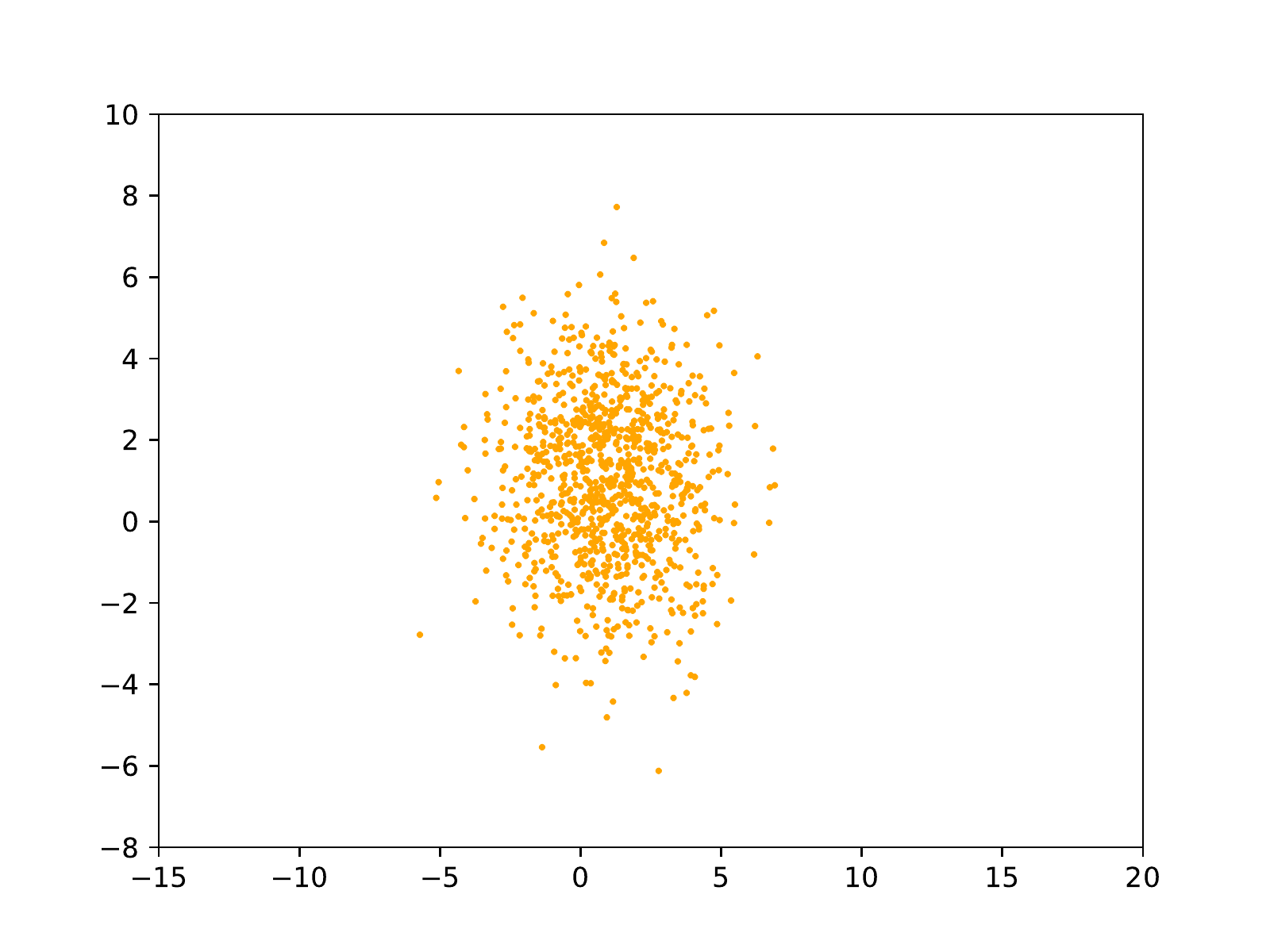}
	\end{minipage}
	\begin{minipage}[b]{0.32\linewidth}
		\includegraphics[height=3.5cm,width=5.5cm]{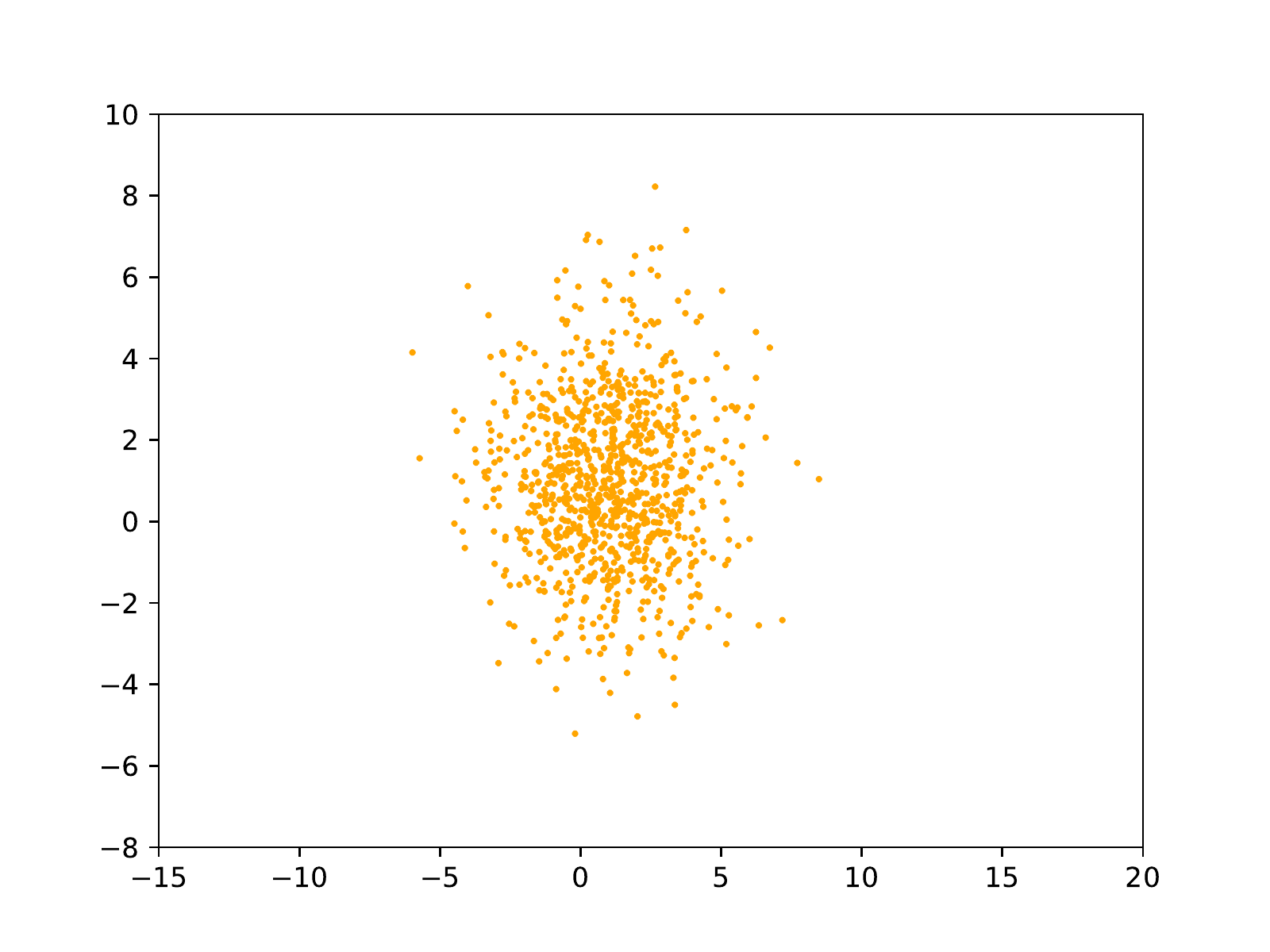}
	\end{minipage}
	\caption{Distribution of training samples at different adaptivity iteration numbers in MCNF. From left to right and from top to bottom, $k=1,2,3,5,30,250$.}
	\label{gauss_frac_adap_MC}
\end{figure}
\begin{figure}[h!]
	\centering
	\subfigure[Training samples at different adaptive iterations. From left to right, $k=1,2,3$.]{
		\includegraphics[height=3.5cm,width=5.5cm]{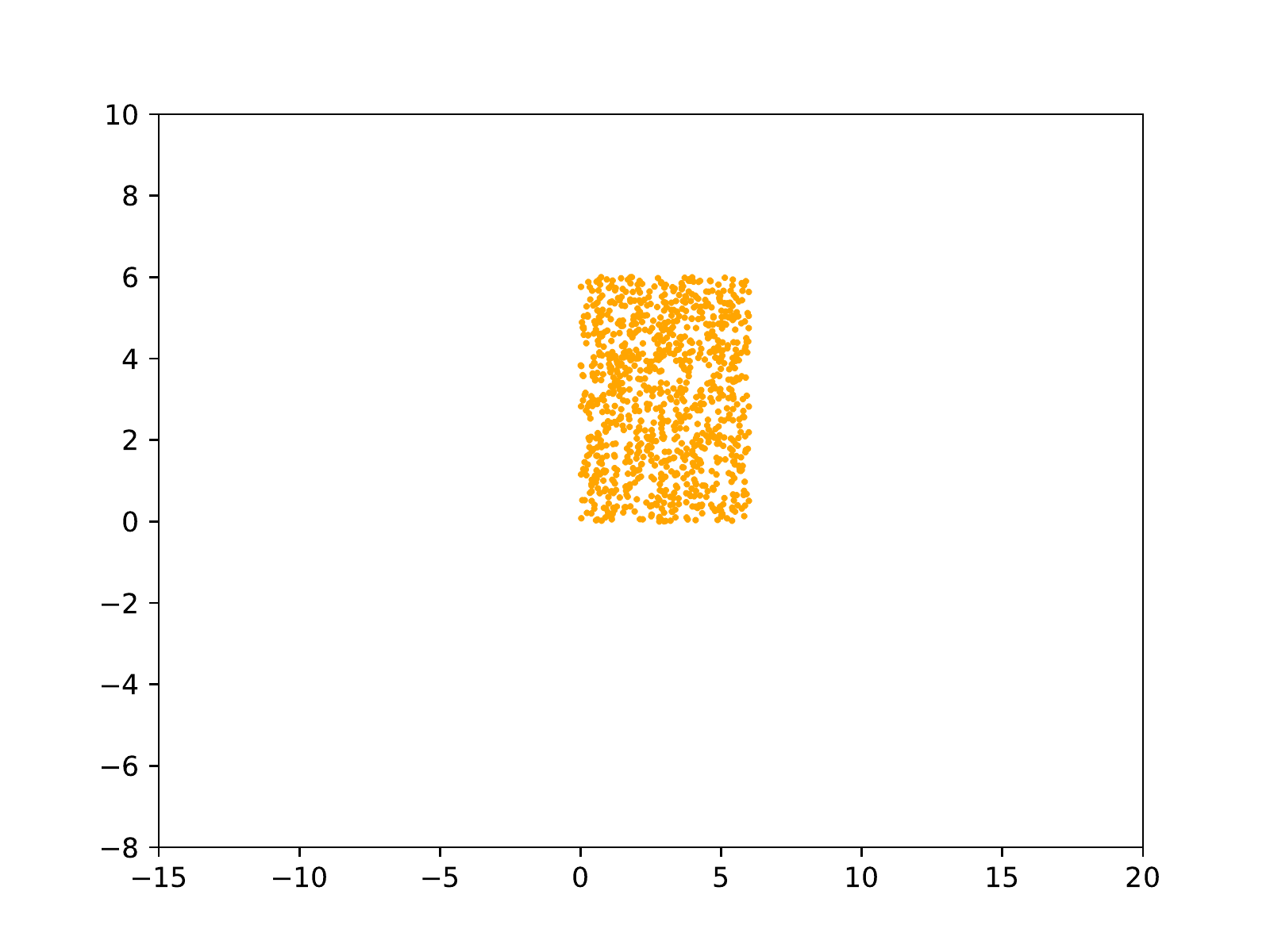}
		\includegraphics[height=3.5cm,width=5.5cm]{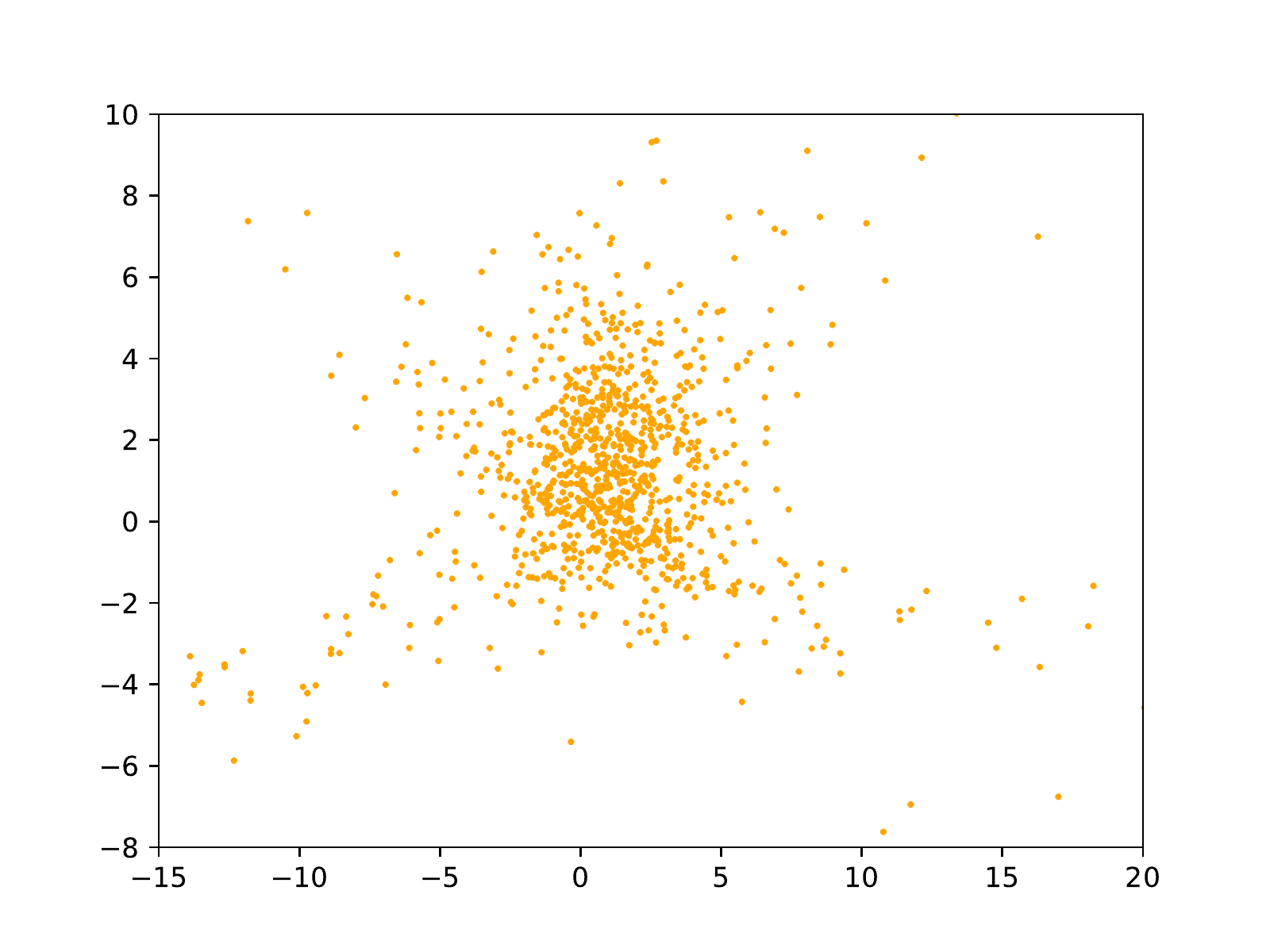}
		\includegraphics[height=3.5cm,width=5.5cm]{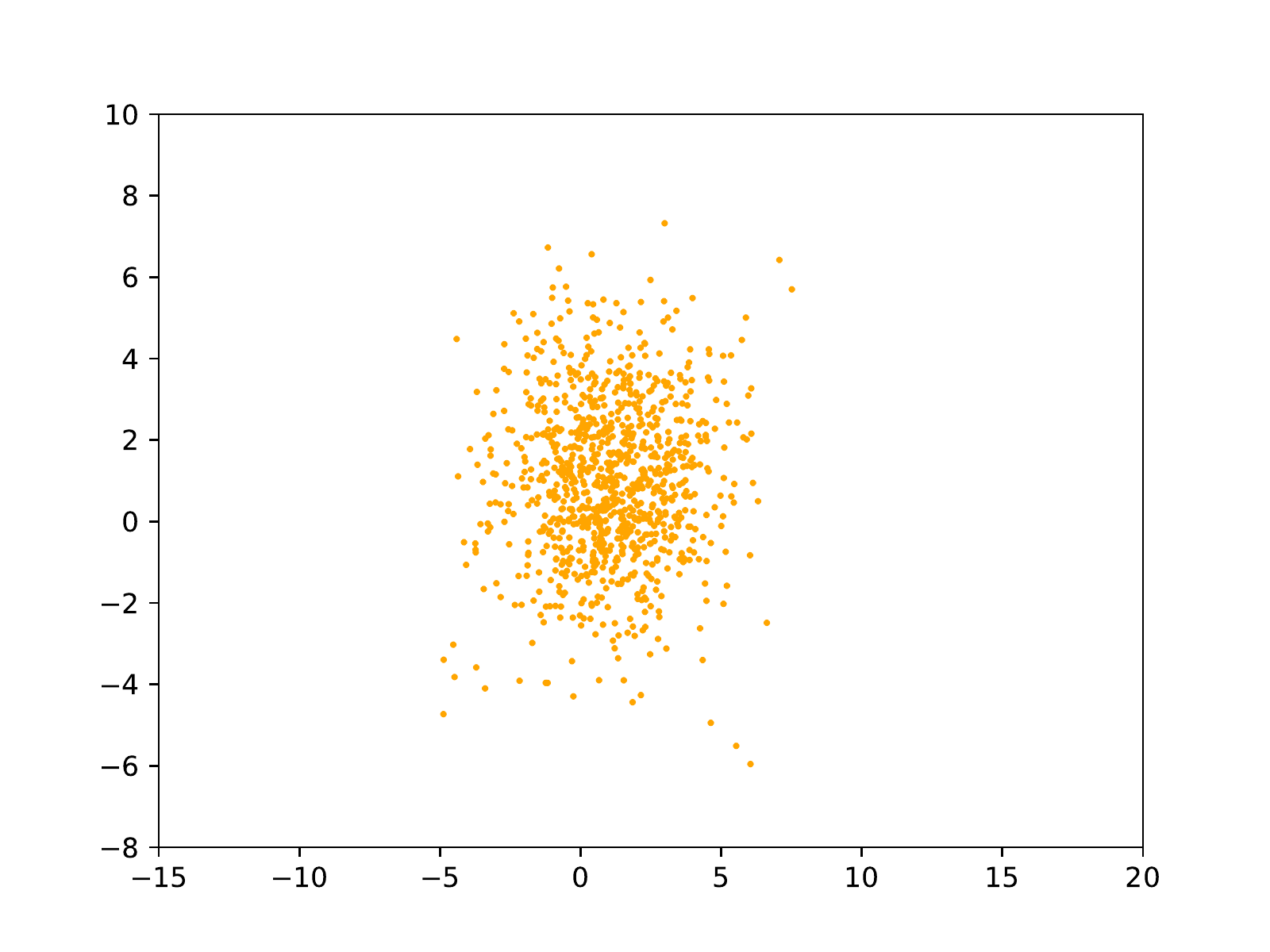}}
	\subfigure[Center points of GRBFs at different adaptive iterations.From left to right, $k=1,2,3$.]{
		\includegraphics[height=3.5cm,width=5.5cm]{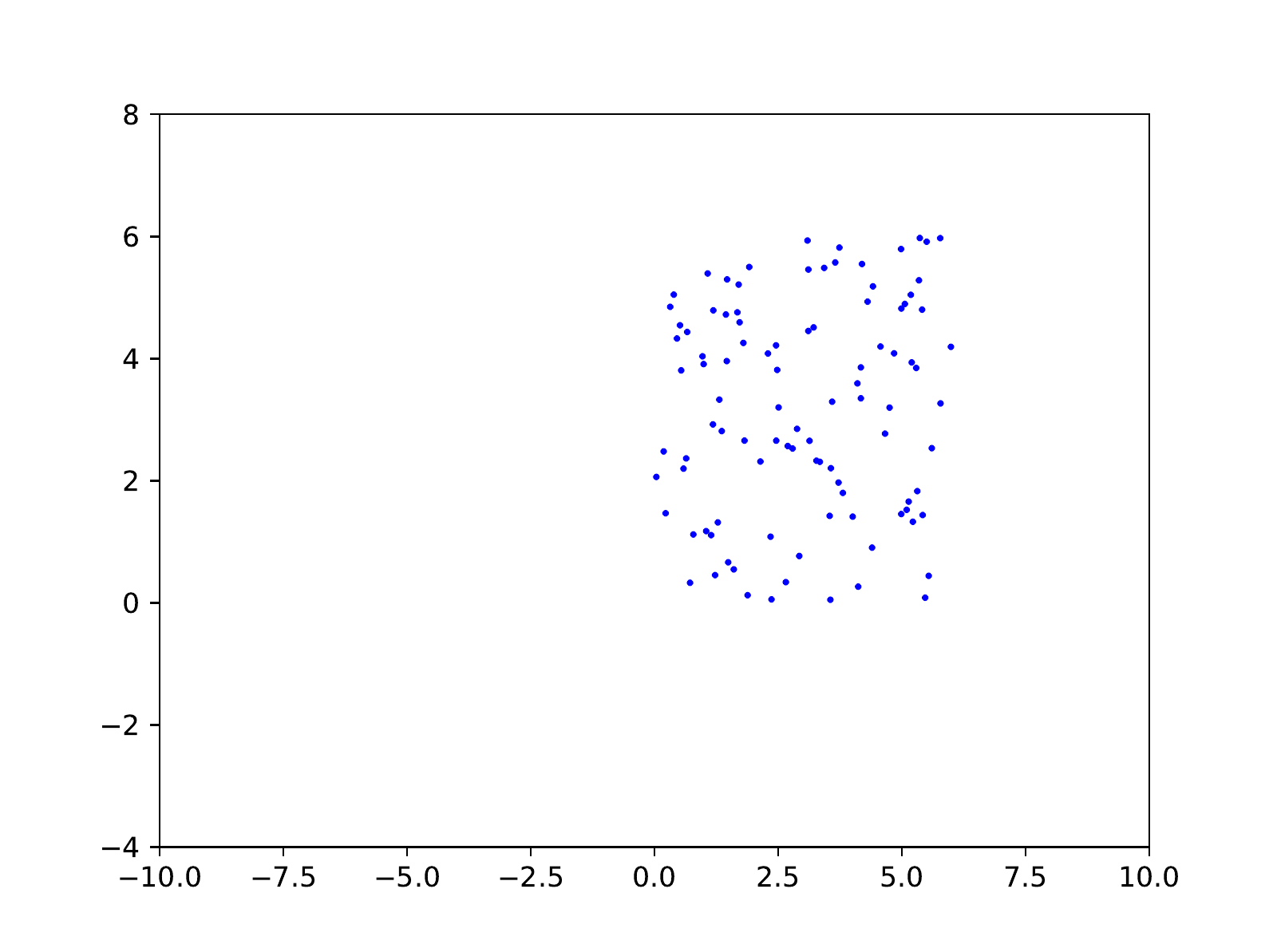}
		\includegraphics[height=3.5cm,width=5.5cm]{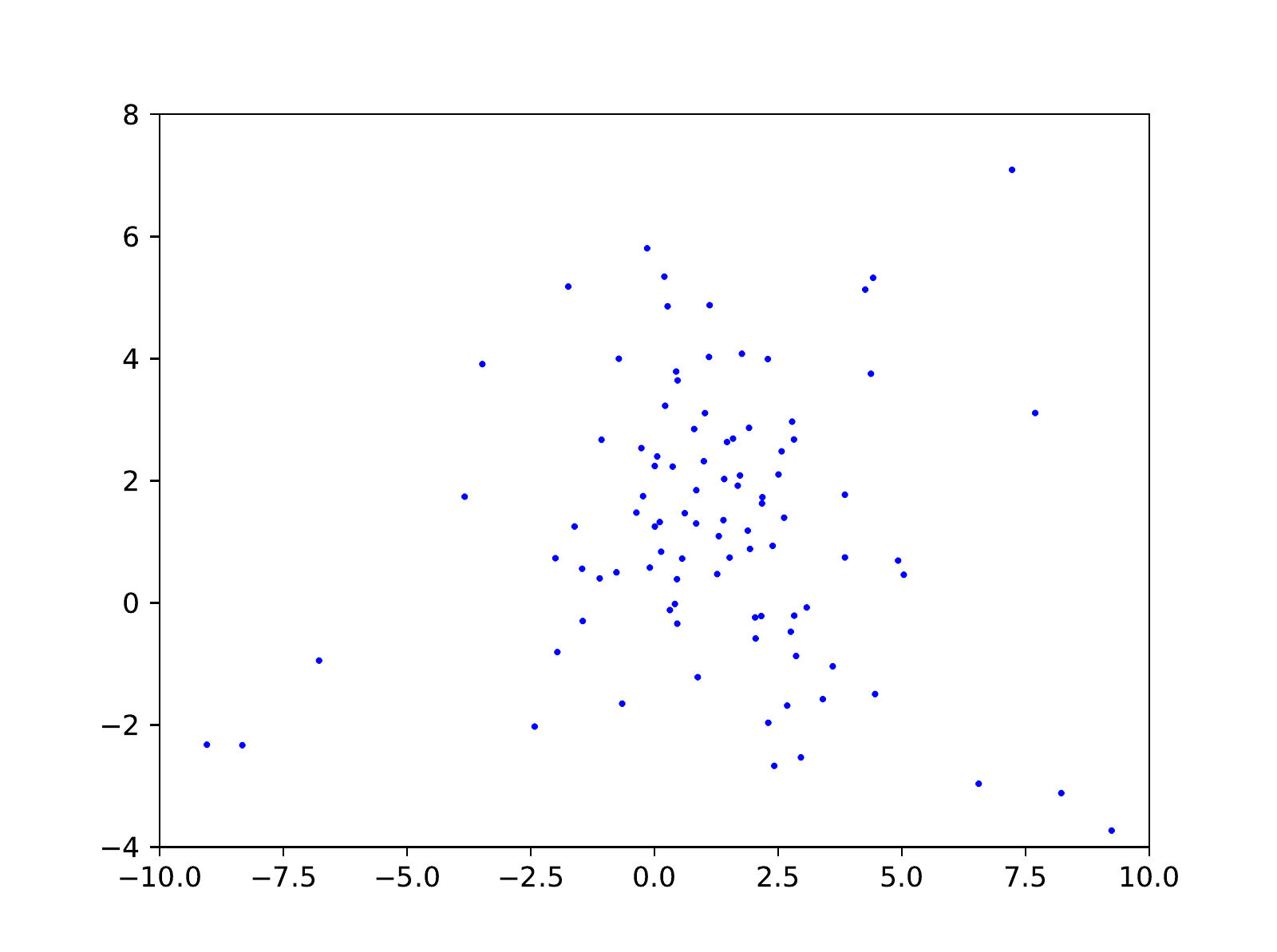}
		\includegraphics[height=3.5cm,width=5.5cm]{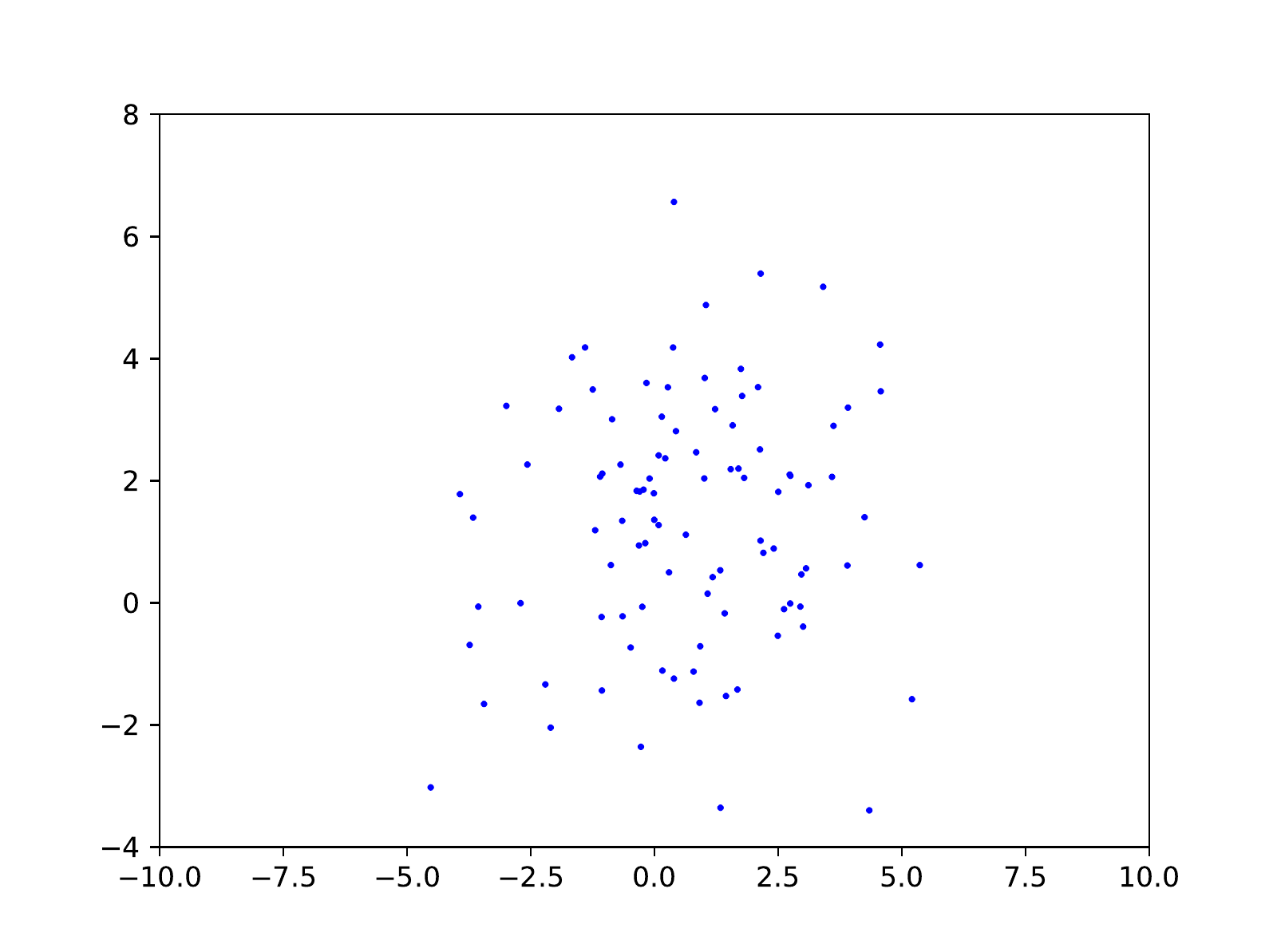}}
	\caption{Adaptivity of GRBFNF for the training set and the centers of GRBF basis functions.}
	\label{gauss_frac_adap_rbf}
\end{figure}

Next, we focus on two experiments to investigate how adaptivity works. Specially, for the MCNF, we choose $600$ adaptivity iterations with 5 epochs for each iteration. And for the GRBFNF, we choose $3$ adaptivity iterations with increasing epochs, 500 epochs in the first adaptivity iteration,  1000 epochs in the second adaptivity iteration and 2000 epochs in the last adaptivity iteration. The time cost of MCNF and GRBFNF is 64 minutes, 46 minutes respectively.
The training points as well as center points of the basis functions for different  adaptivity iteration numbers are presented in Fig. \ref{gauss_frac_adap_MC} and Fig. \ref{gauss_frac_adap_rbf}. One can clearly observe that the training points and center points of the basis functions become increasingly closer to the ground truth as the iteration number increases, showing that adaptive sampling scheme is effective.
\begin{figure}[h]
	\centering
	\subfigure[MCNF, $\alpha=1$.]{
		\includegraphics[height=3.5cm, width=5.5cm]{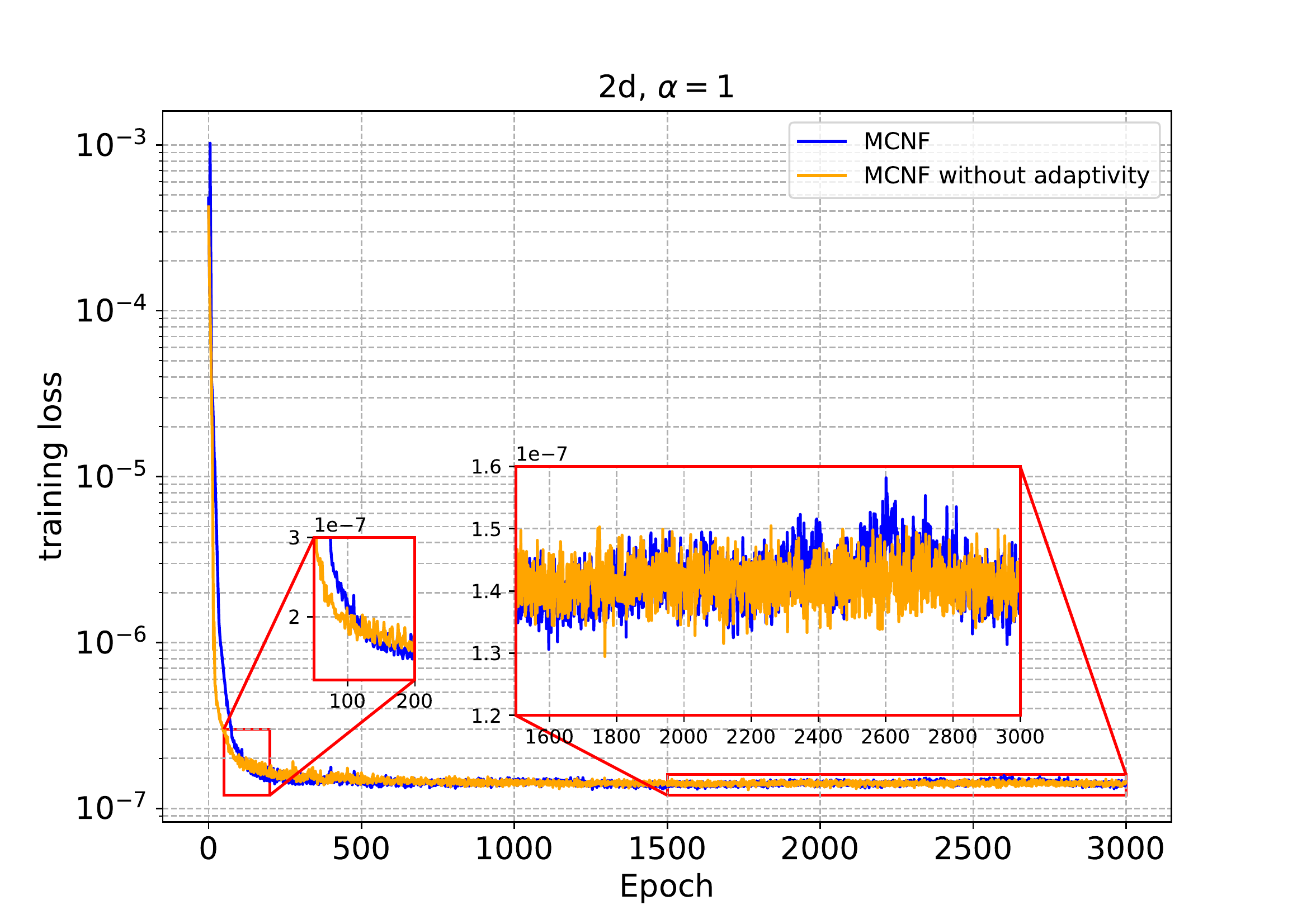}
		\includegraphics[height=3.5cm, width=5.5cm]{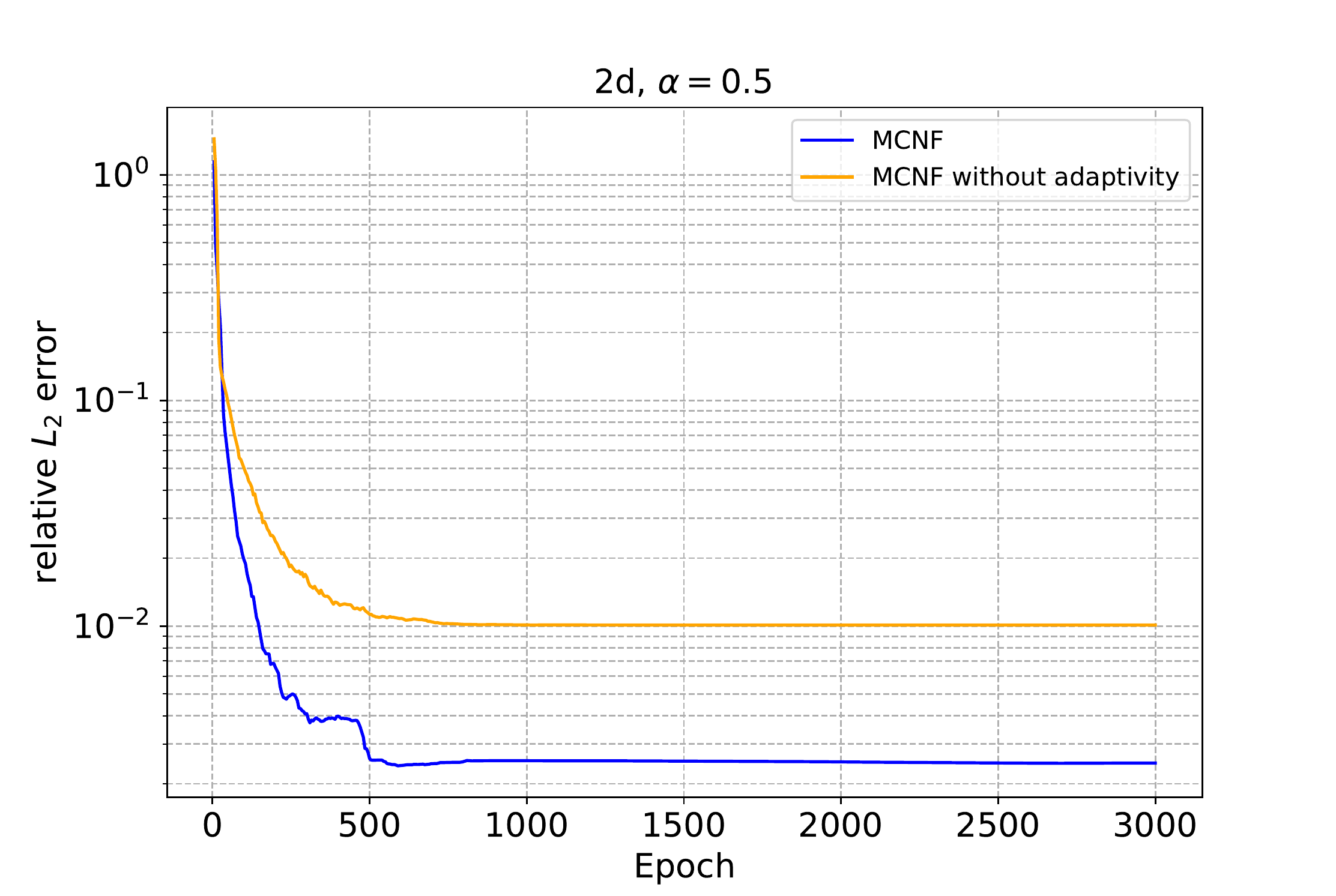}
		\includegraphics[height=3.5cm, width=5.5cm]{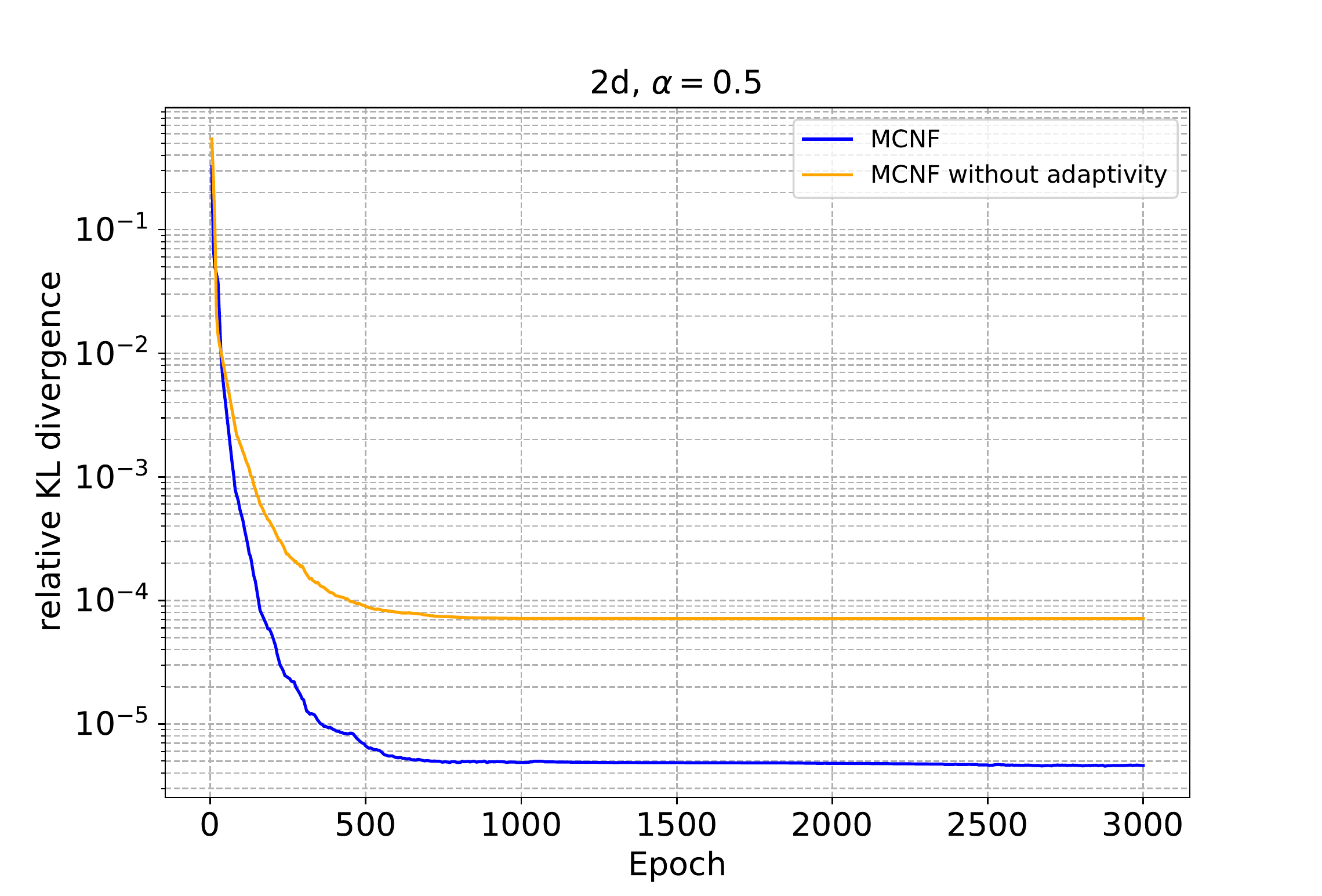}}
	\subfigure[GRBFNF, $\alpha=1$]{
		\includegraphics[height=3.5cm, width=5.5cm]{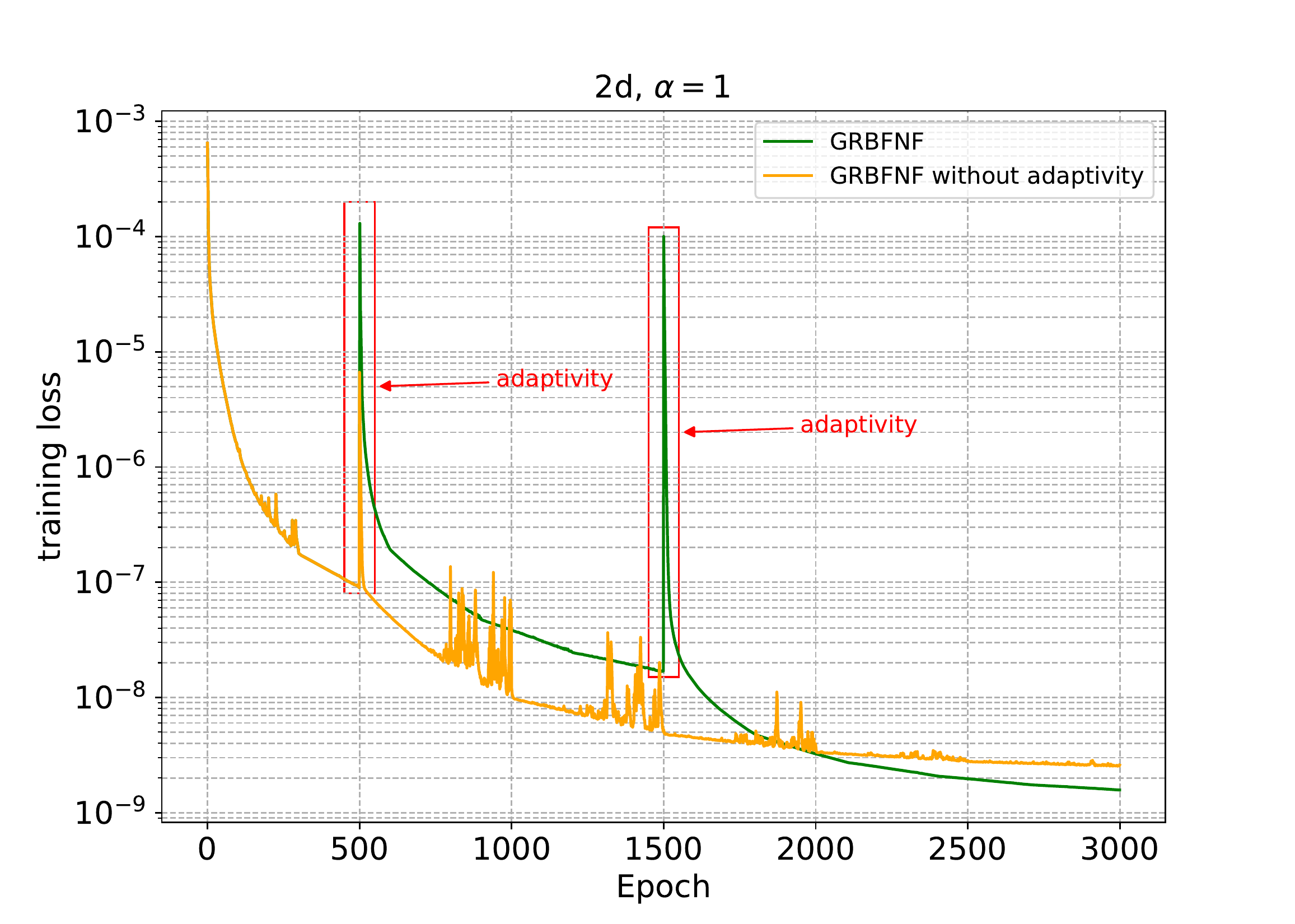}
		\includegraphics[height=3.5cm, width=5.5cm]{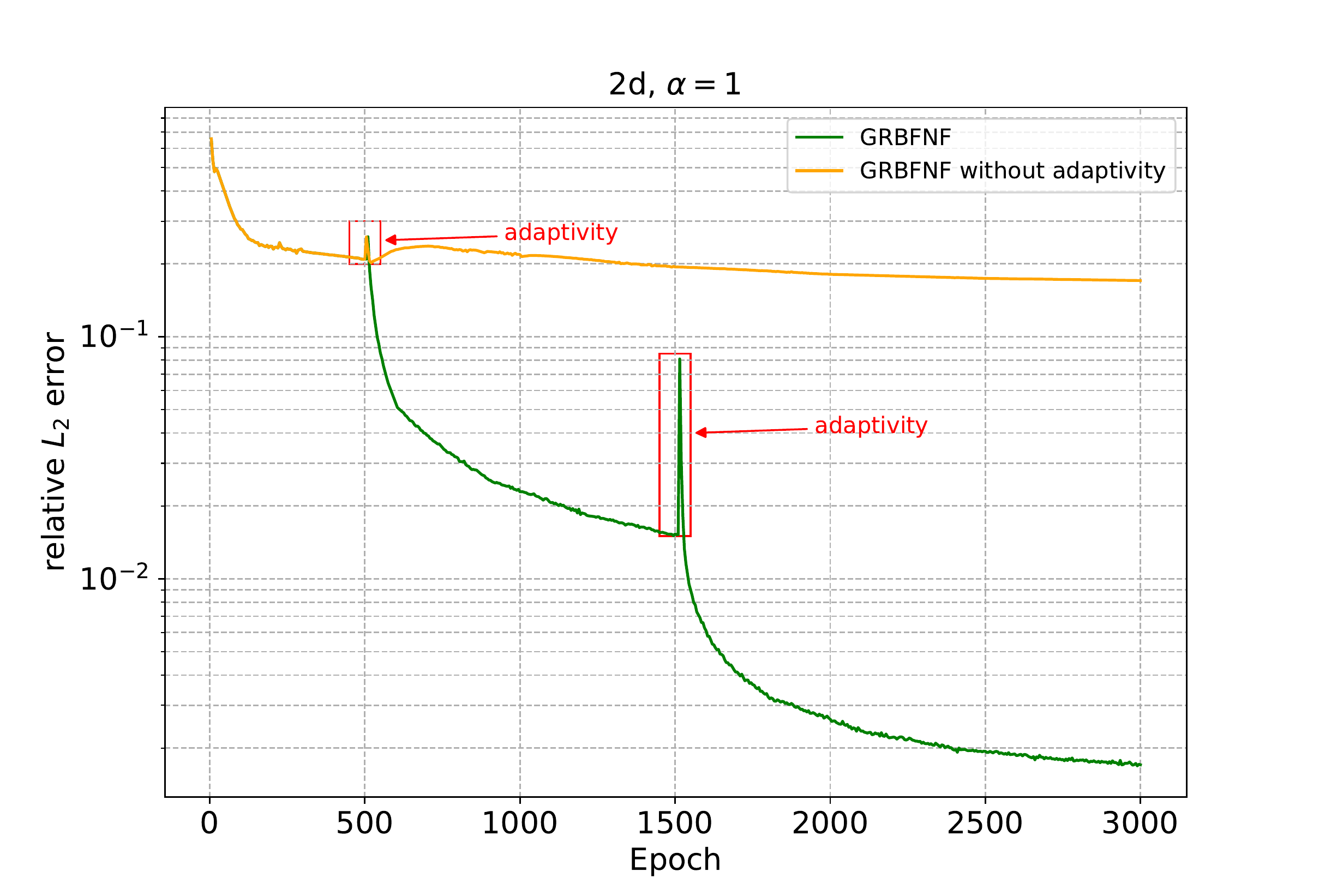}
		\includegraphics[height=3.5cm, width=5.5cm]{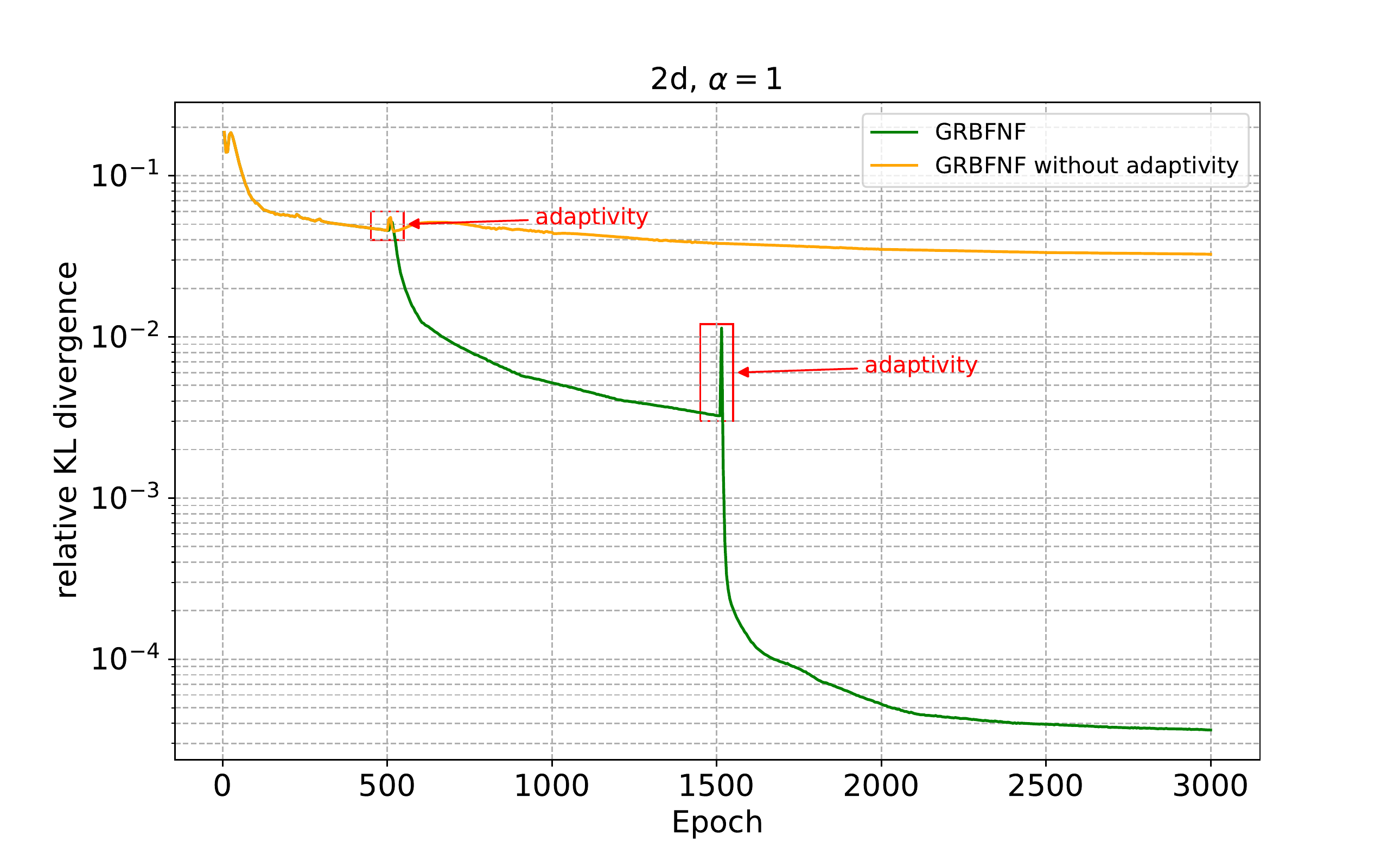}}
	\caption{Comparison between adaptive and non-adaptive methods. Top row: MCNF. Bottom row: GRBFNF. Left: training loss. Middle: relative $L_2$ error. Right: relative KL divergence. }
	\label{frac_MC_RBF_no}
\end{figure}

\begin{figure}[H]
	\centering 
	\includegraphics[height=4.5cm,width=5.5cm]{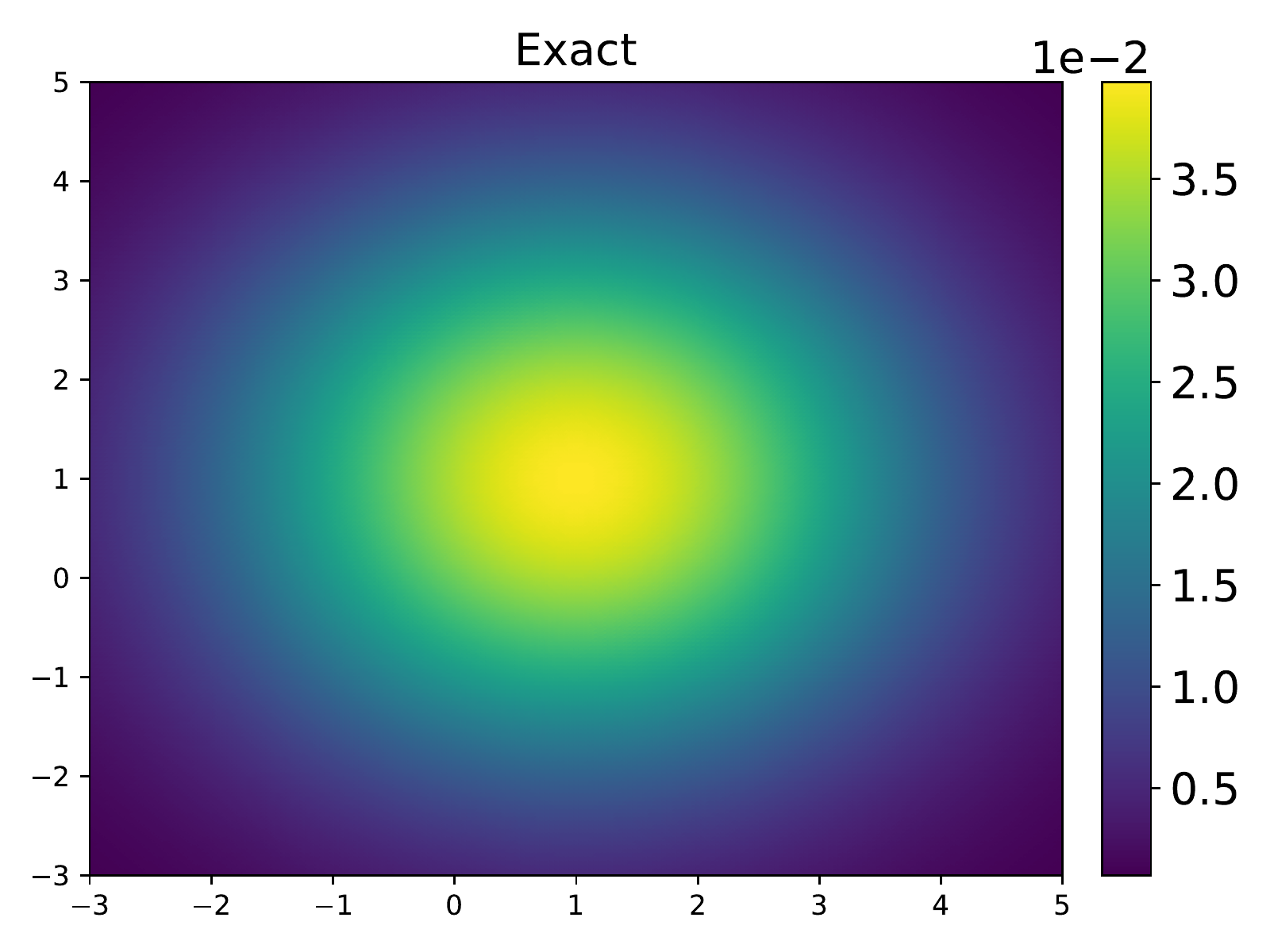}
	\caption{The reference solution of FFP with only the fractional Laplacian term.}
	\label{fig:true_FFP_frac}
\end{figure}

\begin{figure}
	\subfigure[MCNF]{
		\begin{minipage}[b]{0.23\linewidth}
		\includegraphics[height=3.5cm,width=4cm]{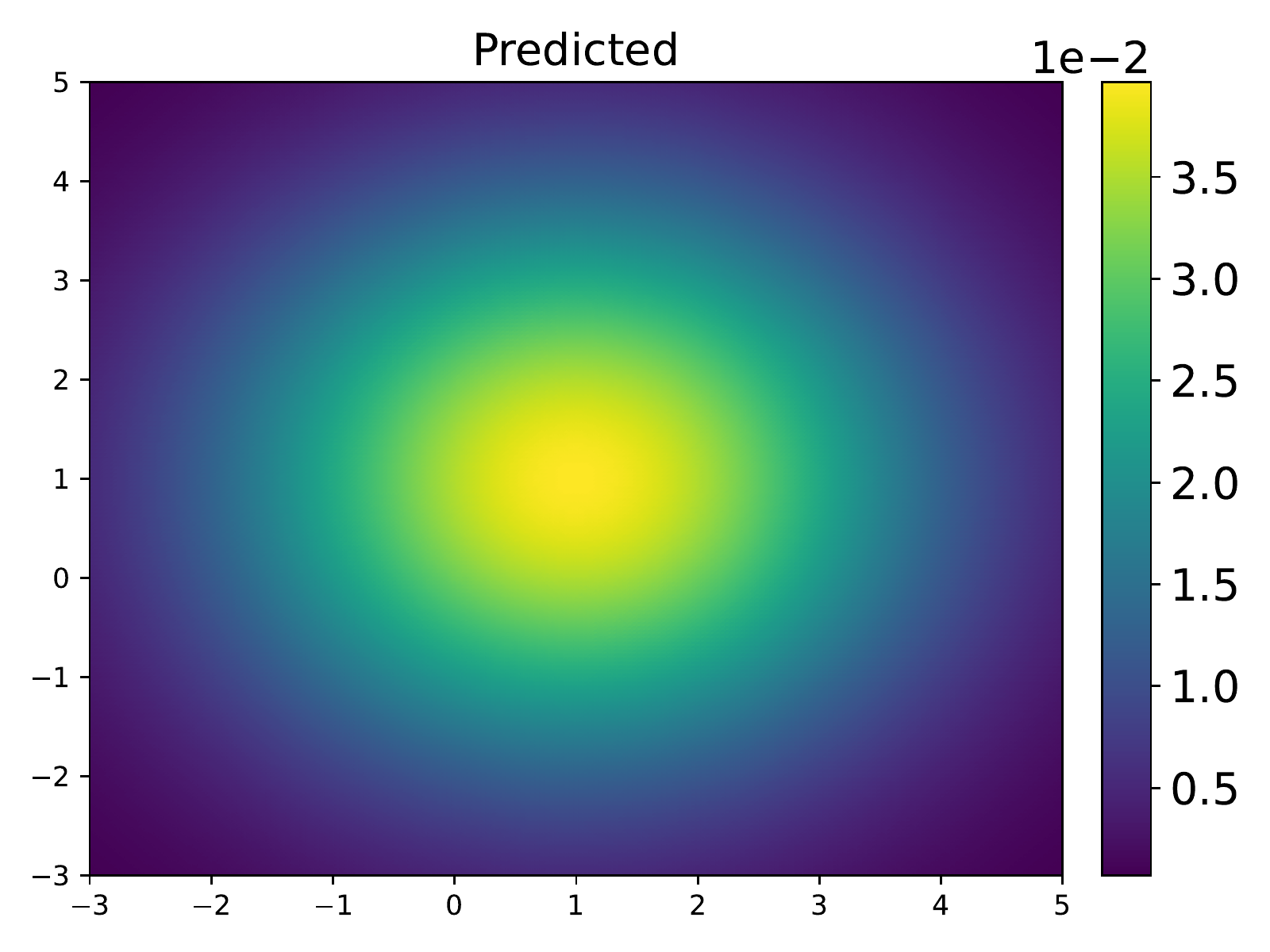}\vspace{4pt}
		\includegraphics[height=3.5cm,width=4cm]{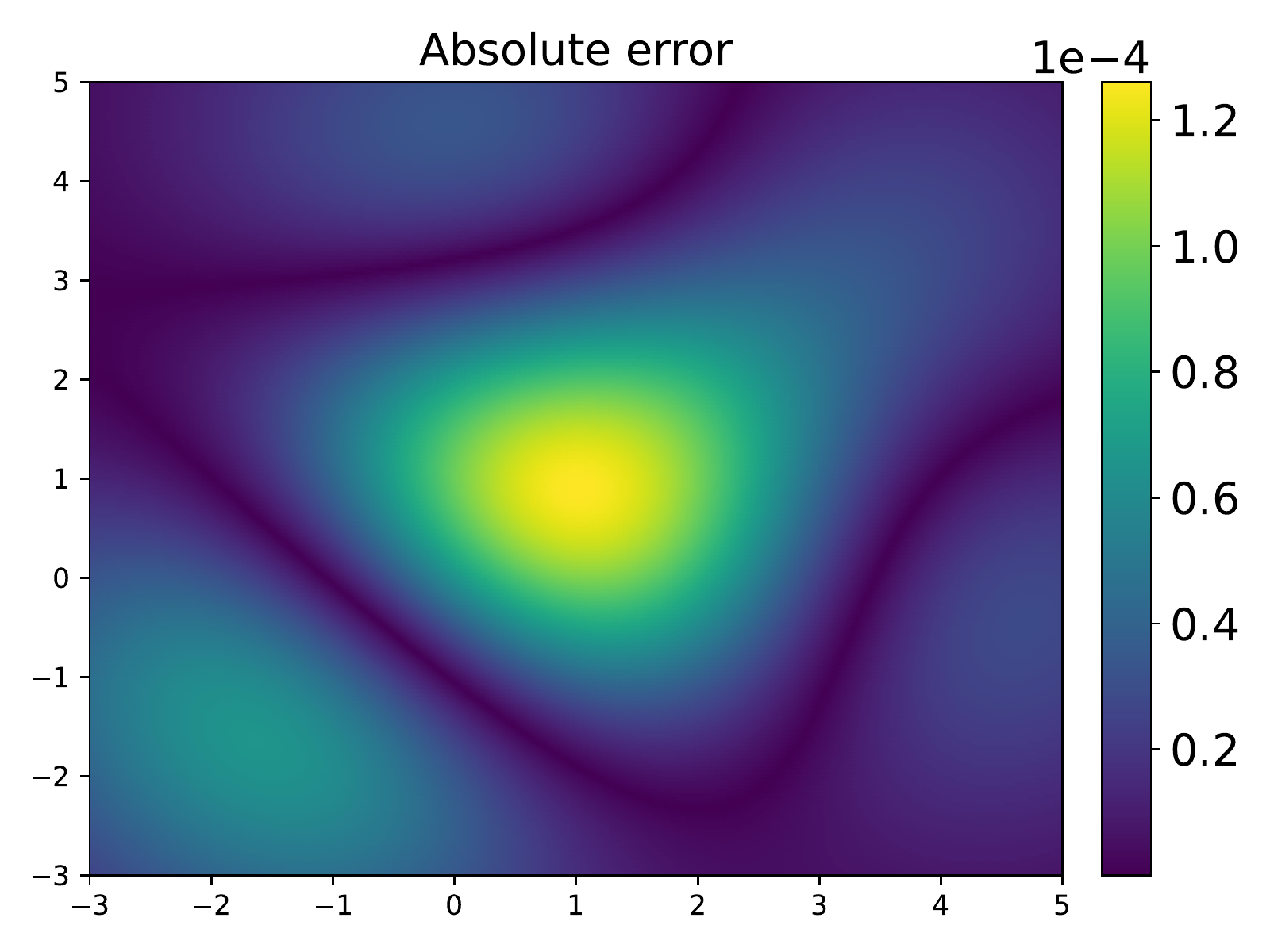}
		\end{minipage}
}
\subfigure[MCNF without adaptivity]{
\begin{minipage}[b]{0.23\linewidth}
	\includegraphics[height=3.5cm,width=4cm]{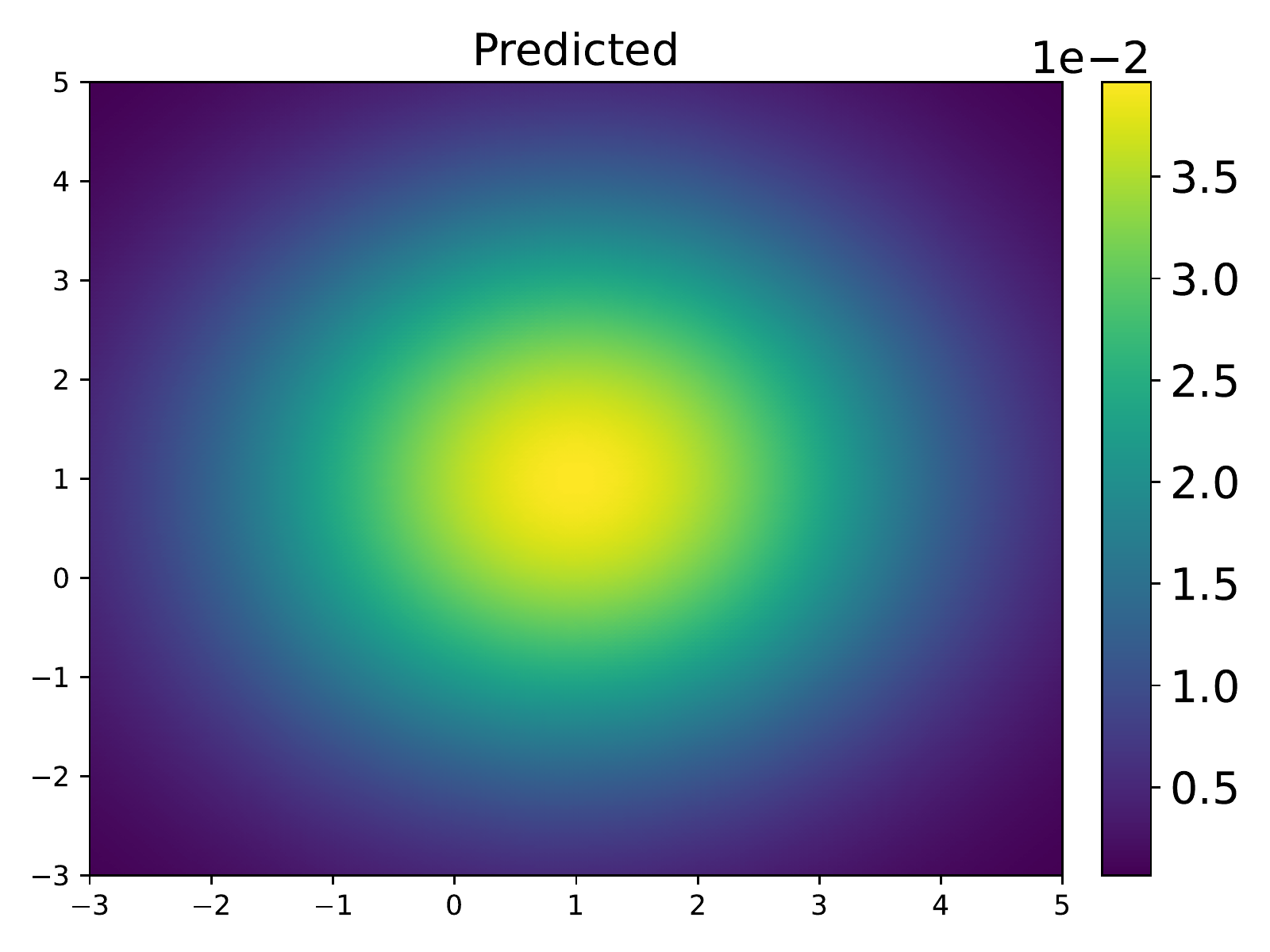}\vspace{4pt}
\includegraphics[height=3.5cm,width=4cm]{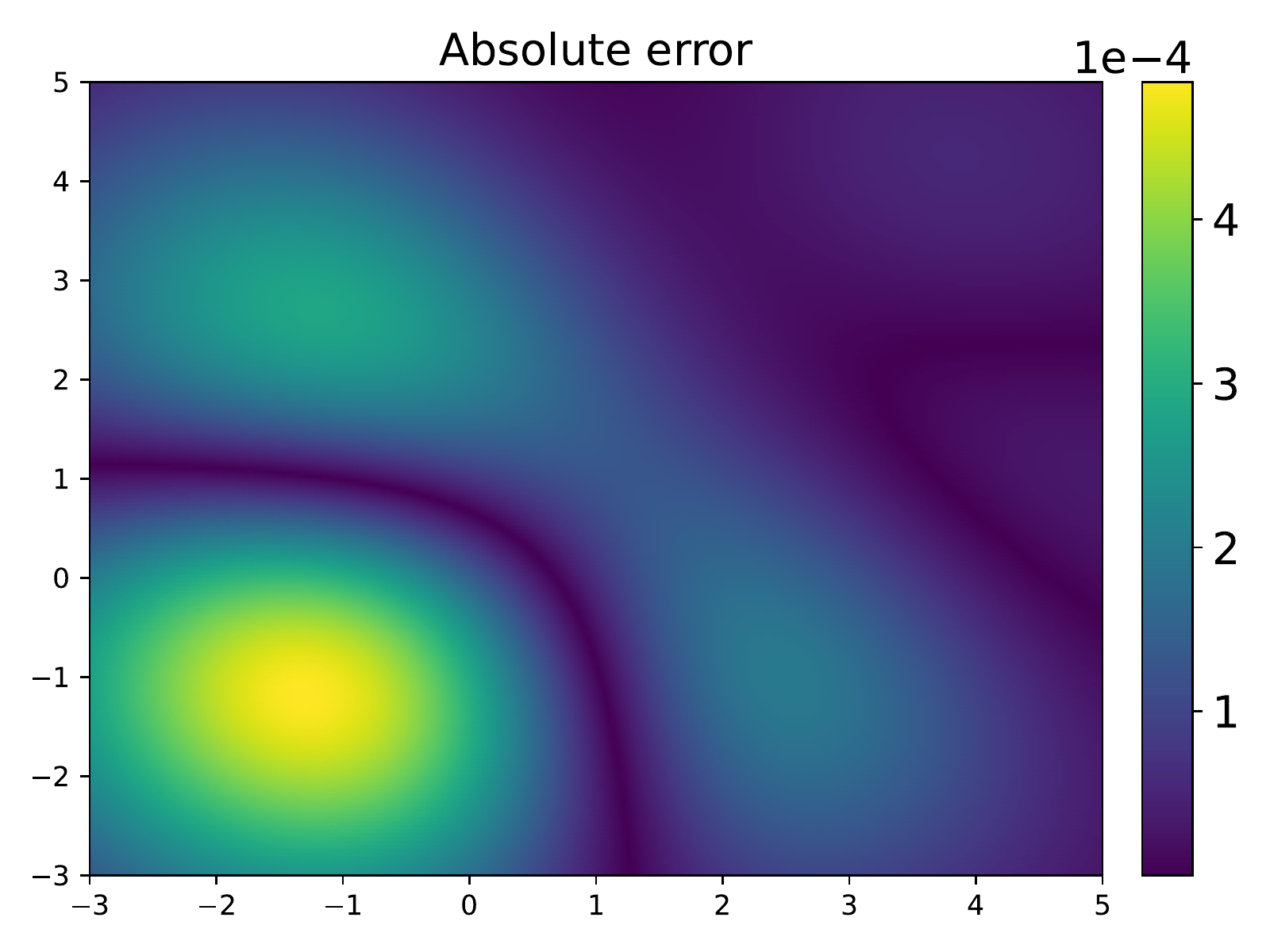}
\end{minipage}
}
\subfigure[GRBFNF]{
	\begin{minipage}[b]{0.23\linewidth}
	\includegraphics[height=3.5cm,width=4cm]{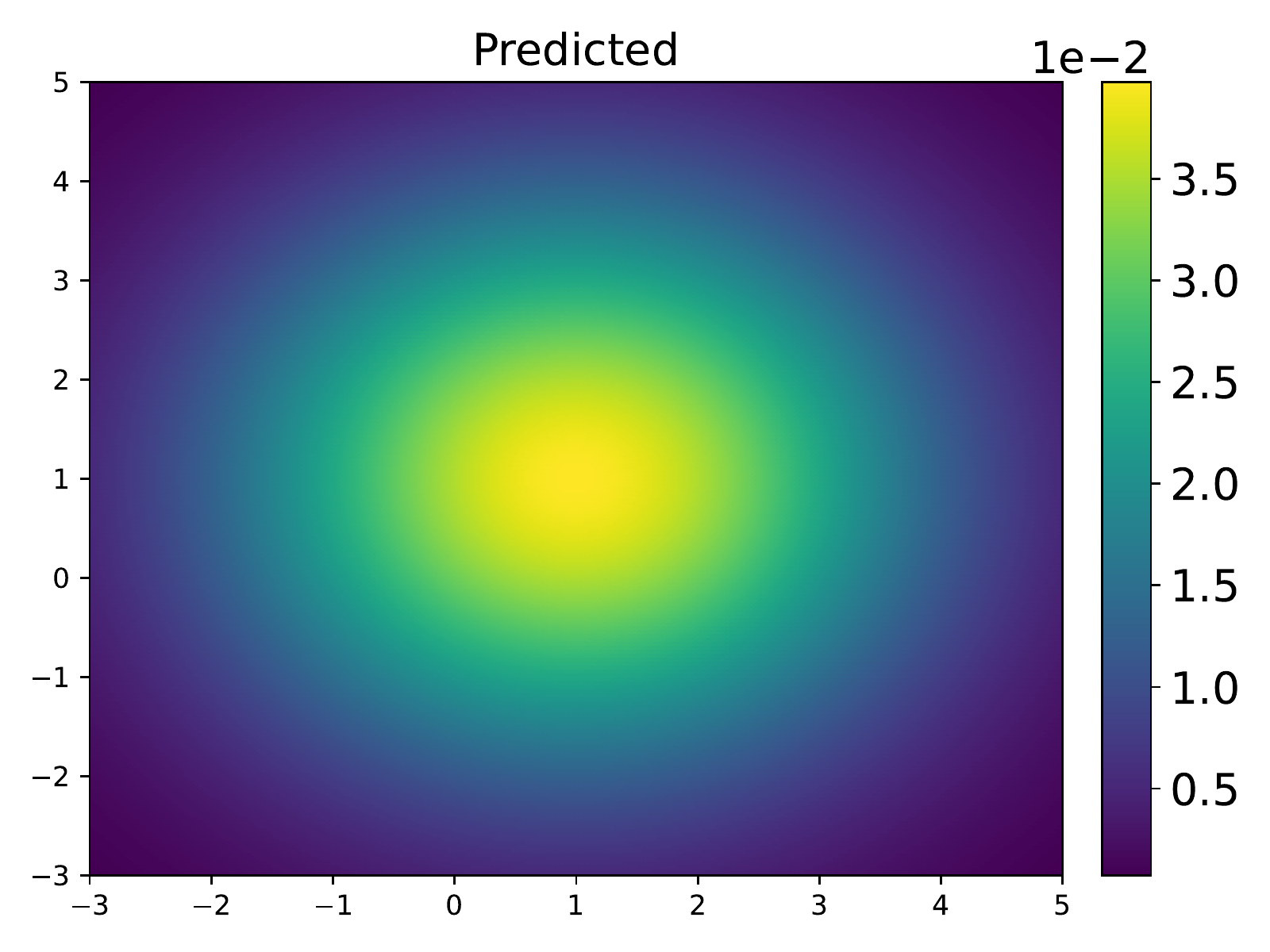}\vspace{4pt}
\includegraphics[height=3.5cm,width=4cm]{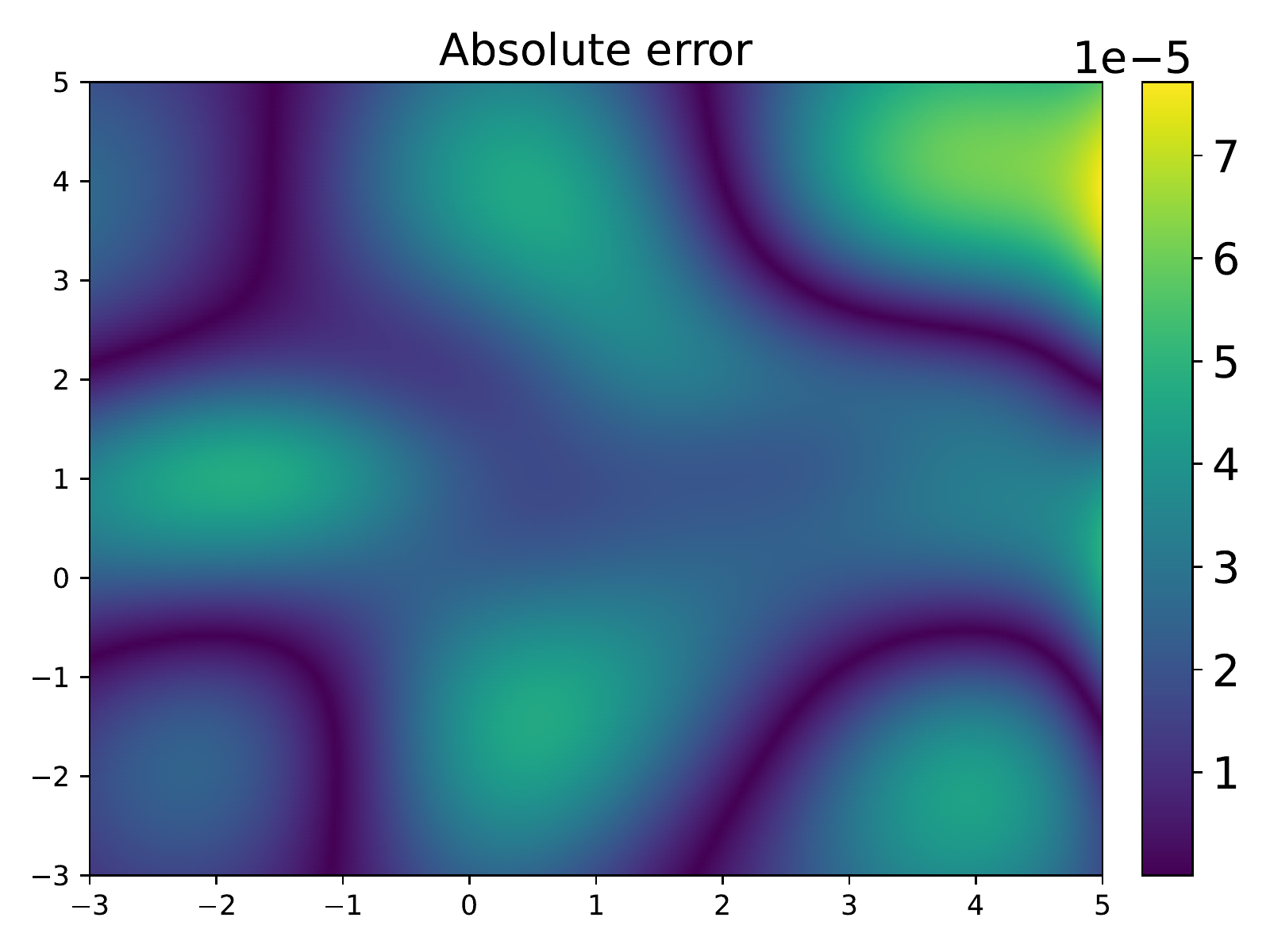}
\end{minipage}
}
\subfigure[GRBFNF without adaptivity]{
	\begin{minipage}[b]{0.24\linewidth}
	\includegraphics[height=3.5cm,width=4cm]{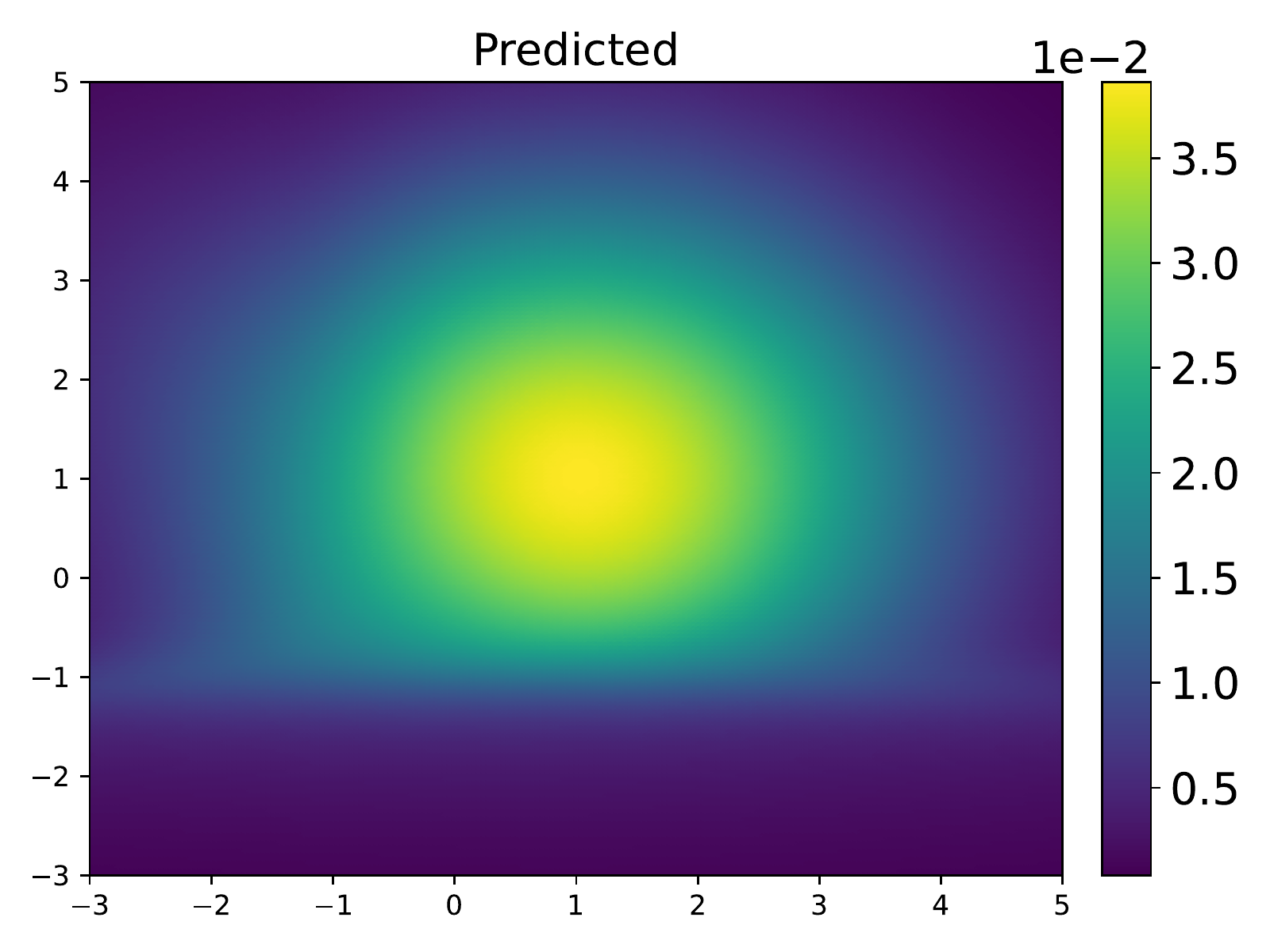}\vspace{4pt}
\includegraphics[height=3.5cm,width=4cm]{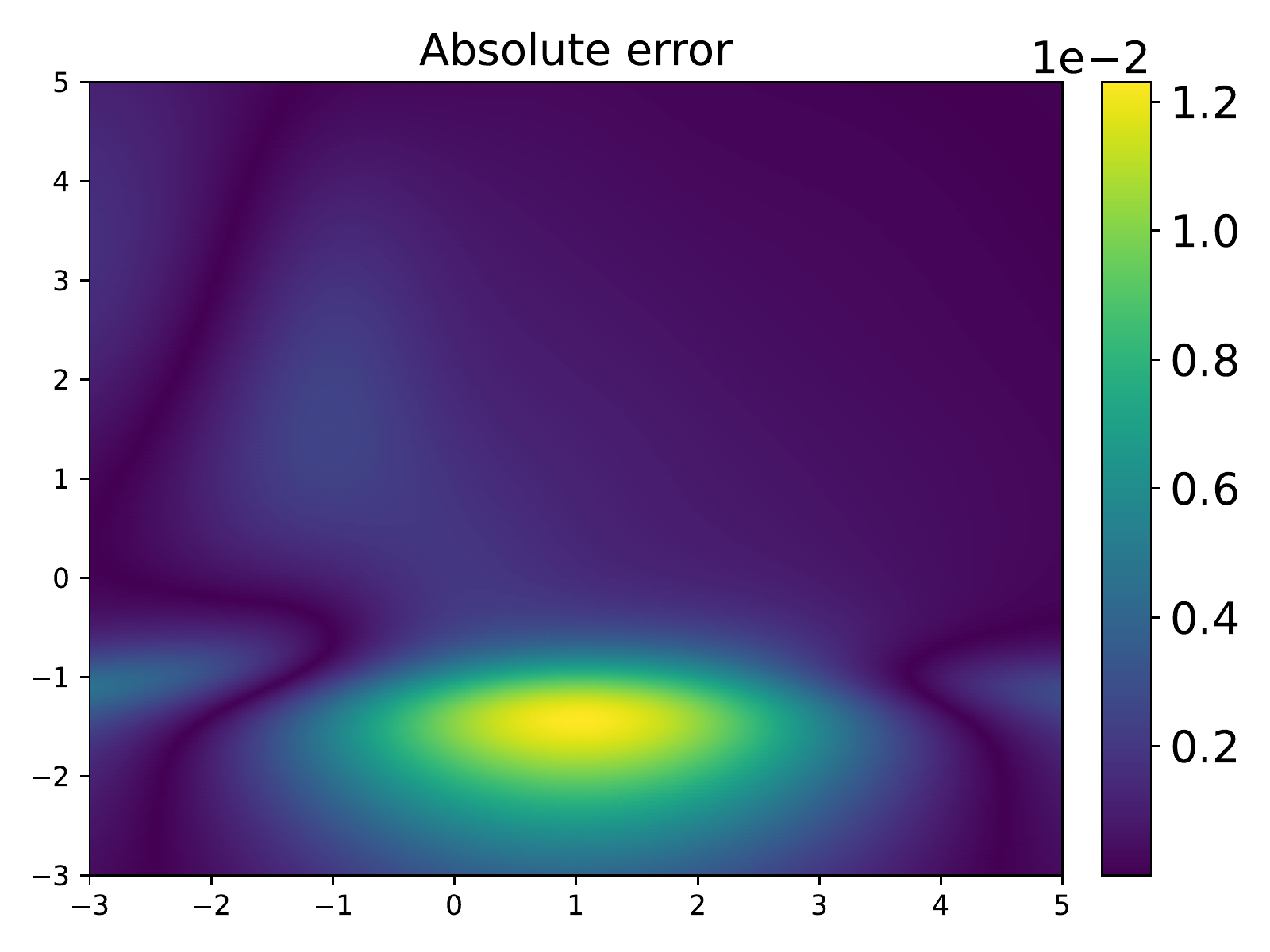}
\end{minipage}
}
	\caption{Comparison between the predicted solutions and the reference solutions. Top row: numerical solution. Bottom row: Absolute error between the numerical solution and the exact solution.}
	\label{gauss_frac_solu}
\end{figure}

\begin{figure}[h]
	\centering
	\subfigure{
		\includegraphics[height=3.5cm, width=5.5cm]{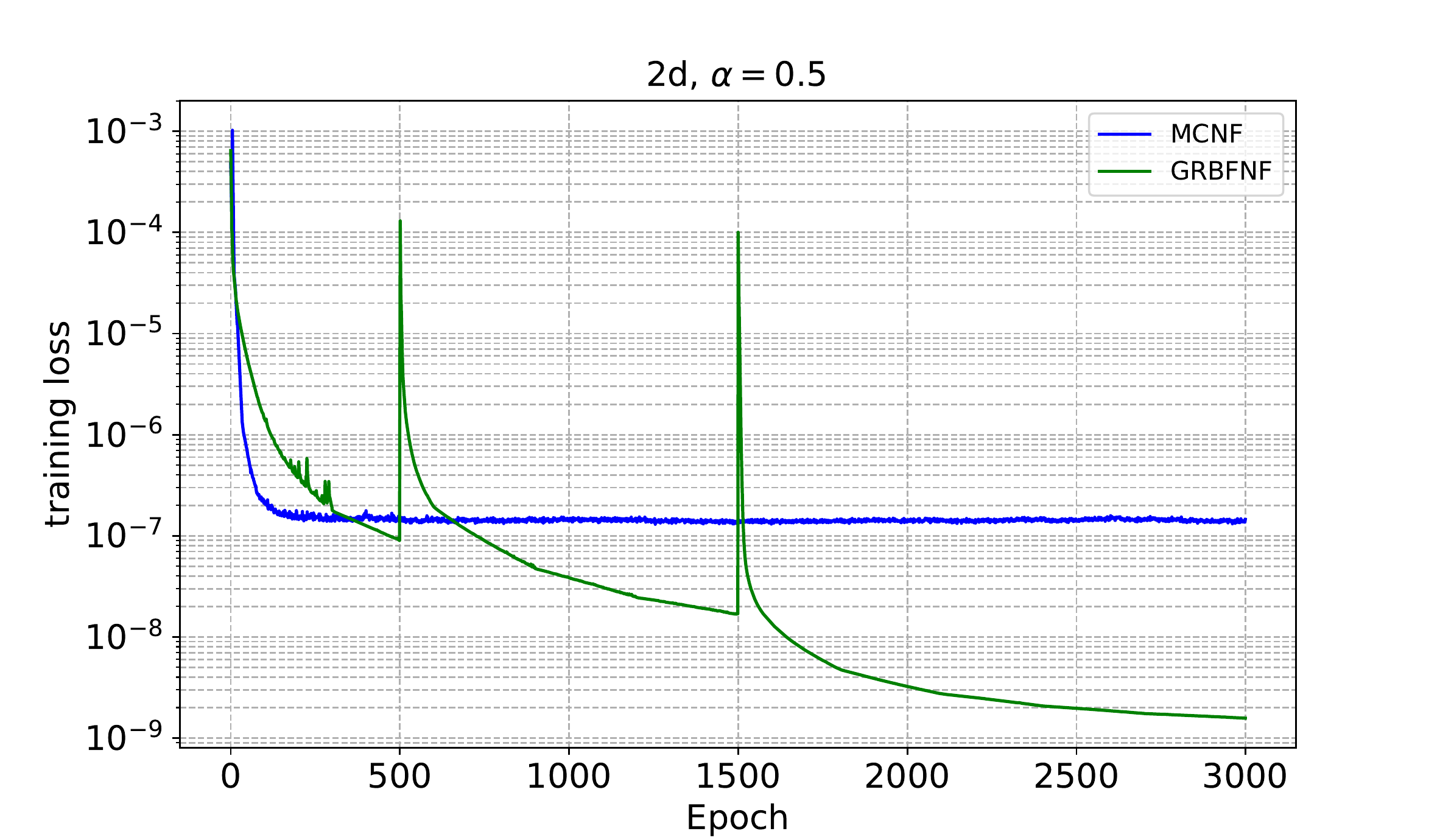}
		\includegraphics[height=3.5cm, width=5.5cm]{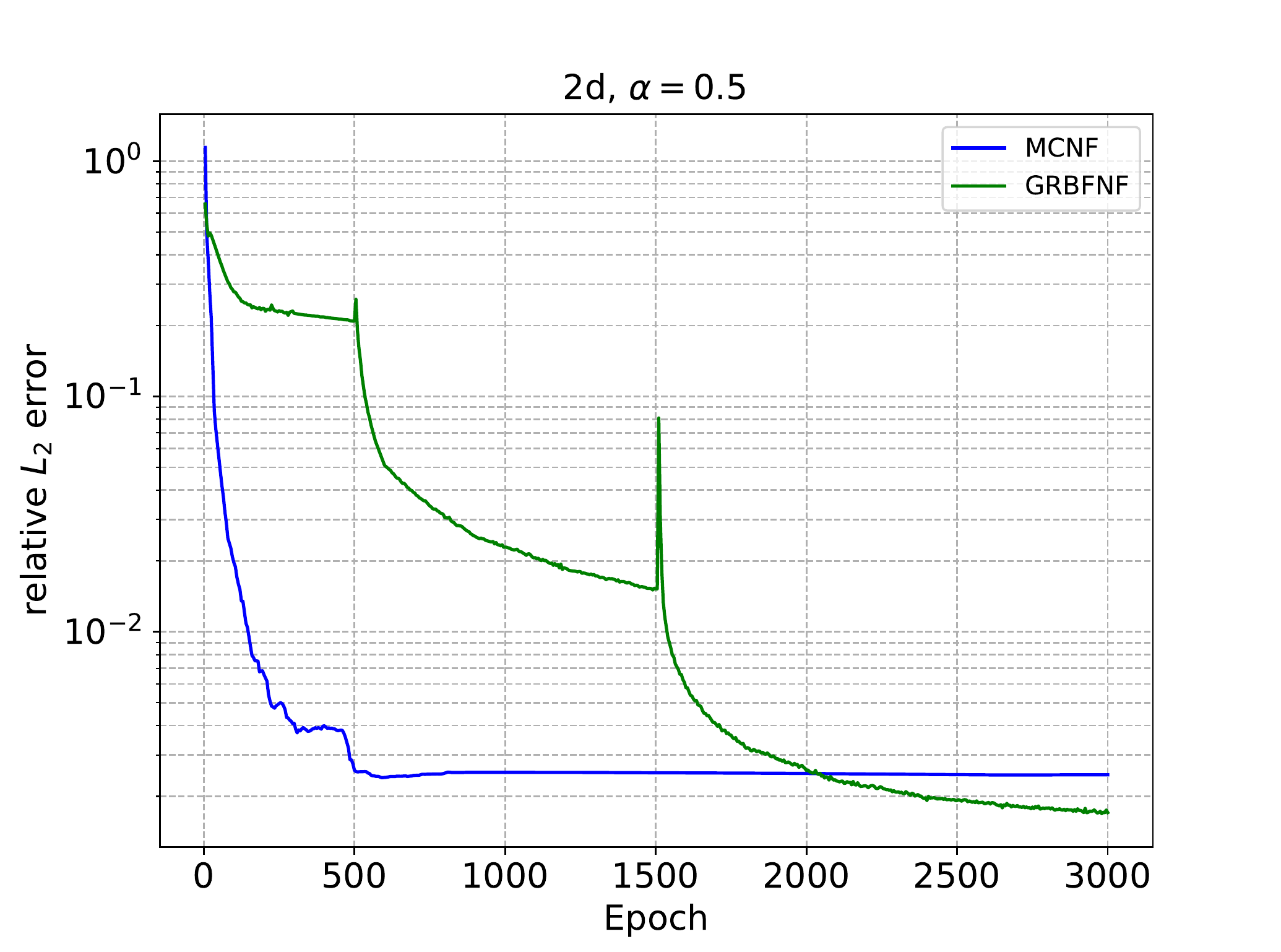}
		\includegraphics[height=3.5cm, width=5.5cm]{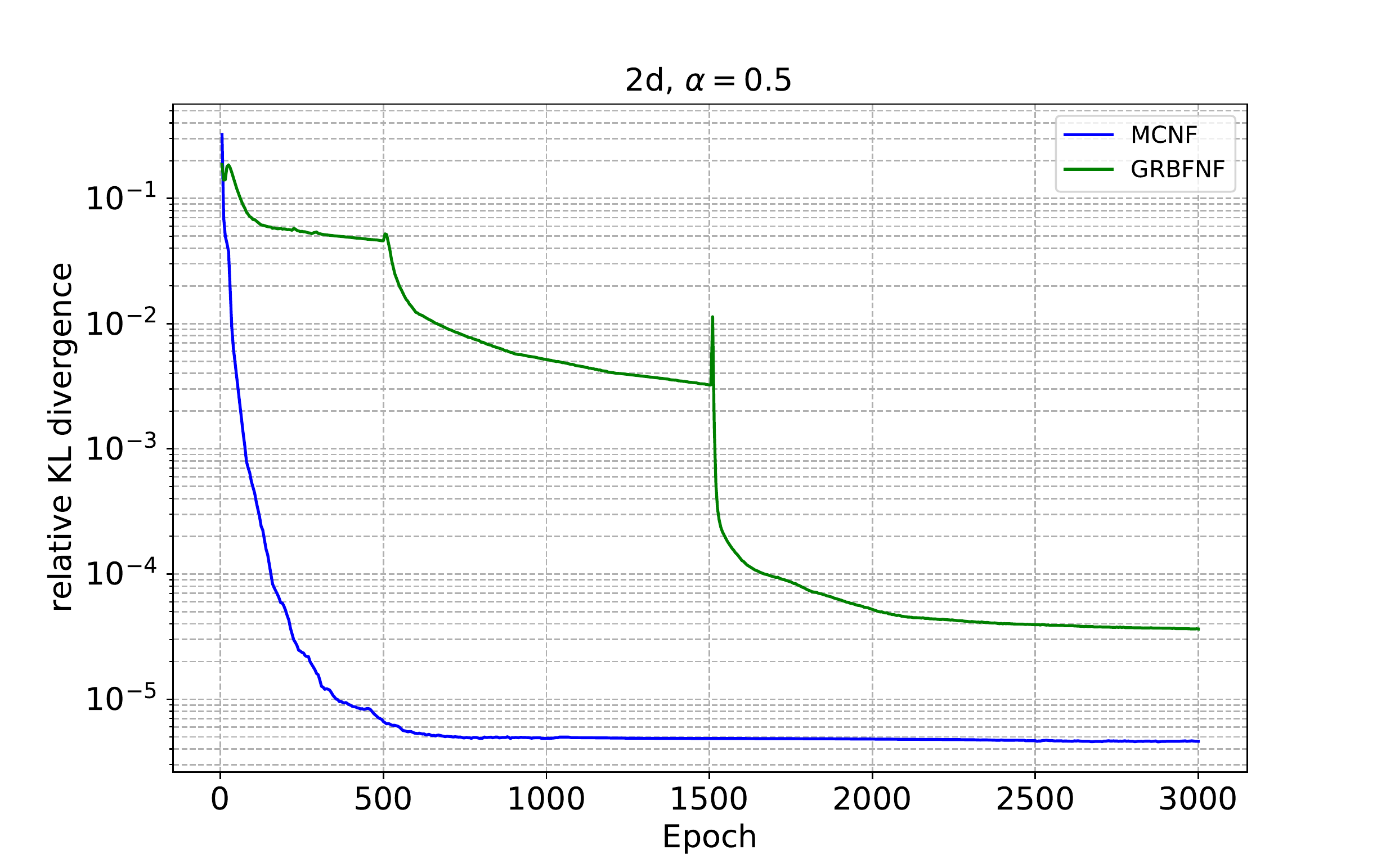}}
	\caption{Convergence behavior of MCNF and GRBFNF. Left: training loss. Middle: relative $L_2$ error. Right: relative KL divergence. }
	\label{RBFNF_MC_sta_frac}
\end{figure}

What's more, we compare our adaptive methods with non-adaptive methods in Fig. \ref{frac_MC_RBF_no}. It can be seen that, the accuracy of adaptive algorithm is higher than that of the non-adaptive algorithm especially for GRBFNF. The computational area of non-adaptive method is always $[0,6]^2$, which certainly affects the performance outside this area. That is to say, the numerical solution can approximate the ground truth well inside predetermined area while fail to capture the information outside this area especially when the prior knowledge is not enough to design a suitable computational area. We drawn the ground truth in Fig. \ref{fig:true_FFP_frac}. The comparison between the predicted solution and the exact solution are presented in Fig. \ref{gauss_frac_solu}, from where we can clearly observe that the non-adaptive methods show larger errors in the area outside the computational area $[0,6]^2$. On the other hand, our methods update the training points adaptively, which can effectively alleviate the limitation of a fixed computational area. Both the solutions of MCNF and GRBFNF yield excellent agreement with the exact solution.  The relative $L_2$ error and the relative KL divergence with different adaptivity iteration numbers are also provided in Fig.\ref{RBFNF_MC_sta_frac}. The relative $L_2$ error of GRBFNF is smaller than MCNF while the relative KL divergence of GRBFNF is larger than MCNF.


\begin{figure}[h]
	\centering
	\begin{minipage}[b]{0.43\linewidth}
		\includegraphics[height=5cm,width=7cm]{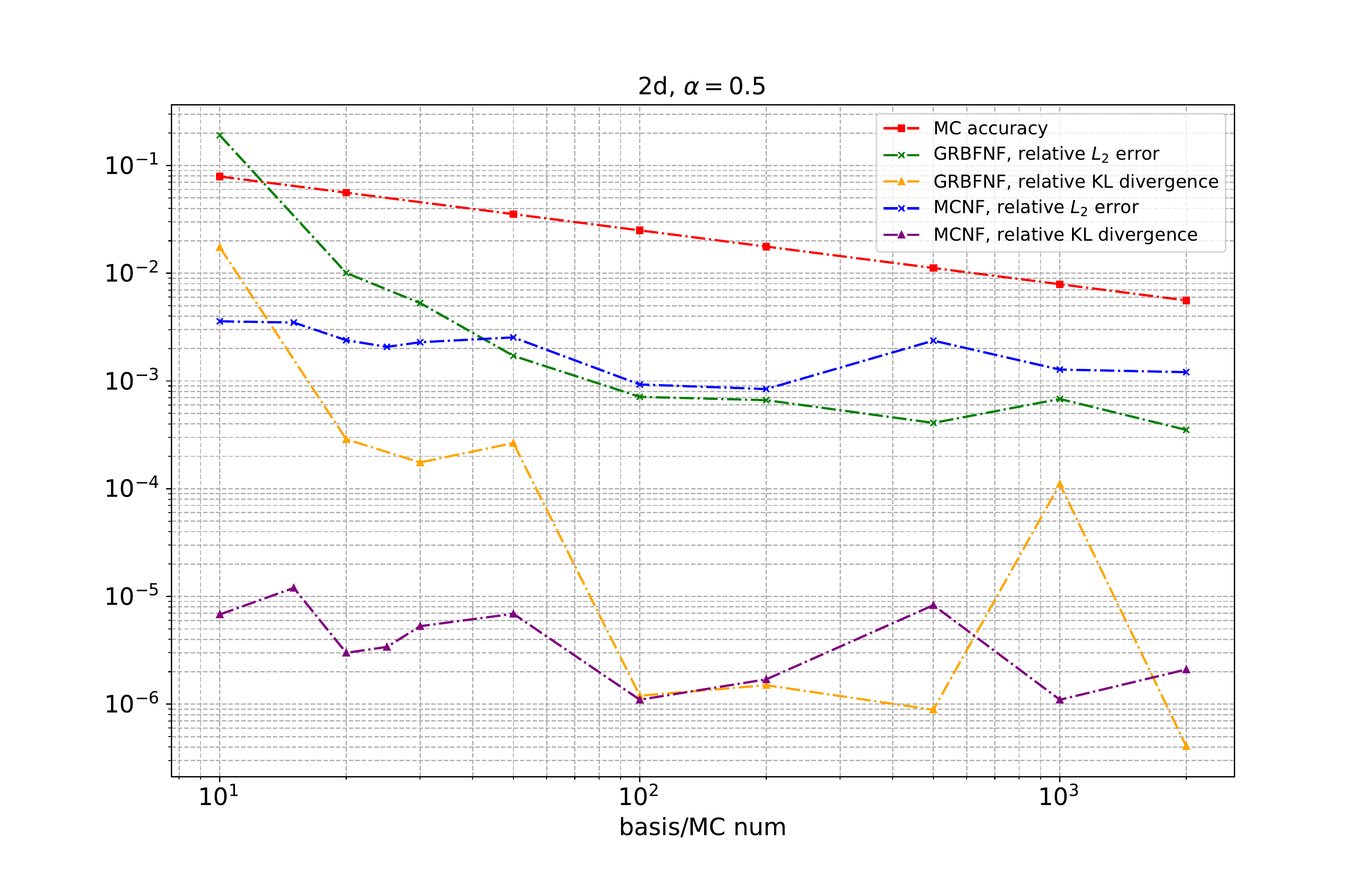}
	\end{minipage}
	\begin{minipage}[b]{0.43\linewidth}
		\includegraphics[height=5cm,width=7cm]{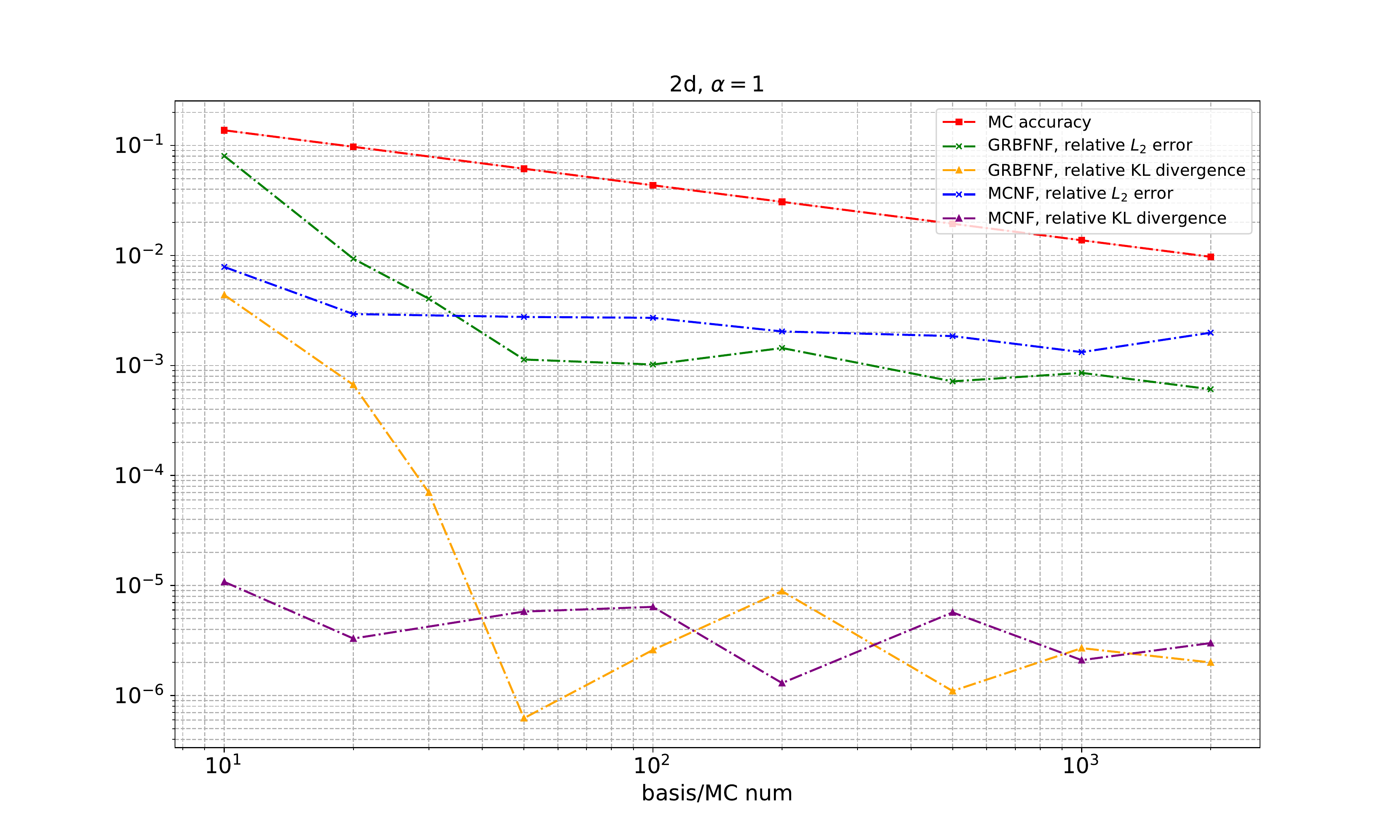}
	\end{minipage}
	
	\begin{minipage}[b]{0.43\linewidth}
		\includegraphics[height=5cm,width=7cm]{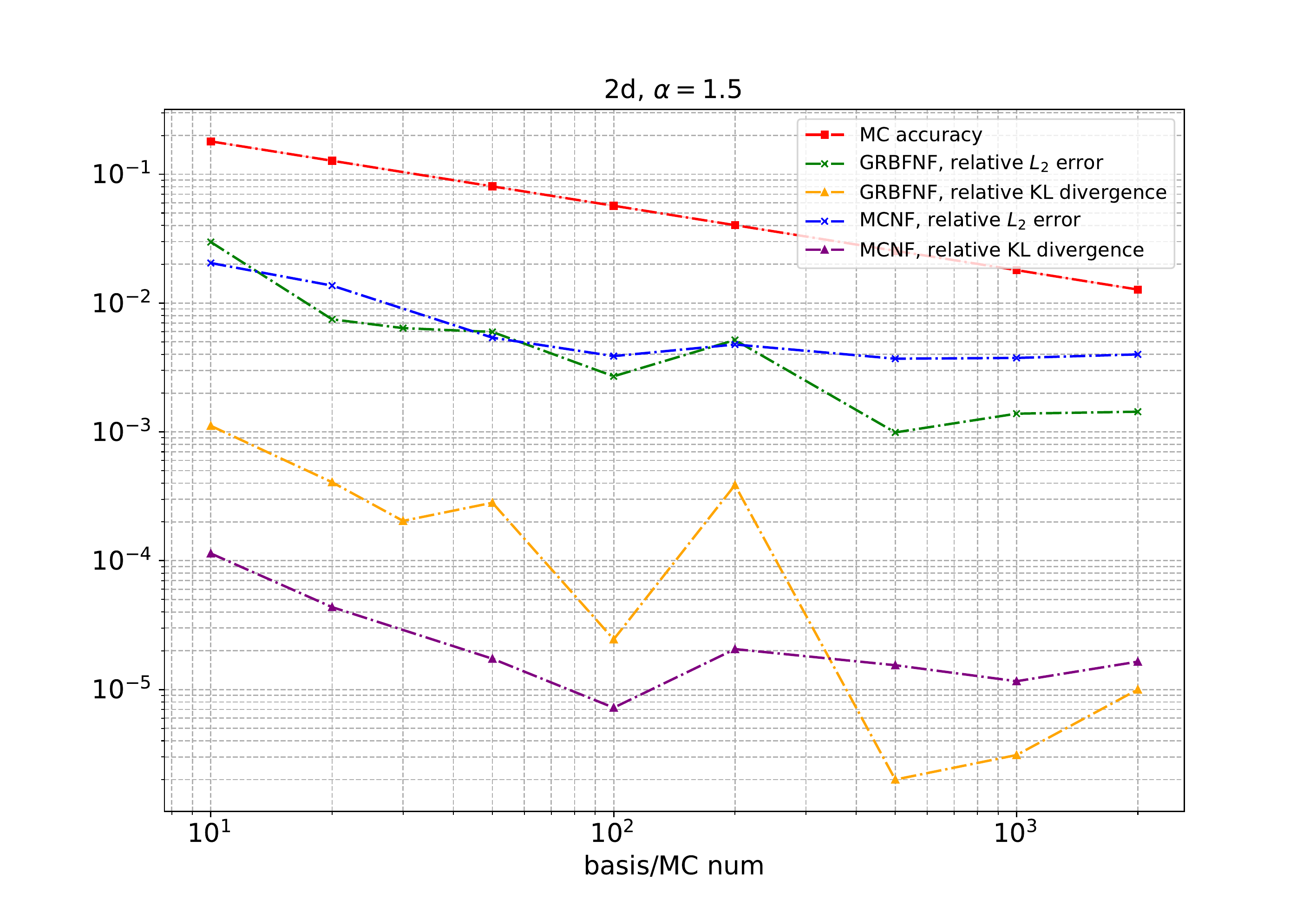}
	\end{minipage}
	\begin{minipage}[b]{0.43\linewidth}
		\includegraphics[height=5cm,width=7cm]{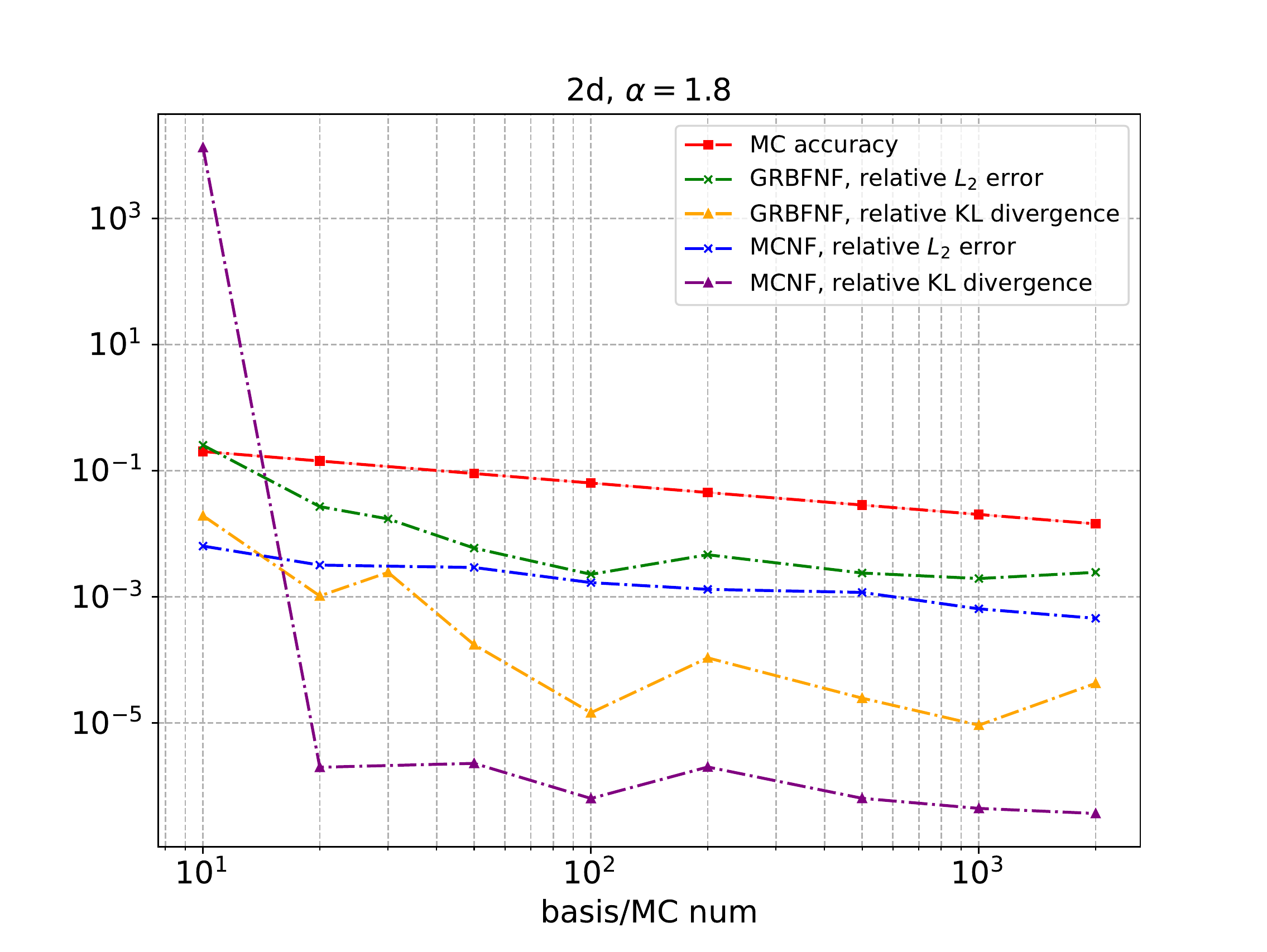}
	\end{minipage}
	\caption{Error decay of MCNF and GRBFNF for different $\alpha$ in terms of the number of MC samples and GRBF basis functions. Left: relative $L_2$ error. Right: relative KL divergence. }
	\label{RBFNF_MC_sta_frac_alpha}
\end{figure}

Finally, we take $[-3,3]^2$ to replace the above initial sampling area $[0,6]^2$ and  repeat the experiments to test the performance of MCNF and GRBFNF for fractional FPEs with different fractional order $\alpha$. The results are presented in Fig. \ref{RBFNF_MC_sta_frac_alpha}, where we also display the accuracy of Monte Carlo sampling method to compute associated fractional Laplacian. The numerical error of approximating fractional Laplacian is defined by
$\frac{\sum_i|(-\Delta)^{\frac{\alpha}{2}}[p](\bm{x}_i)-\mathcal{M}[p](\bm{x}_i)|^2}{\sum_i|(-\Delta)^{\frac{\alpha}{2}}[p](\bm{x}_i)|^2}$
where $\mathcal{M}[p]$ denotes numerical approximation. Both MCNF and GRBFNF arrive good agreement with the ground truth for $\alpha=0.5, 1, 1.5, 1.8$.

\subsection{Bimodal distribution}
To test the performance of {MCNF} and {GRBFNF} with respect to a bimodal distribution, we consider
\begin{equation}
\left\{
\begin{aligned}
&\nabla\cdot(\bm{g}(\bm{x})p(\bm{x}))+0.05\Delta p(\bm{x})-(-\Delta)^{\alpha/2}p(\bm{x})=f(\bm{x}), \quad \bm{x}\in\mathbb{R}^2,\\
&\int_{\mathbb{R}^2}p(\bm{x}){\rm d}\bm{x}=1, \quad p(\bm{x})\geq0,
\end{aligned}\right.
\end{equation}
where $\bm{g}(x)=0.2\bm{x}$, $$f(\bm{x})=\frac{1}{5\pi}\nabla\cdot\big(\exp(-2\|\bm{x}-\bm{1}_2\|_2^2)\big)-\frac{1}{\pi}B(2,\alpha)2^{\frac{\alpha}{2}}\left({_1F_1}\bigg(\frac{2+\alpha}{2}; 1; -2\|\bm{x}\|^2_2\bigg)+{_1F_1}\bigg(\frac{2+\alpha}{2}; 1; -2\|\bm{x}-\bm{1}_2\|^2_2\bigg)\right).$$
The true solution is $$p(\bm{x})=\frac{1}{\pi}\bigg(\exp\Big(-2\|\bm{x}\|^2_2\Big)+\exp\Big(-2\|\bm{x}-\bm{1}_2\|^2_2\Big)\bigg).$$

For the NF, we take $L=8$ affine coupling layers with $32$ hidden neurons. The initial training set is generated via uniformly distributed points in $[-3, 3]^2$. The sample size is $5000$ and the batch size is set to be $1024$. Both {MCNF} and {GRBFNF} are applied.
For the {MCNF}, the number of the samples used to approximate fractional Laplacian is $100$, $r_0 = 0.3$, $r_{\epsilon} = 0.0001$. $800$ adaptivity iterations with $5$ epochs for each adaptivity iteration are conducted. The initial learning rate is $0.001$ with $80\%$ decay each $3000$ steps. For the {GRBFNF}, the number of the basis functions is $100$ and the initial center points of basis functions are generated from a uniform distribution in area $[-3, 3]$. $3$ adaptivity iterations with increasing epochs are conducted for this problem, i.e. 500 epochs for the first adaptivity iteration, 1000 epochs for the second adaptivity iteration, and 2500 epochs for the last adaptive iteration. The learning rate is $0.01$ with half decay every $300$ steps and is reset to $0.005$ after each adaptivity step.

\begin{figure}[h]
	\centering
	\subfigure[MCNF, $\alpha=0.5$.]{
		\includegraphics[height=3.5cm, width=5.5cm]{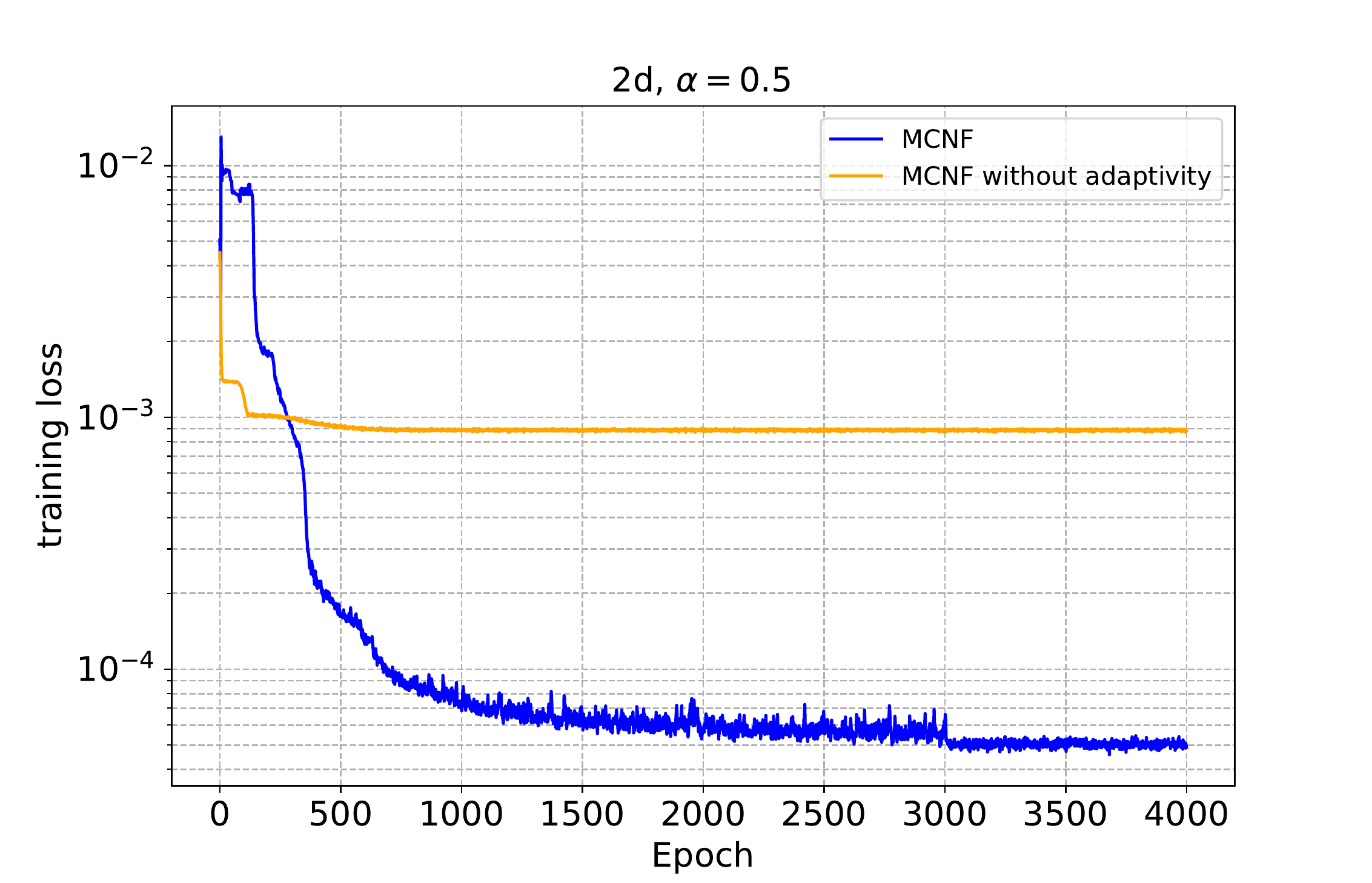}
		\includegraphics[height=3.5cm, width=5.5cm]{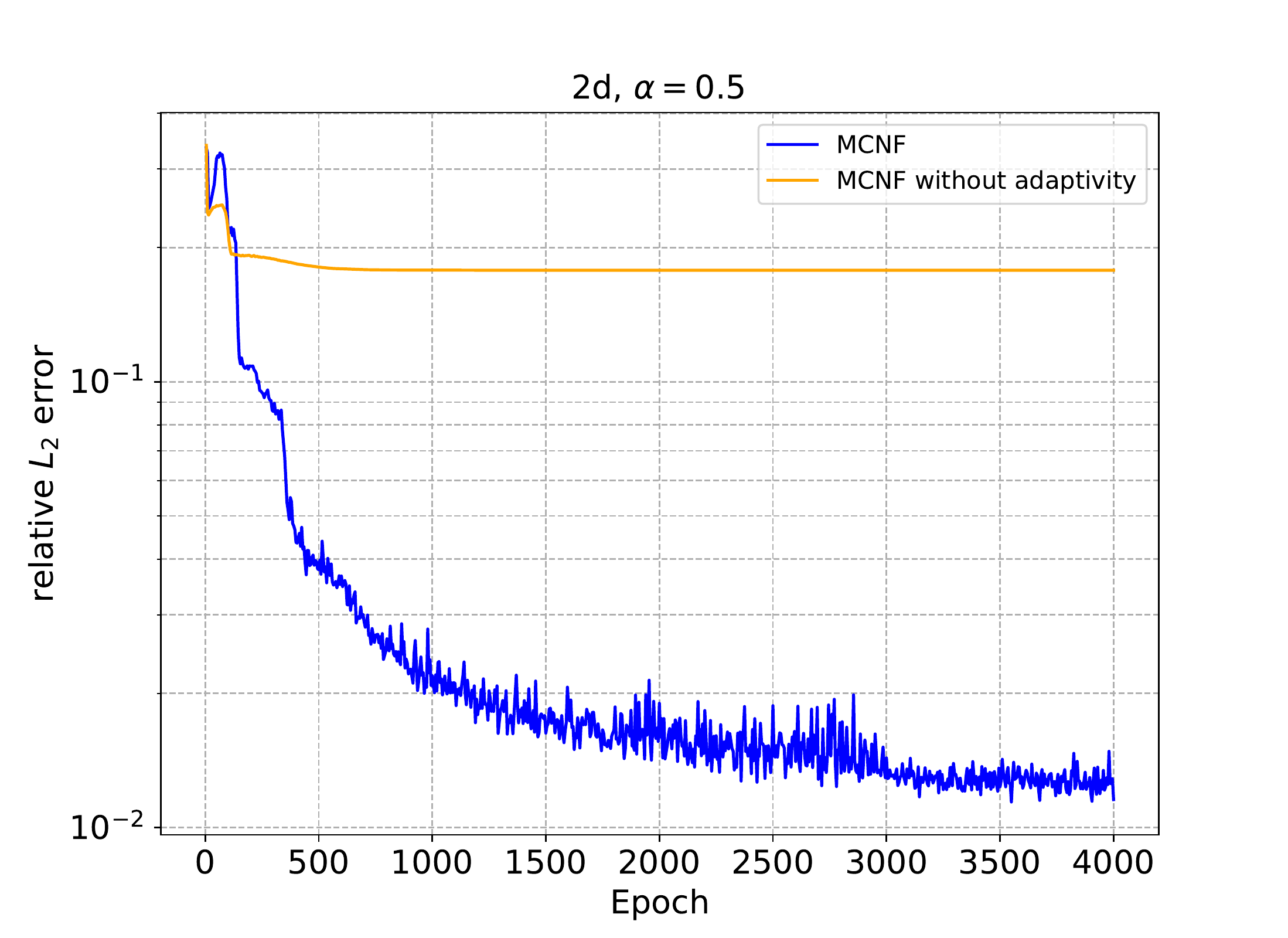}
		\includegraphics[height=3.5cm, width=5.5cm]{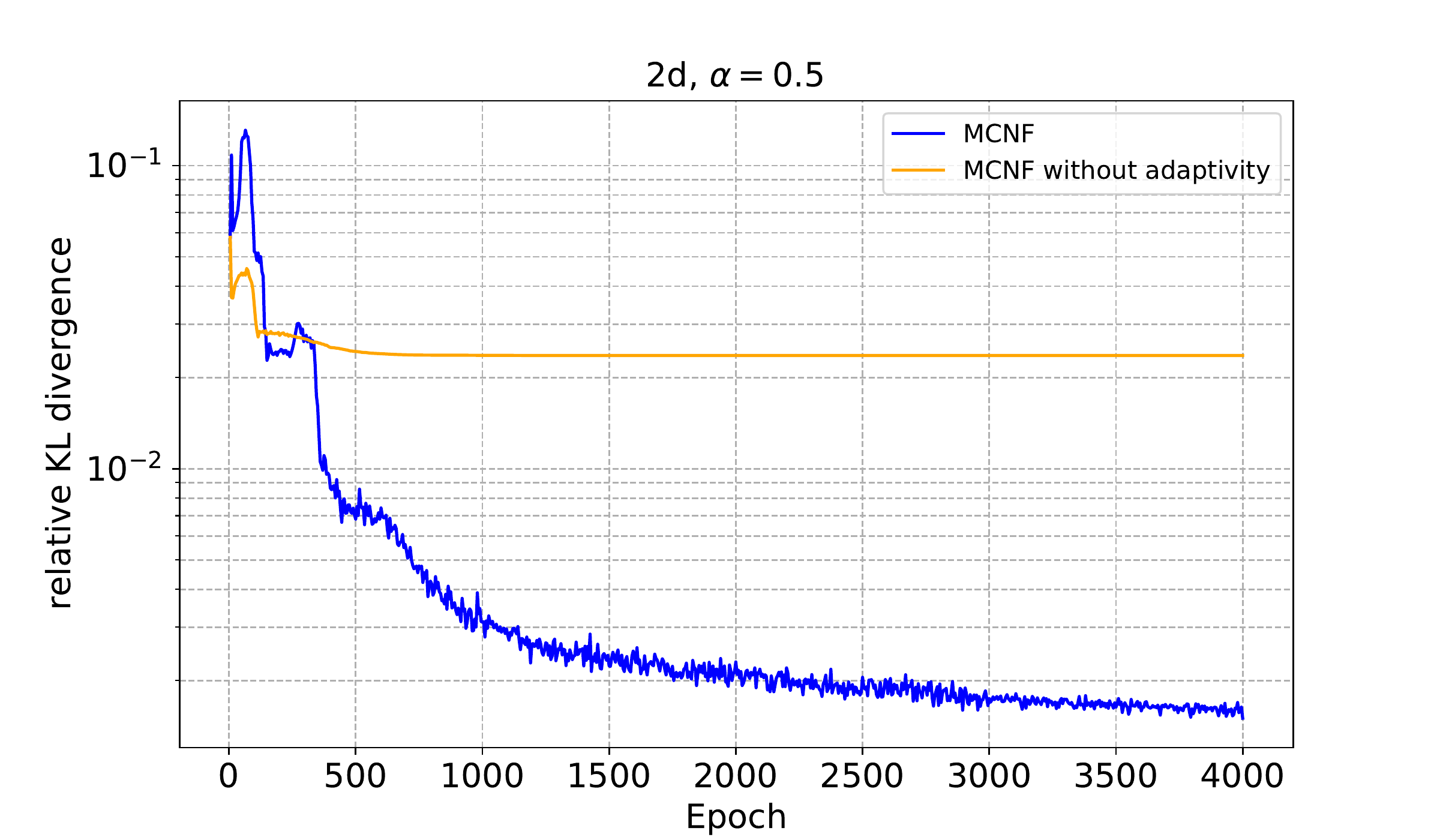}}
	\subfigure[GRBFNF, $\alpha=0.5$]{
		\includegraphics[height=3.5cm, width=5.5cm]{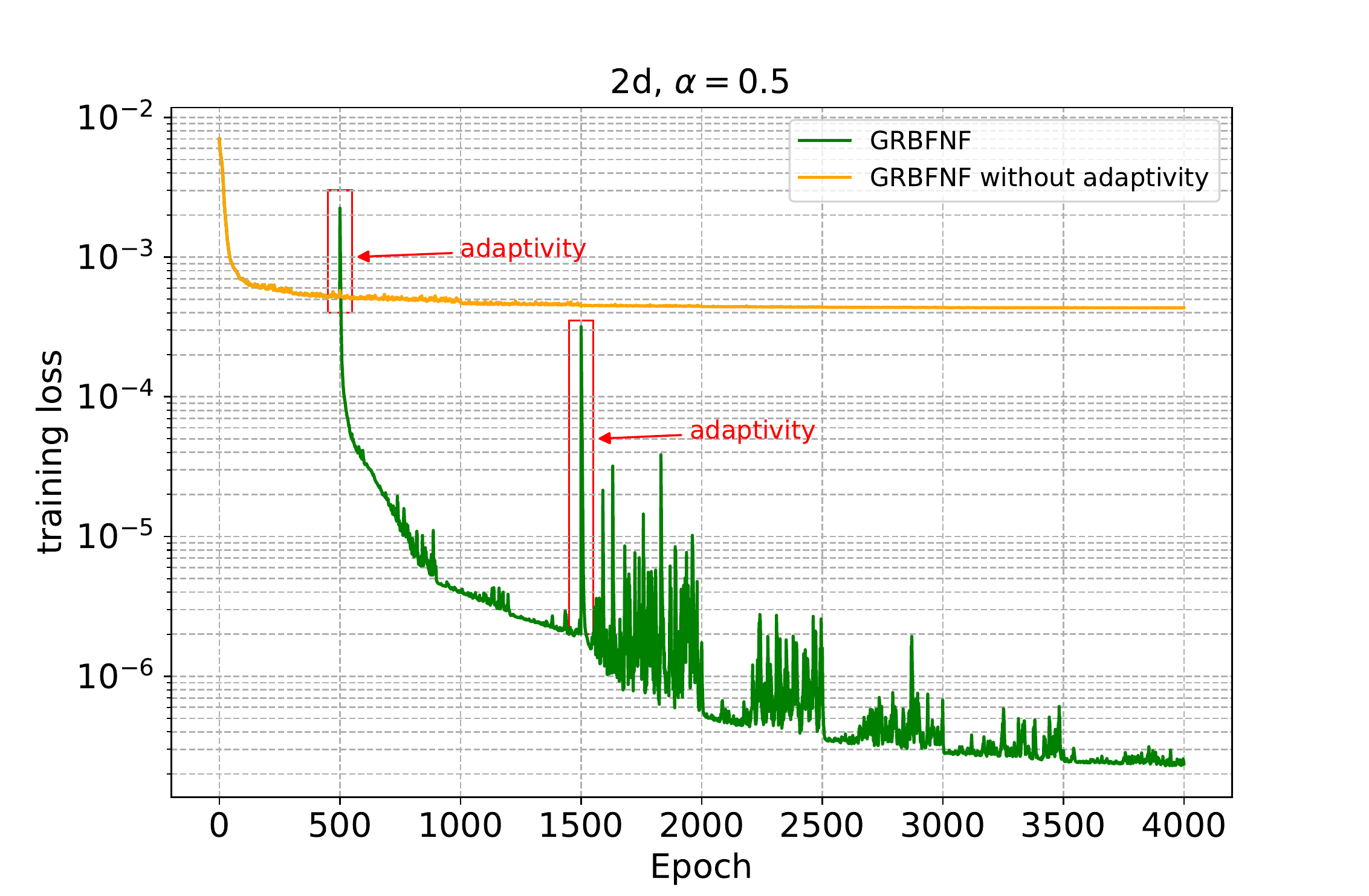}
		\includegraphics[height=3.5cm, width=5.5cm]{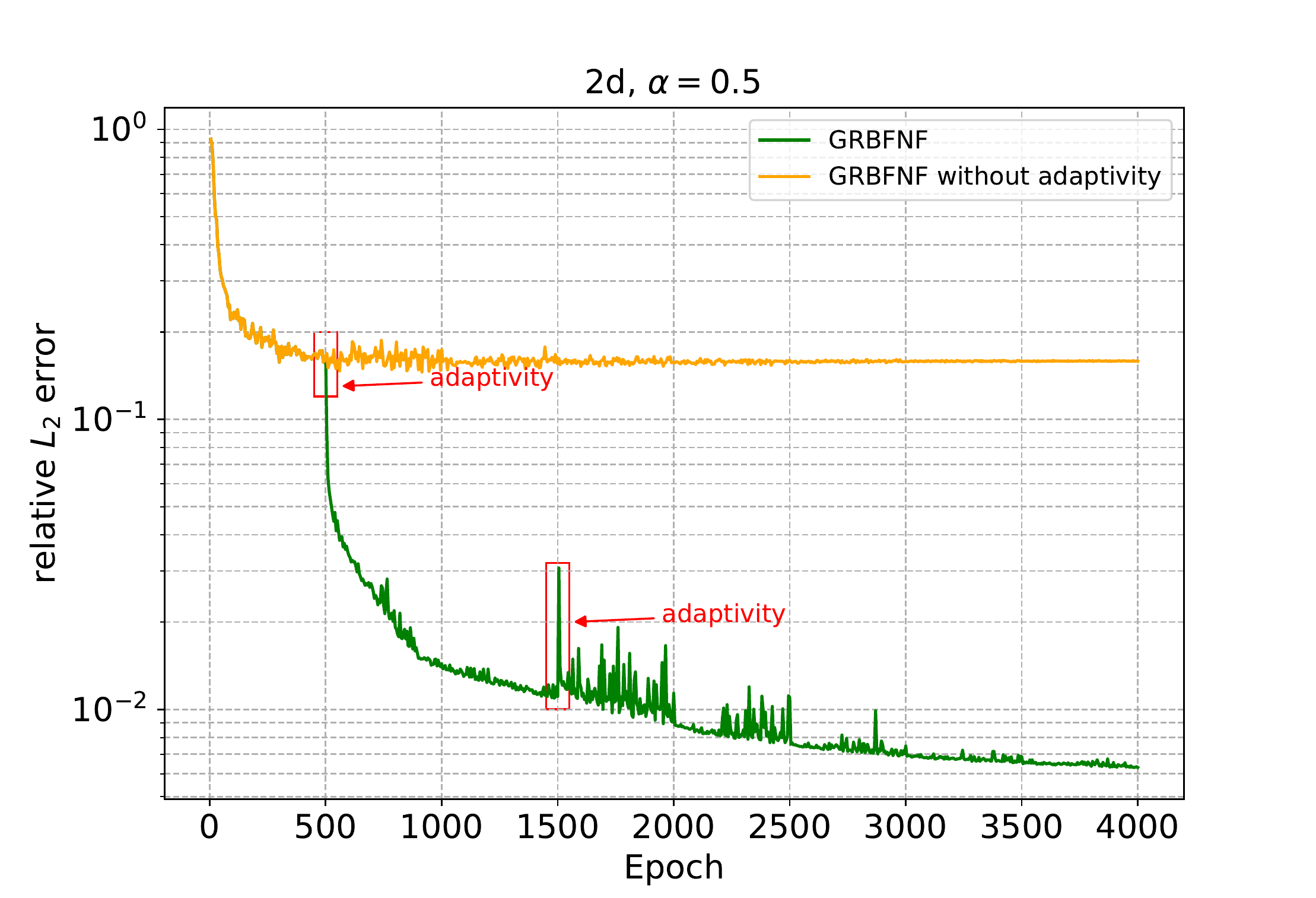}
		\includegraphics[height=3.5cm, width=5.5cm]{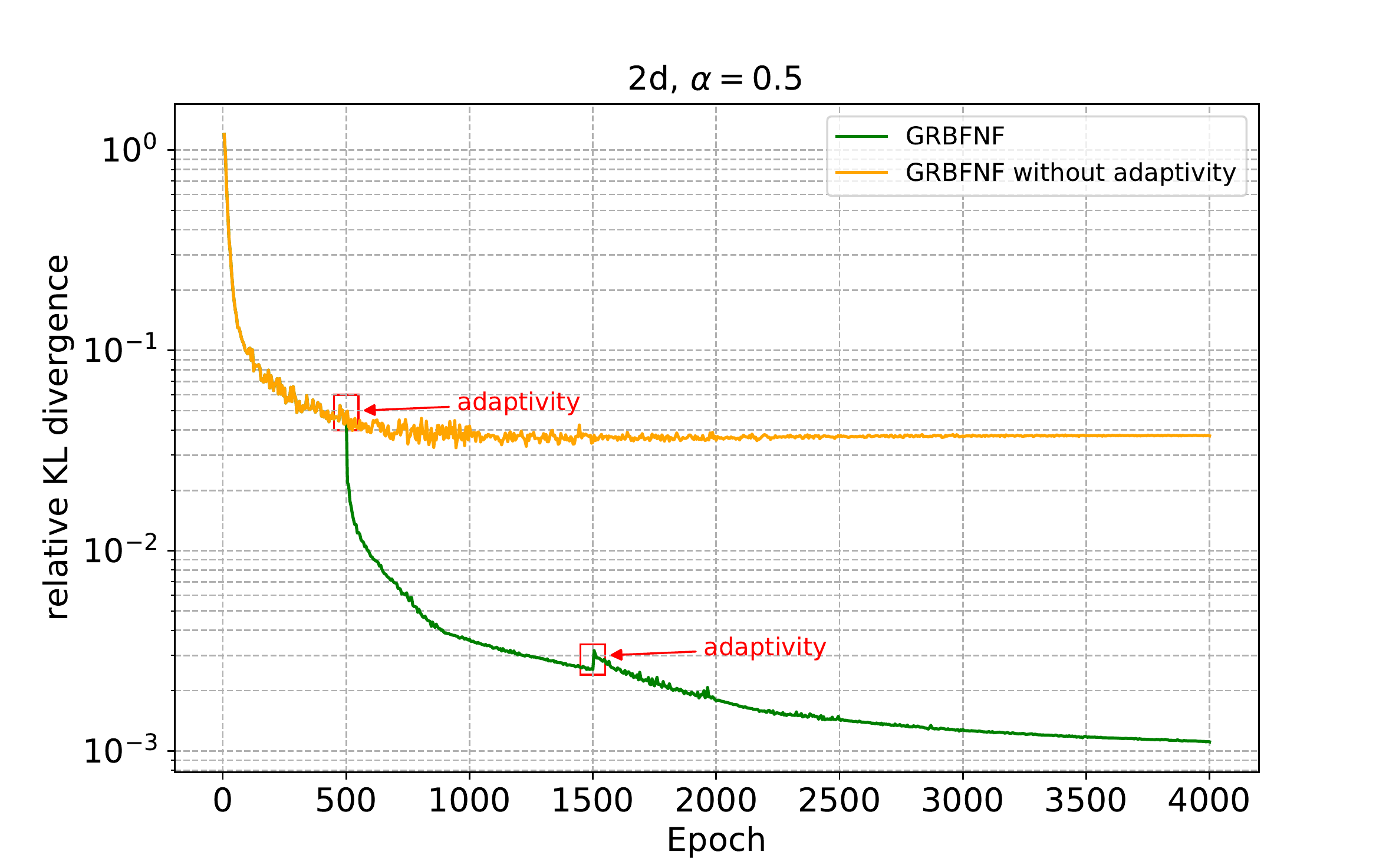}}
	\caption{Convergence behavior of MCNF and GRBFNF with and without adaptive sampling. Left: training loss. Middle: relative $L_2$ error. Right: relative KL divergence. }
	\label{sta_ga2d2_MC_RBF_no}
\end{figure}

\begin{figure}
	\centering
	\includegraphics[height=4cm,width=5cm]{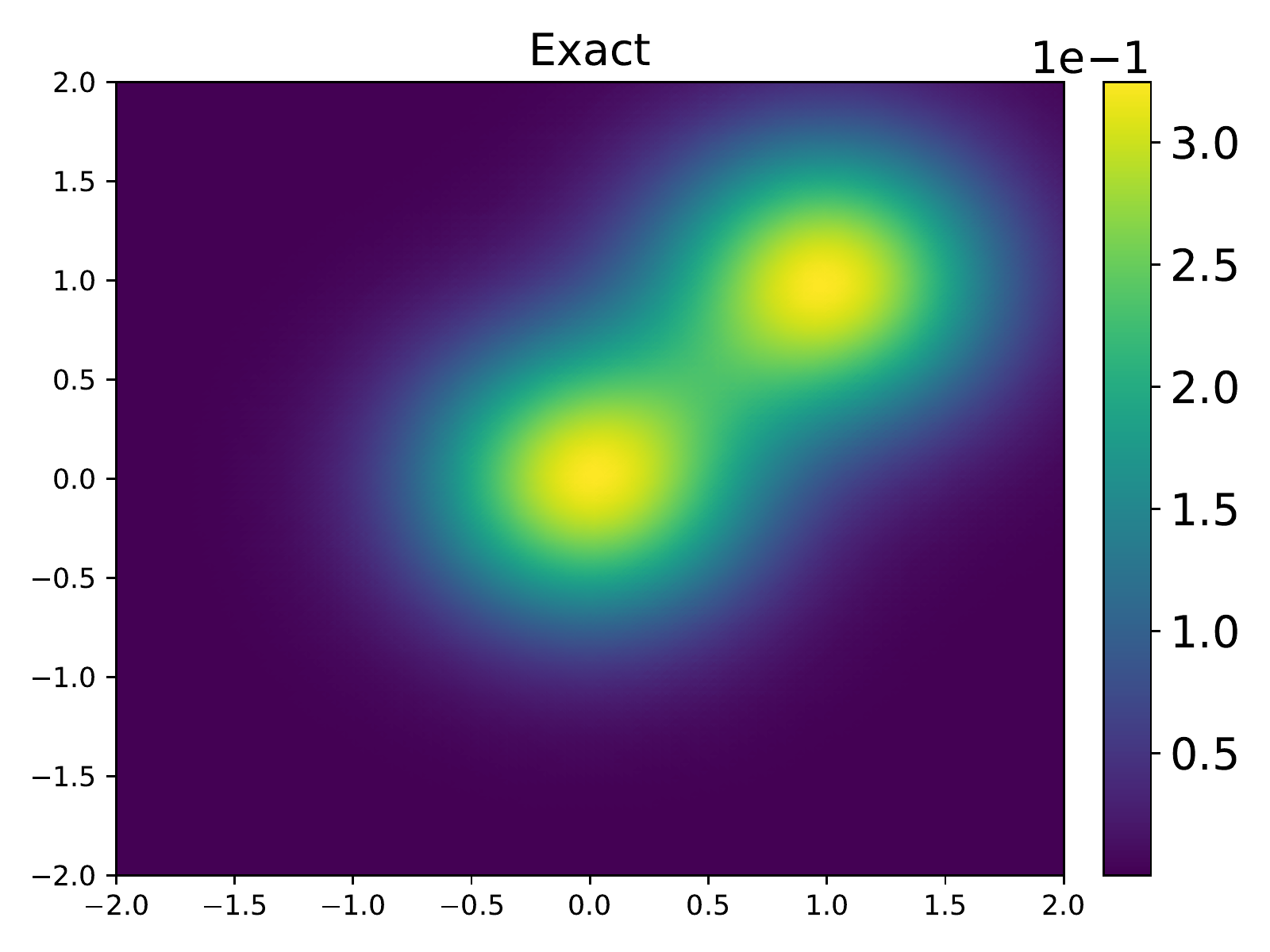}
	\caption{The ground truth of bimodal distribution.}
	\label{fig:true_FFP_bim}
\end{figure}

\begin{figure}[H]
	\centering
	\subfigure[{MCNF}]{
		\begin{minipage}[b]{0.23\linewidth}
			\includegraphics[height=3.5cm,width=4cm]{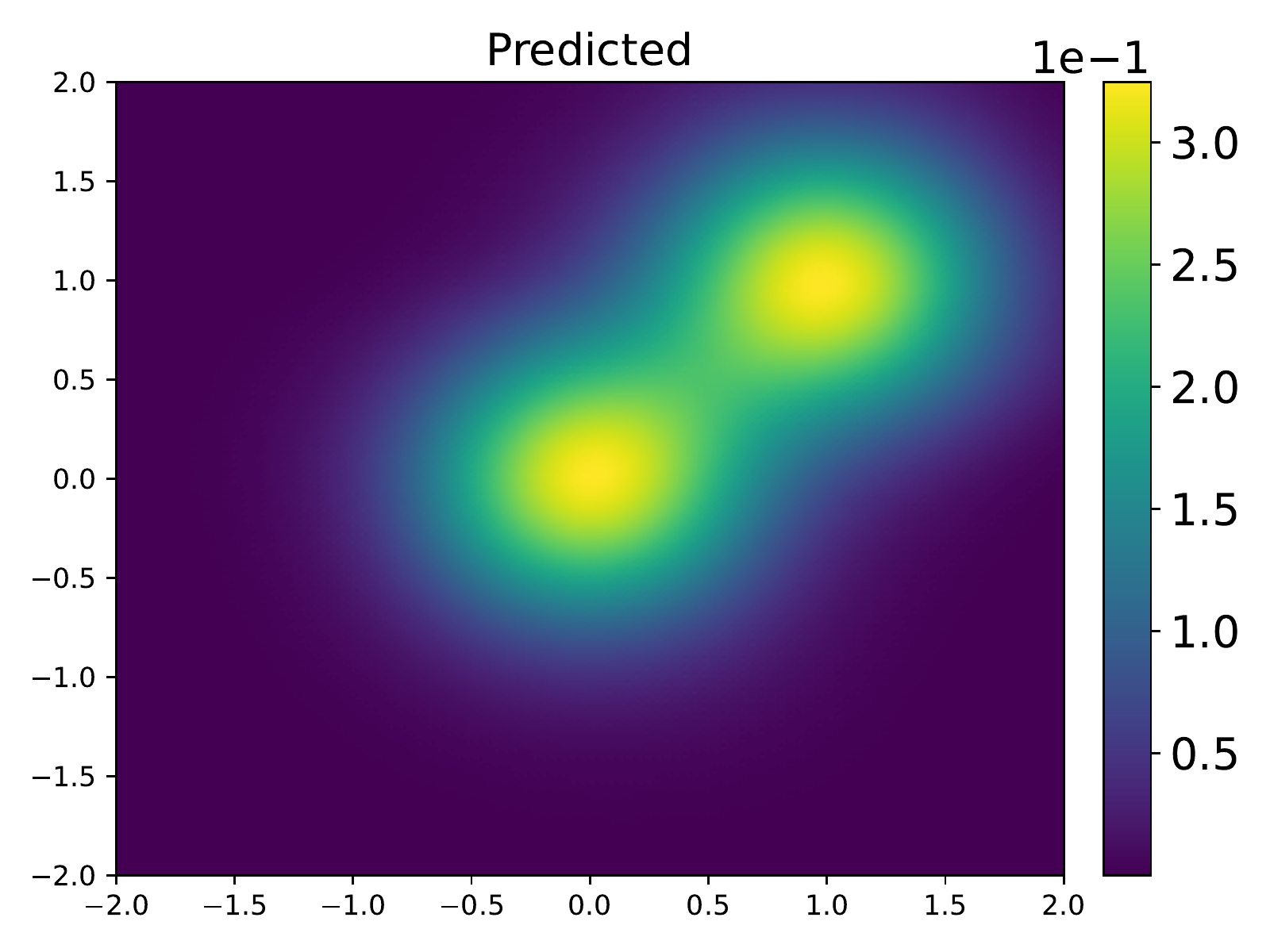}\vspace{4pt}
			\includegraphics[height=3.5cm,width=4cm]{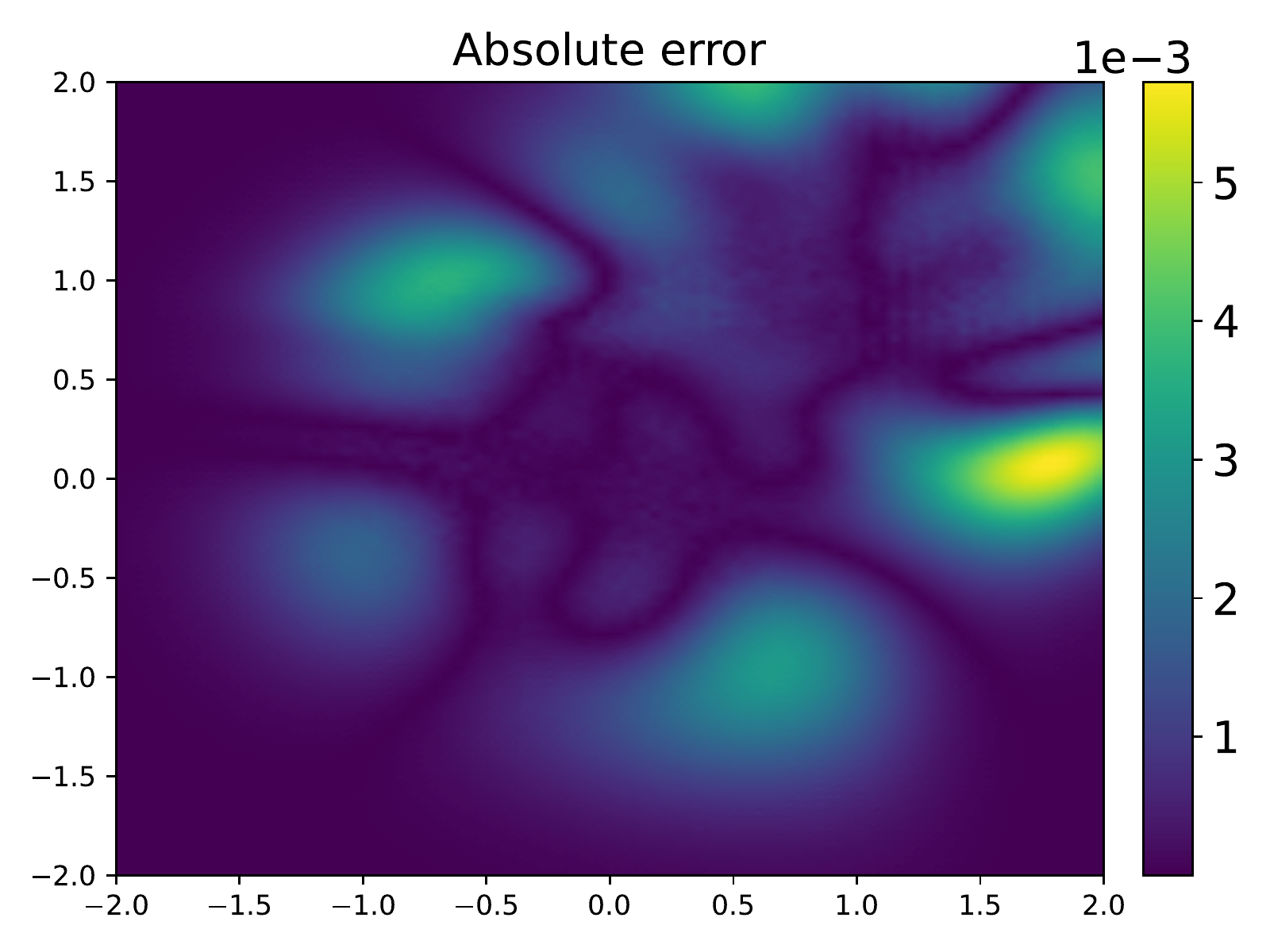}
	\end{minipage}}
	\subfigure[{MCNF} without adaptivity]{
		\begin{minipage}[b]{0.23\linewidth}
			\includegraphics[height=3.5cm,width=4cm]{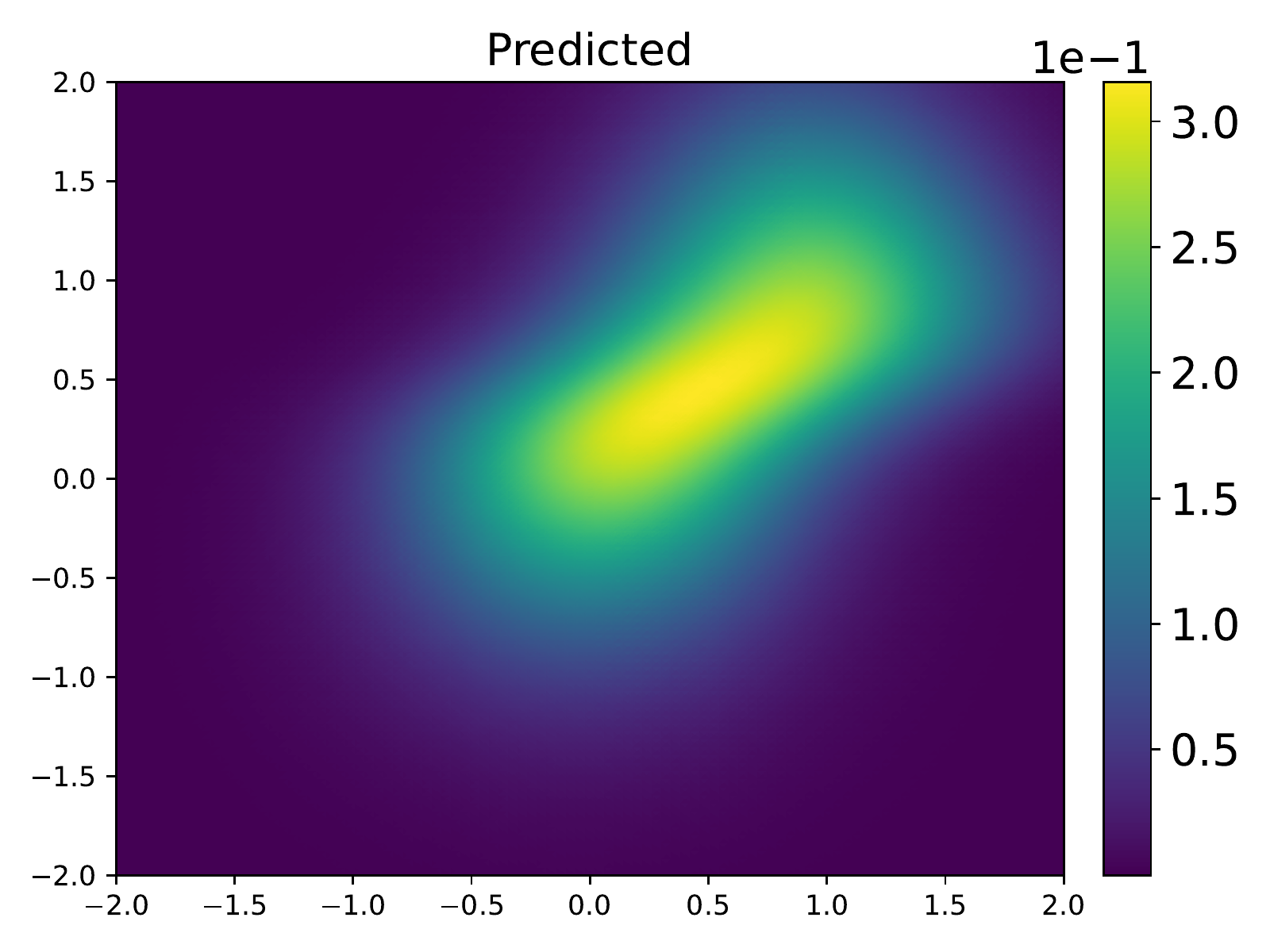}
			\quad
			\includegraphics[height=3.5cm,width=4cm]{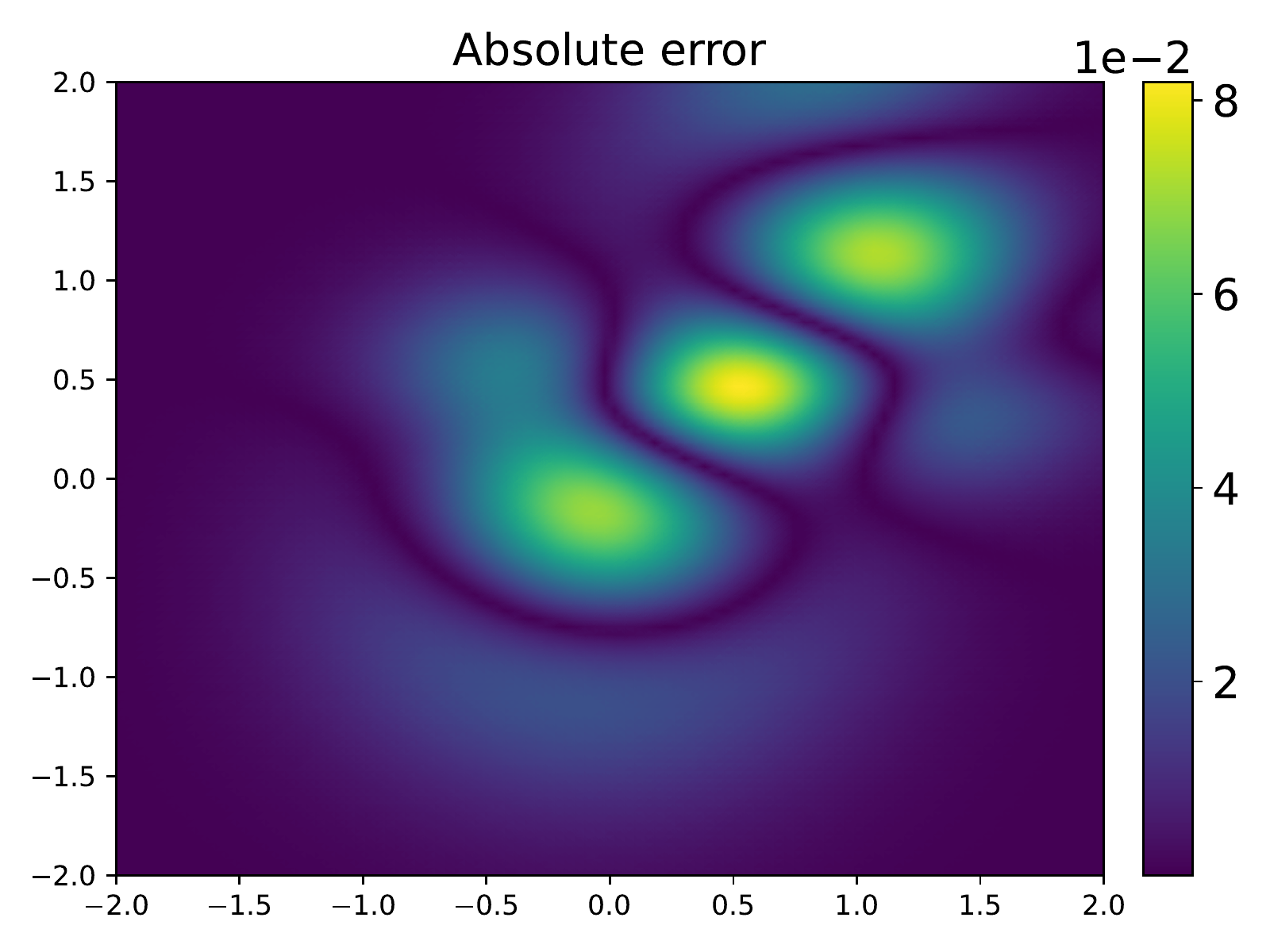}
	\end{minipage}}
	\subfigure[{GRBFNF}]{
		\begin{minipage}[b]{0.23\linewidth}
			\quad
			\includegraphics[height=3.5cm,width=4cm]{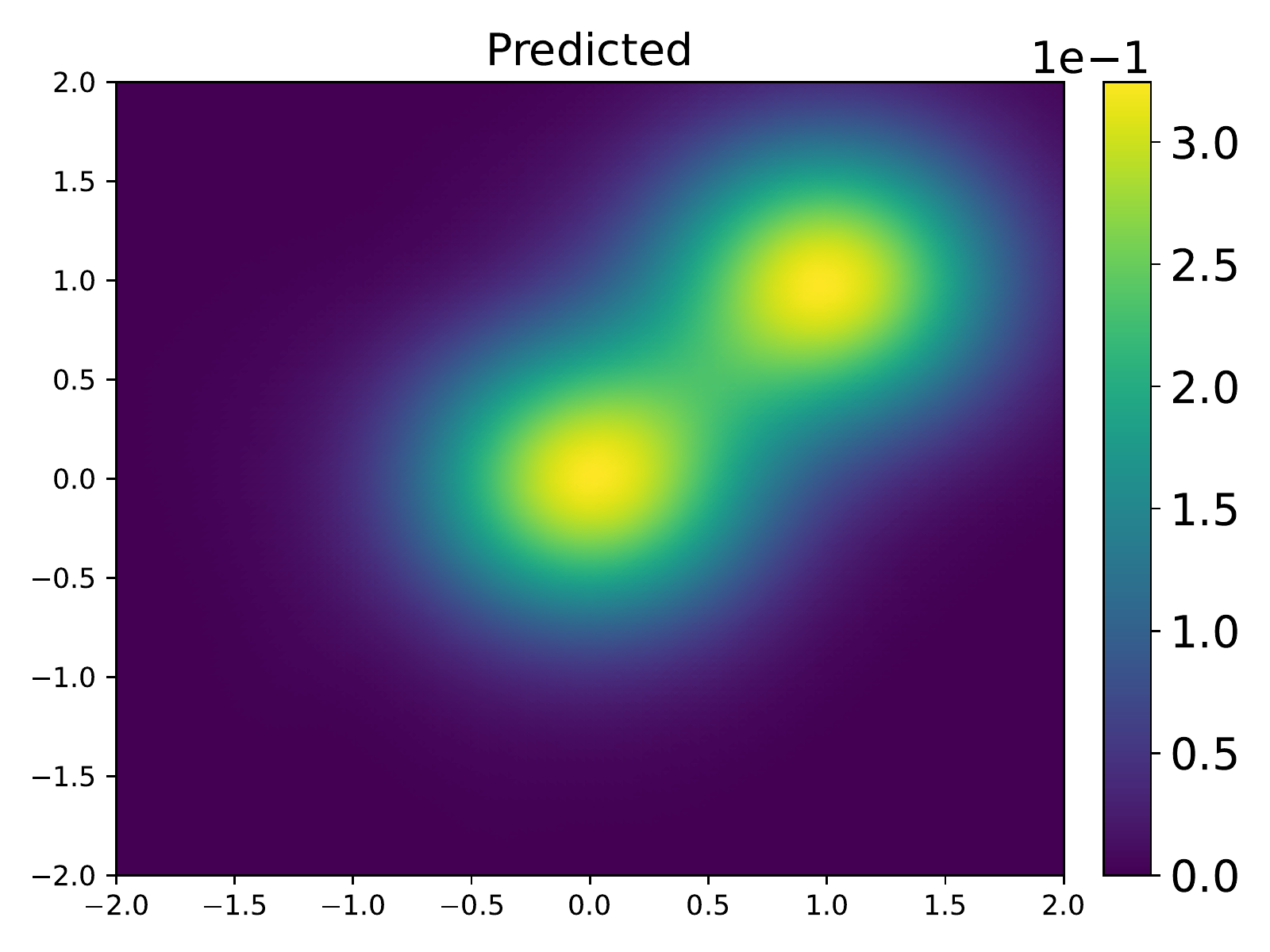}
			\quad
			\includegraphics[height=3.5cm,width=4cm]{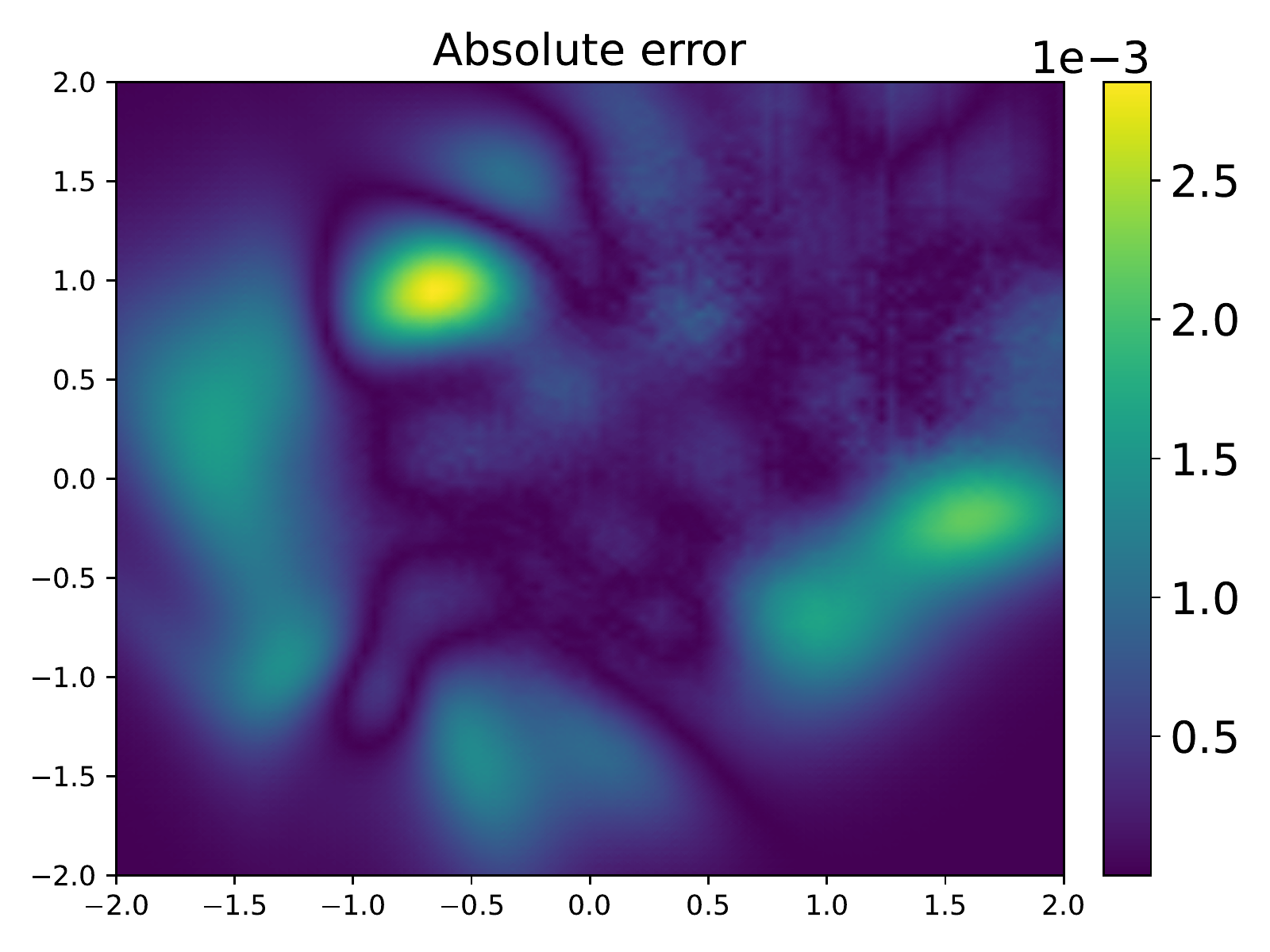}
	\end{minipage}}
	\subfigure[{GRBFNF} without adaptivity]{
		\begin{minipage}[b]{0.24\linewidth}
			\includegraphics[height=3.5cm,width=4cm]{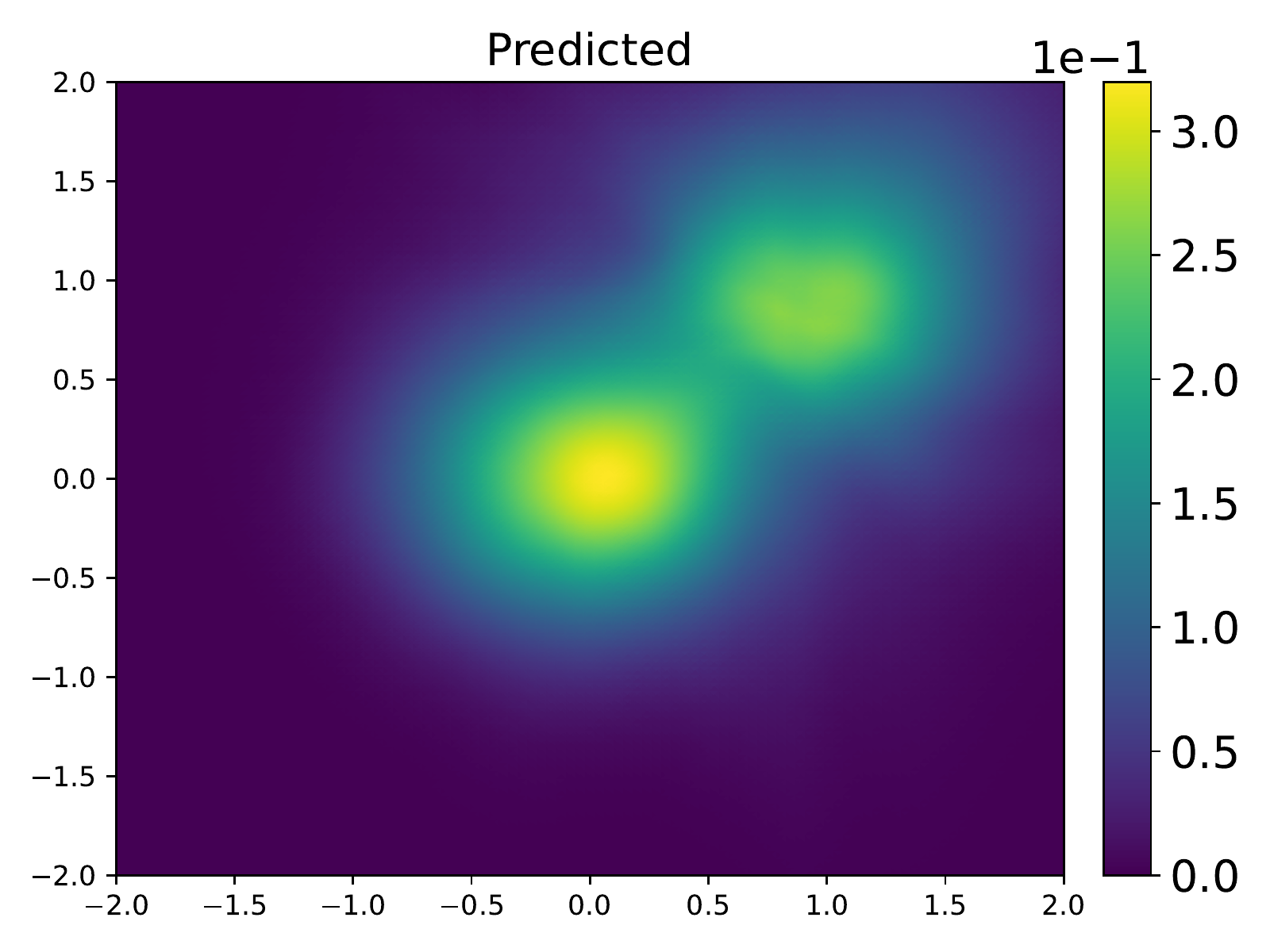}
			\includegraphics[height=3.5cm,width=4cm]{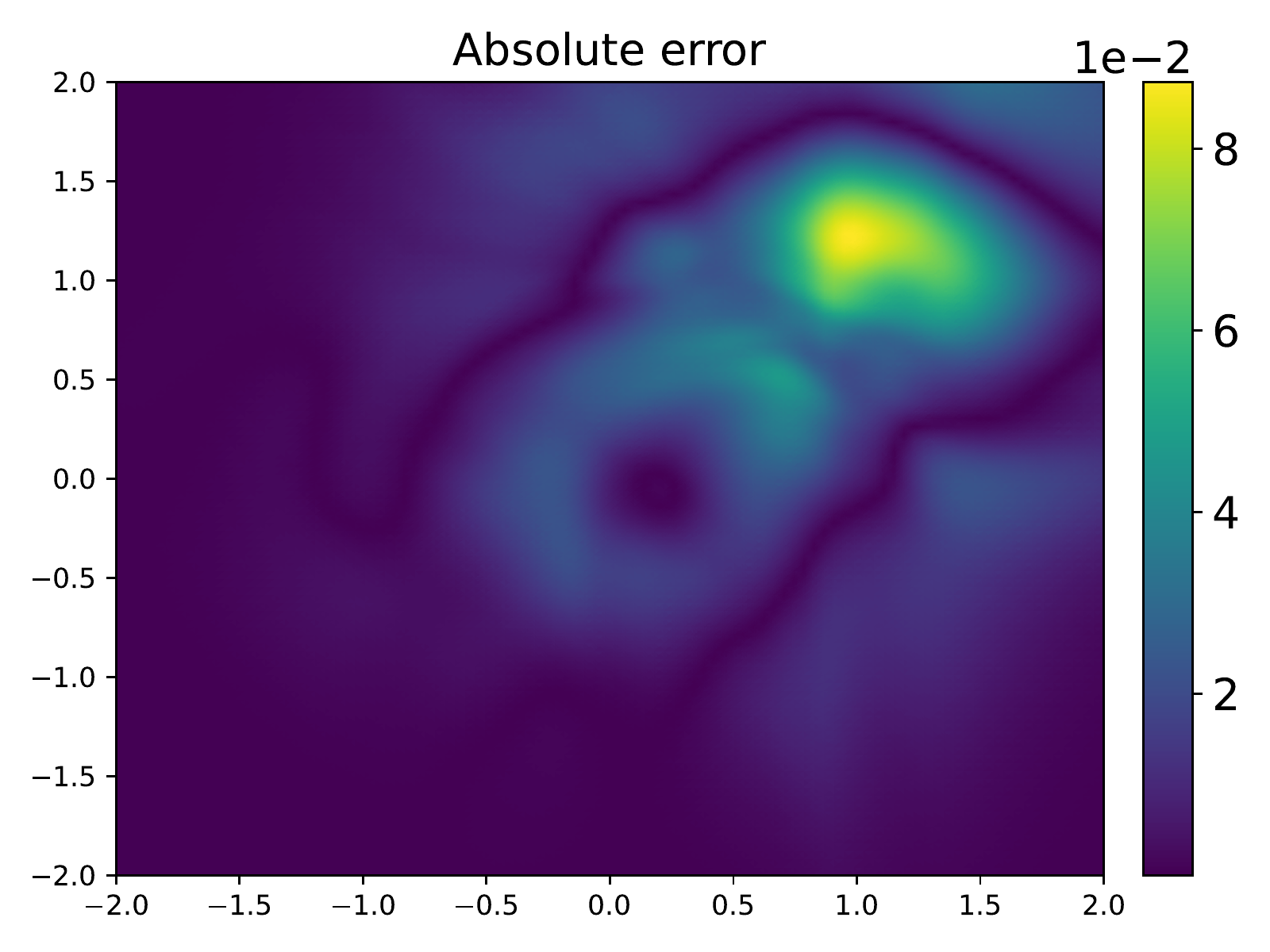}
	\end{minipage}}
	\caption{Comparison between the predicted solutions and the reference solutions. Top row: numerical solution. Bottom row: Absolute error between the numerical solution and the exact solution.}
	\label{bim_solu}
\end{figure}

\begin{figure}[h]
	\centering
	\subfigure[$\alpha=0.5$]{
		\includegraphics[height=3.5cm, width=5.5cm]{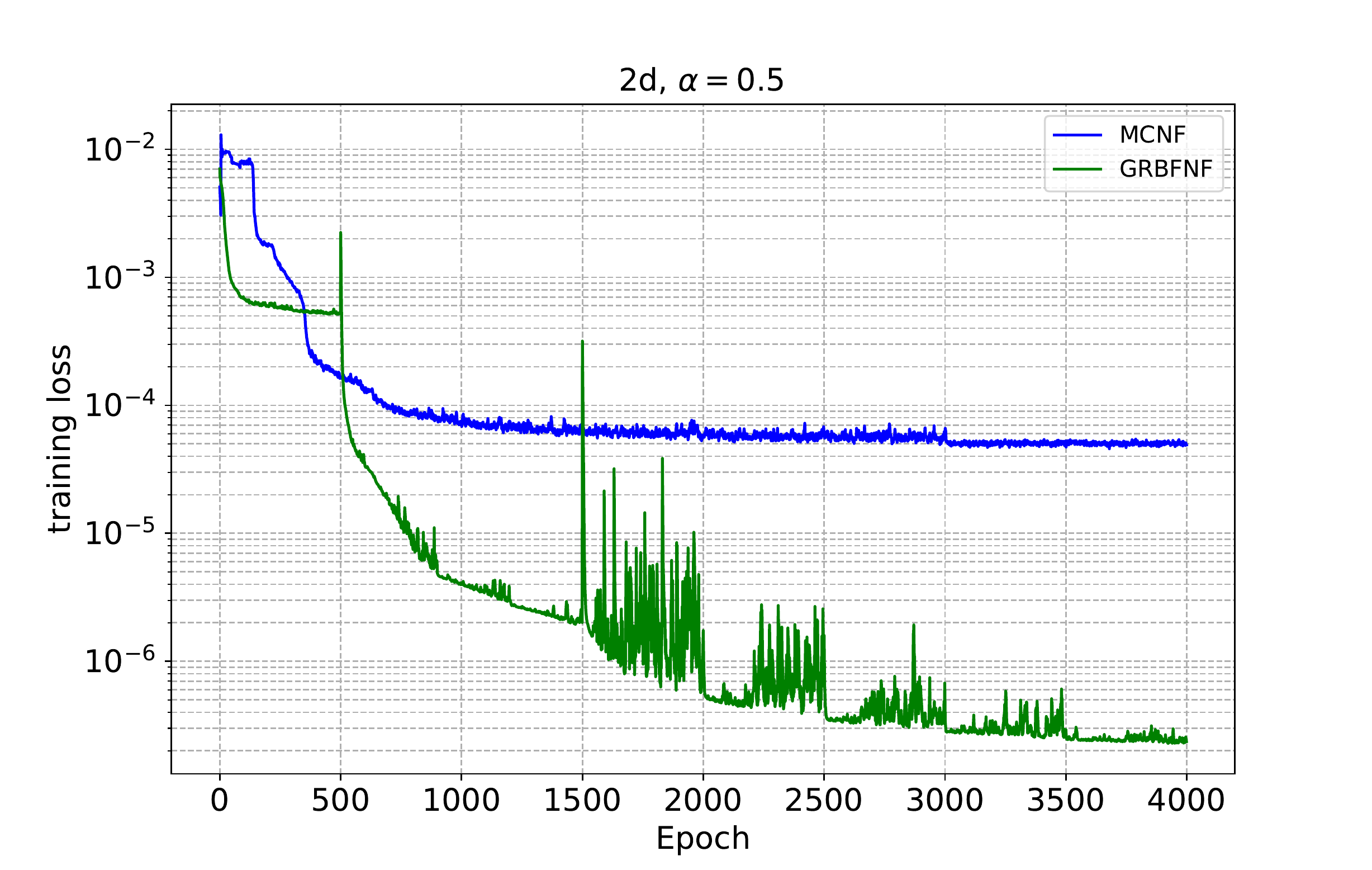}
		\includegraphics[height=3.5cm, width=5.5cm]{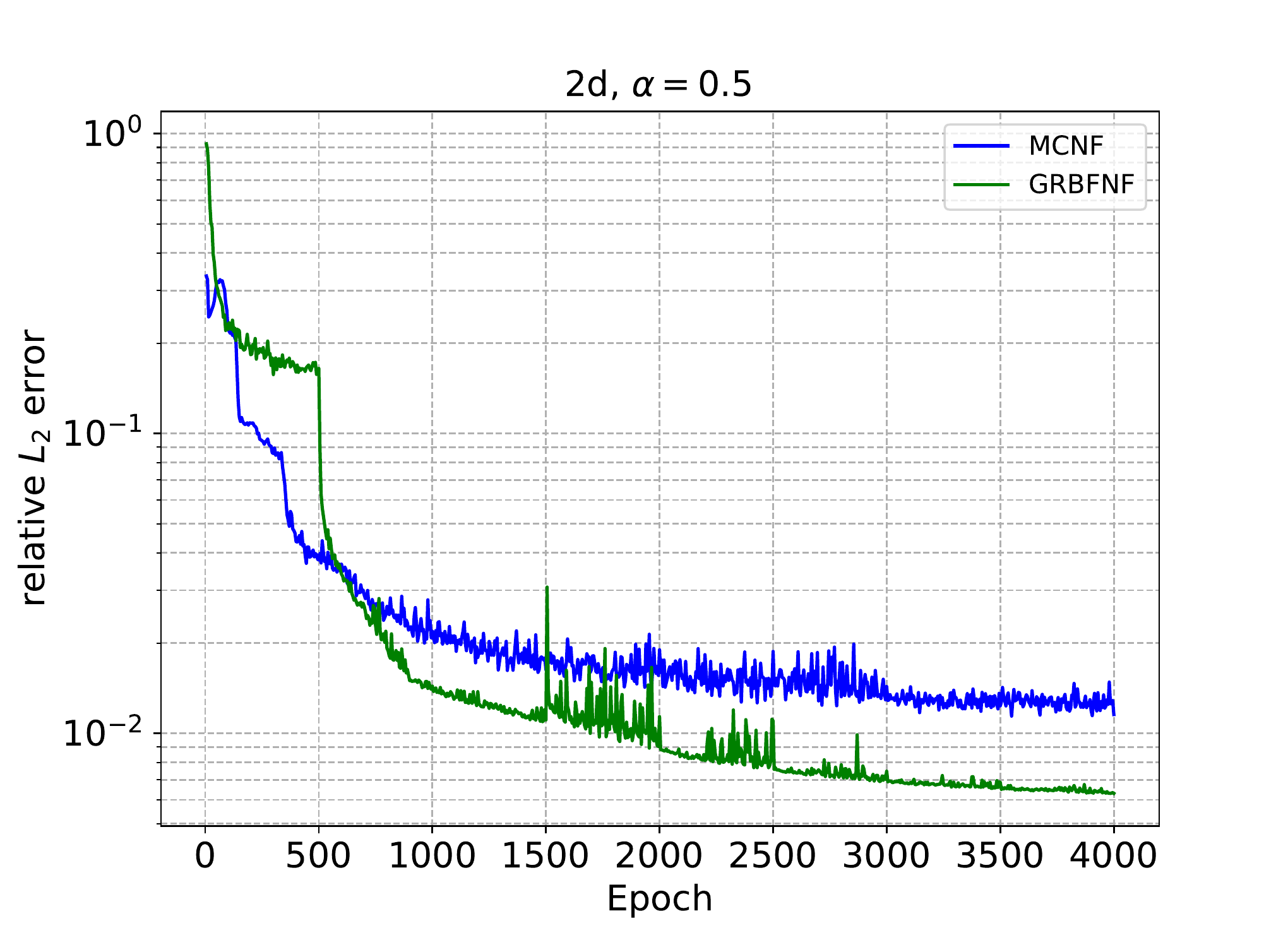}
		\includegraphics[height=3.5cm, width=5.5cm]{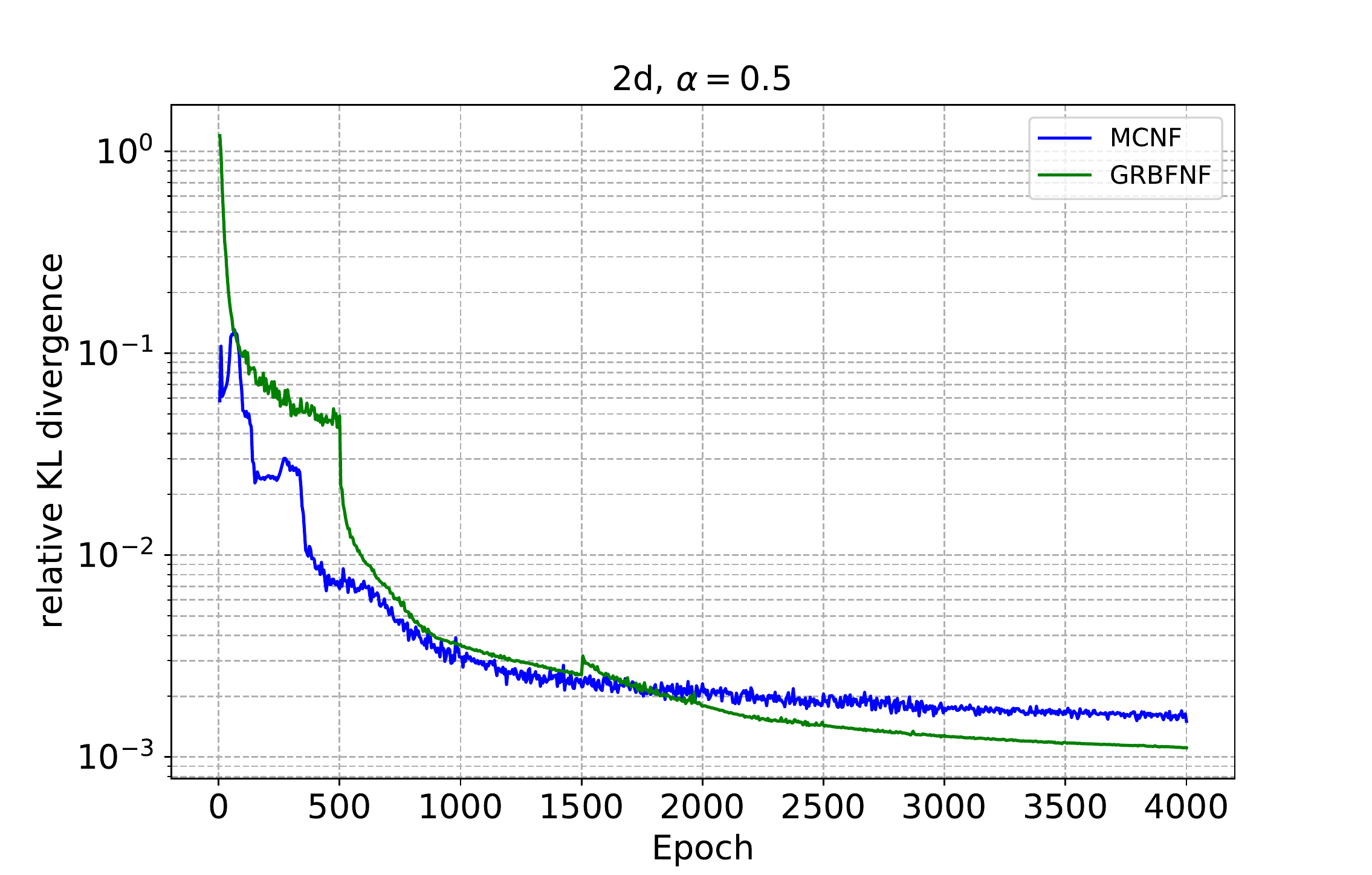}}
	\caption{Convergence behavior of MCNF and GRBFNF for bimodal distribution. Left: training loss. Middle: relative $L_2$ error. Right: relative KL divergence. The unit of time is second.}
	\label{RBFNF_MC_sta_ga2d2}
\end{figure}

We also apply MCNF and GRBFNF without adaptivity to solve this problem. The relative $L_2$ error and the relative KL divergence are provided in Fig. \ref{sta_ga2d2_MC_RBF_no}, which again verifies the strength of the adaptive methods. Although in this example, the unknown PDF is mainly concentrated in the initial sampling area which is different from the previous example, uniform samples used by non-adaptive methods fail to yield an accurate approximation and the adaptive sampling may improve the results by at least one order of magnitude. The exact solution is presented in Fig. \ref{fig:true_FFP_bim}.
 The comparison between the predicted solution and the exact solution are presented in Fig. \ref{bim_solu}.  Both MCNF and GRBFNF can approximate the exact solution well. The training loss, the relative $L_2$ error and the relative KL divergence are presented in Fig. \ref{RBFNF_MC_sta_ga2d2}. The GRBFNF shows better performance than MCNF in this example.

\subsection{High dimensional fractional Fokker-Planck equations}
In this part, we consider a high-dimensional FPE.
\begin{equation}
\left\{\begin{aligned}
&\nabla\cdot(\bm{g}(\bm{x})p(\bm{x}))+\Delta p(\bm{x})-(-\Delta)^{\alpha/2}p(\bm{x})=f(\bm{x}), \quad \bm{x}\in\mathbb{R}^{d},\\
&\int_{\mathbb{R}^2}p(\bm{x}){\rm d}\bm{x}=1, \quad p(\bm{x})\geq0.
\end{aligned}\right.
\label{eq_fp_frac}
\end{equation}

where $\bm{g(x)}=\frac{\bm{x-\mu}}{\sigma^2}$, the corresponding analytic solution is
\begin{equation}
p(\bm{x})=\mathcal{N}(\bm{\mu},\Sigma)=\frac{1}{(2\pi)^{d/2} \sigma^{d}}\exp\left(-\frac{1}{2}(\bm{x-\mu})^{\mathrm{T}}\Sigma^{\mathrm{-1}}(\bm{x-\mu})\right).
\end{equation}
We take $d=4, 6, 8$, $\bm{\mu}=\bm{1}_d$ and $\Sigma=\sigma^2\bm{I}_d$, where $\bm{I}_d$ is a $d$-dimensional identity matrix and $\sigma=2$.

For high-dimensional problems, the PDF and the associated loss function may be too small, which results in numerical underflow. For the sake of numerical stability, we magnify the solution by multiplying a large enough constant $C$. Thus $Cp$ satisfies
\begin{equation}
{\frac {\partial (Cp)}{\partial t}}=\mathcal{L}(Cp)-(-\Delta)^{\alpha/2}(Cp).
\label{ffp_eq_miti}
\end{equation}
Actually, the $C$ used here is $1$ for $d=4$, $10$ for $d=6$ and $200$ for $d=8$.
\begin{figure}[!h]
	\centering
	\begin{minipage}[t]{0.3\linewidth}
		\includegraphics[height=4cm, width=4.5cm]{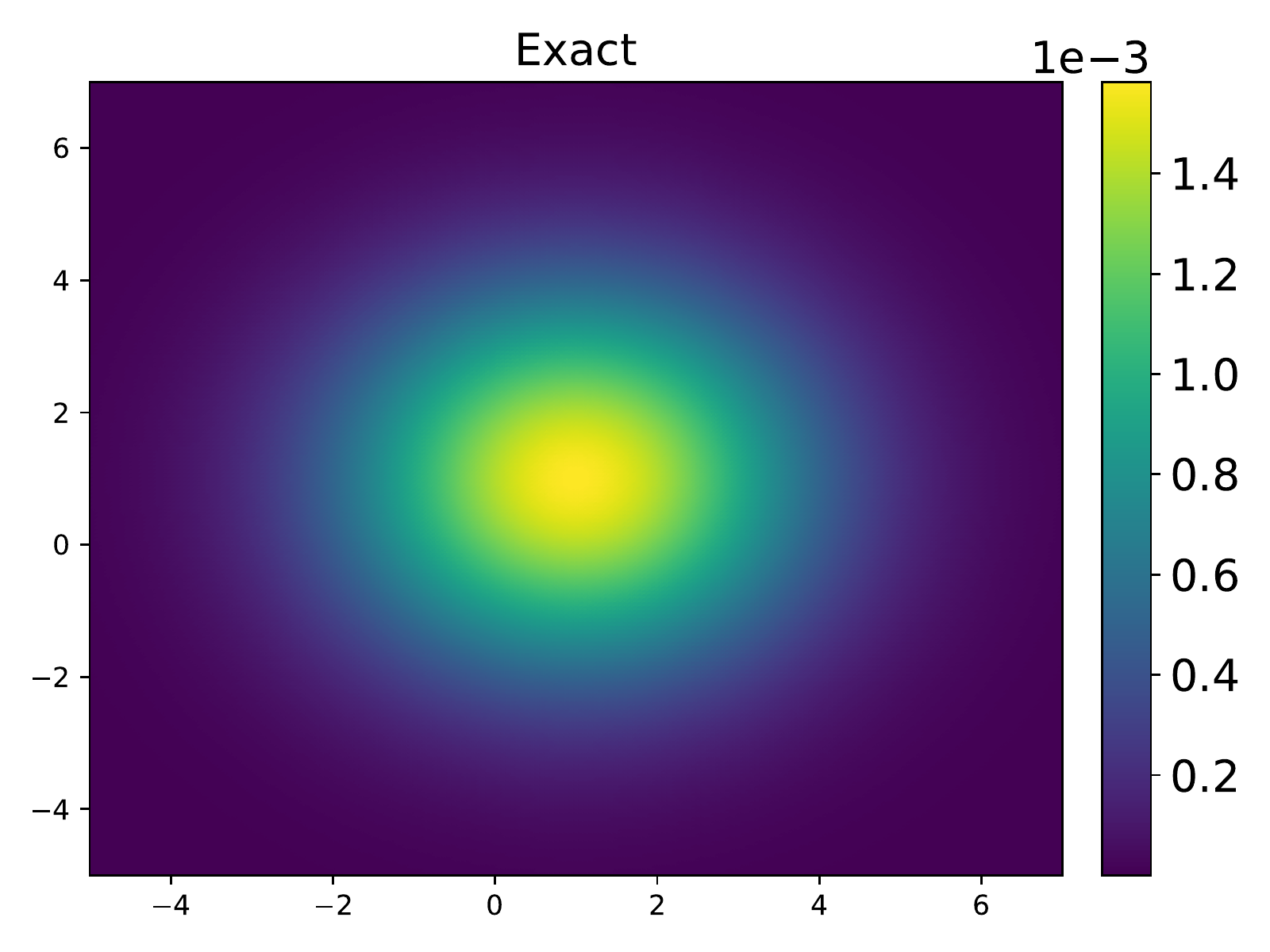}
	\end{minipage}
	\begin{minipage}[t]{0.3\linewidth}	
		\includegraphics[height=4cm, width=4.5cm]{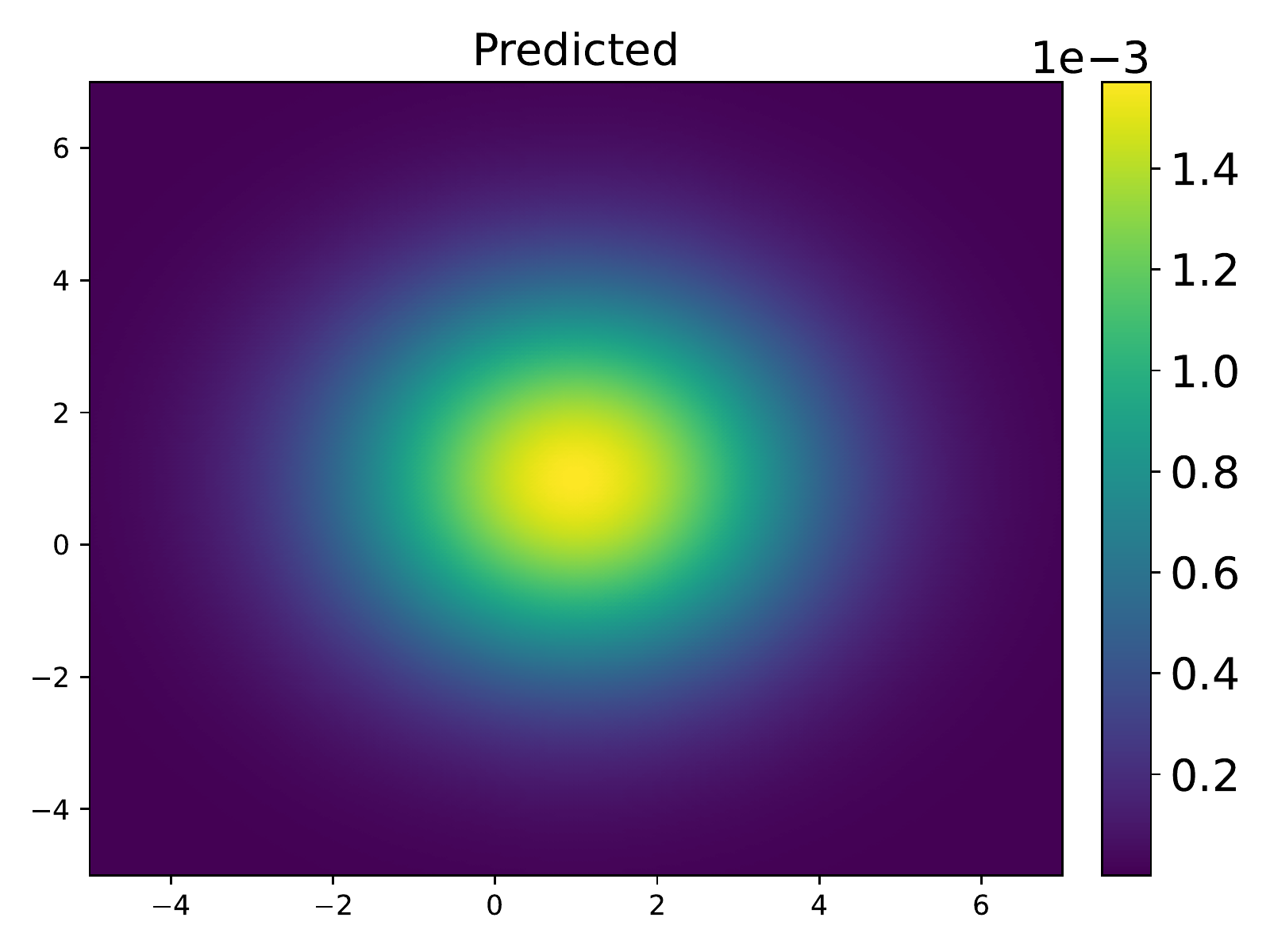}
	\end{minipage}
	\begin{minipage}[t]{0.3\linewidth}	
		\includegraphics[height=4cm, width=4.5cm]{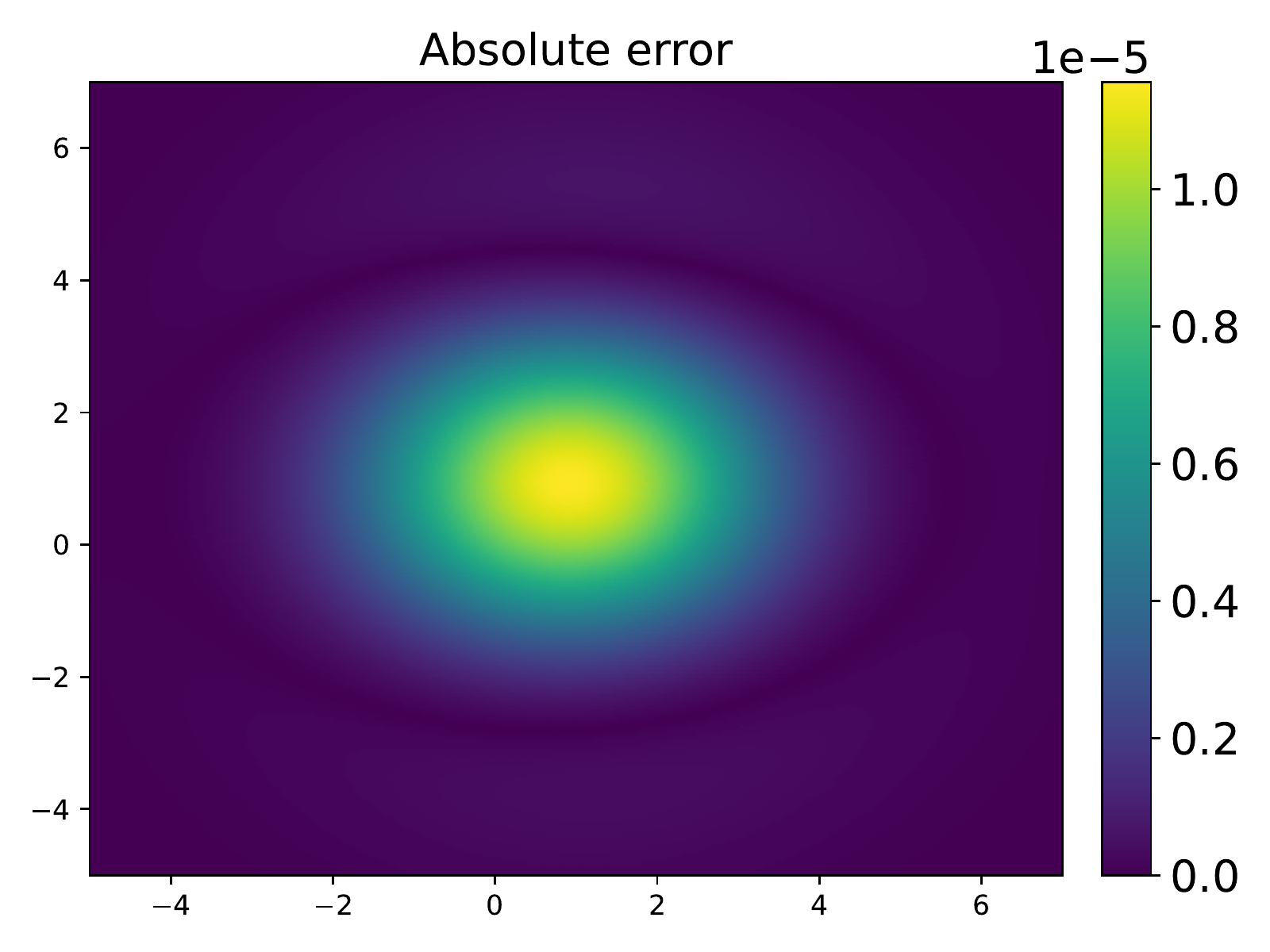}
	\end{minipage}
	\caption{MCNF for 4-dimensional problem, where the first two dimensions are plotted at $x_3=x_4=1$. Predicted solution versus the reference solution. Left: exact solution. Middle: prediction. Right: absolute error.}
	\label{d4}
\end{figure}

\begin{figure}[!h]
	\centering
	\begin{minipage}[t]{0.3\linewidth}
		\includegraphics[height=4cm, width=4.5cm]{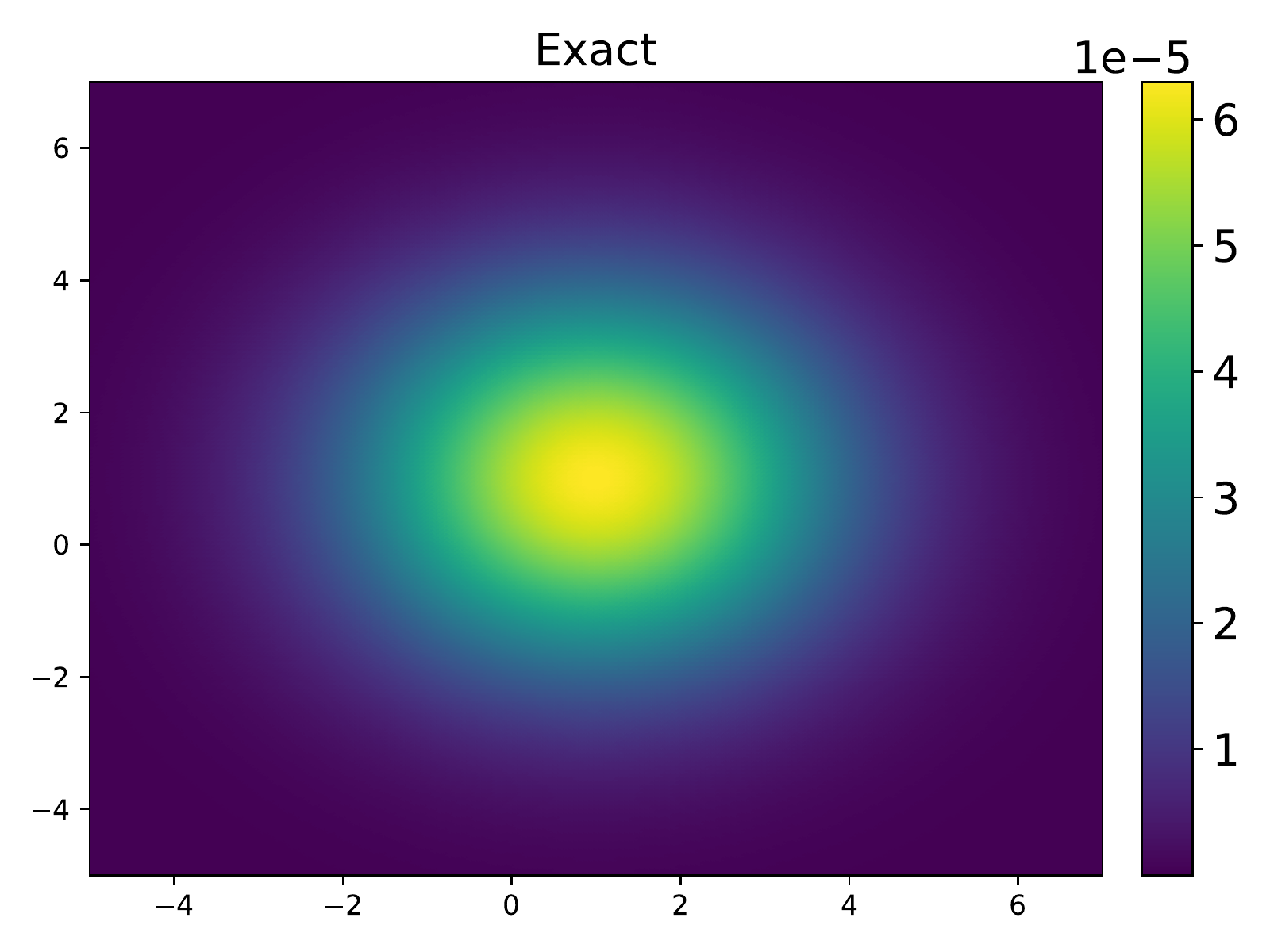}
	\end{minipage}
	\begin{minipage}[t]{0.3\linewidth}	
		\includegraphics[height=4cm, width=4.5cm]{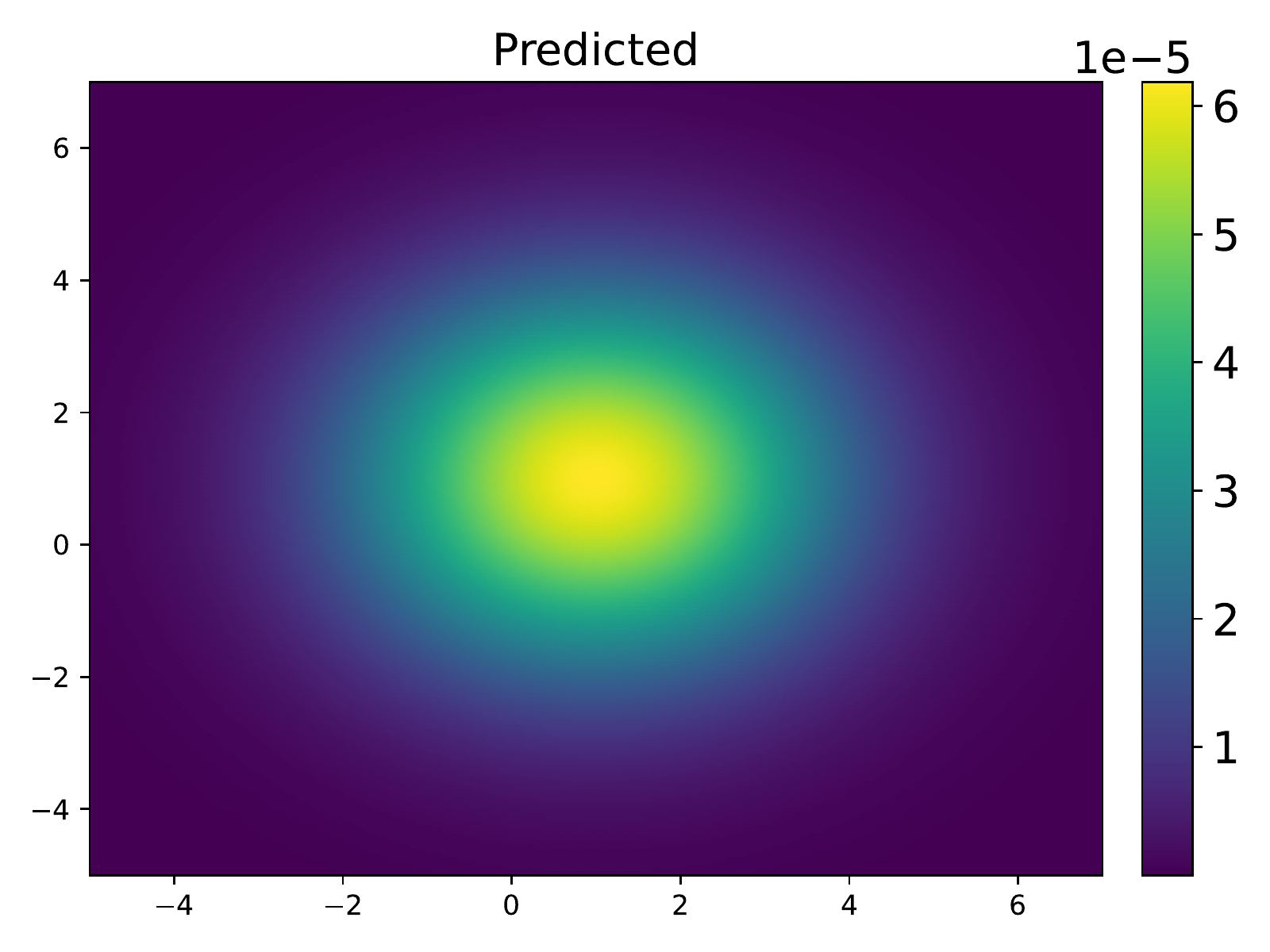}
	\end{minipage}
	\begin{minipage}[t]{0.3\linewidth}	
		\includegraphics[height=4cm, width=4.5cm]{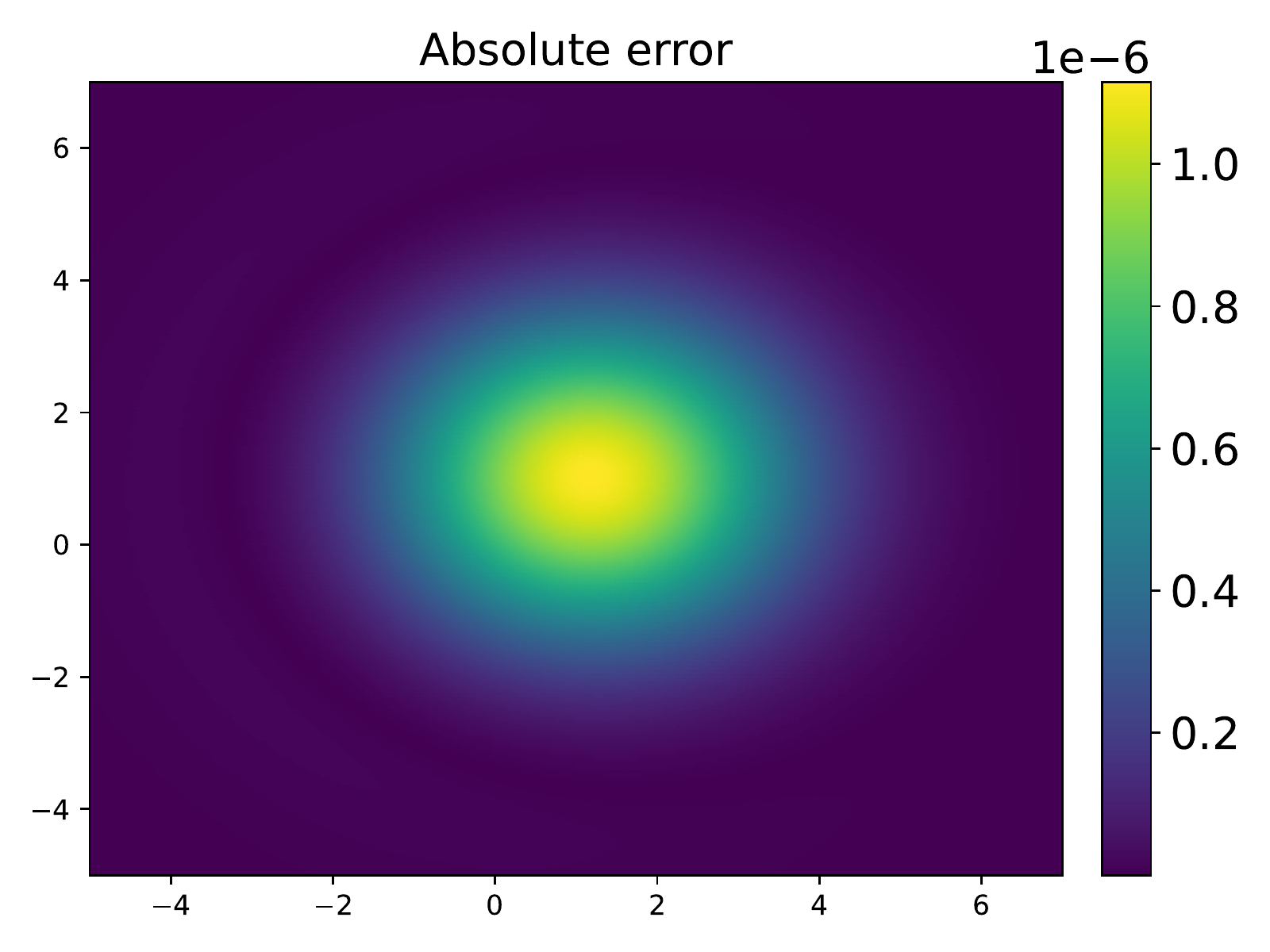}
	\end{minipage}
	\caption{MCNF for 6-dimensional problem, where the first two dimensions are plotted at $x_3=x_4=x_5=x_6=1$. Predicted solution versus the reference solution. Left: exact solution. Middle: prediction. Right: absolute error.}
	\label{d6}
\end{figure}

\begin{figure}[!h]
	\centering
	\begin{minipage}[t]{0.3\linewidth}
		\includegraphics[height=4cm, width=4.5cm]{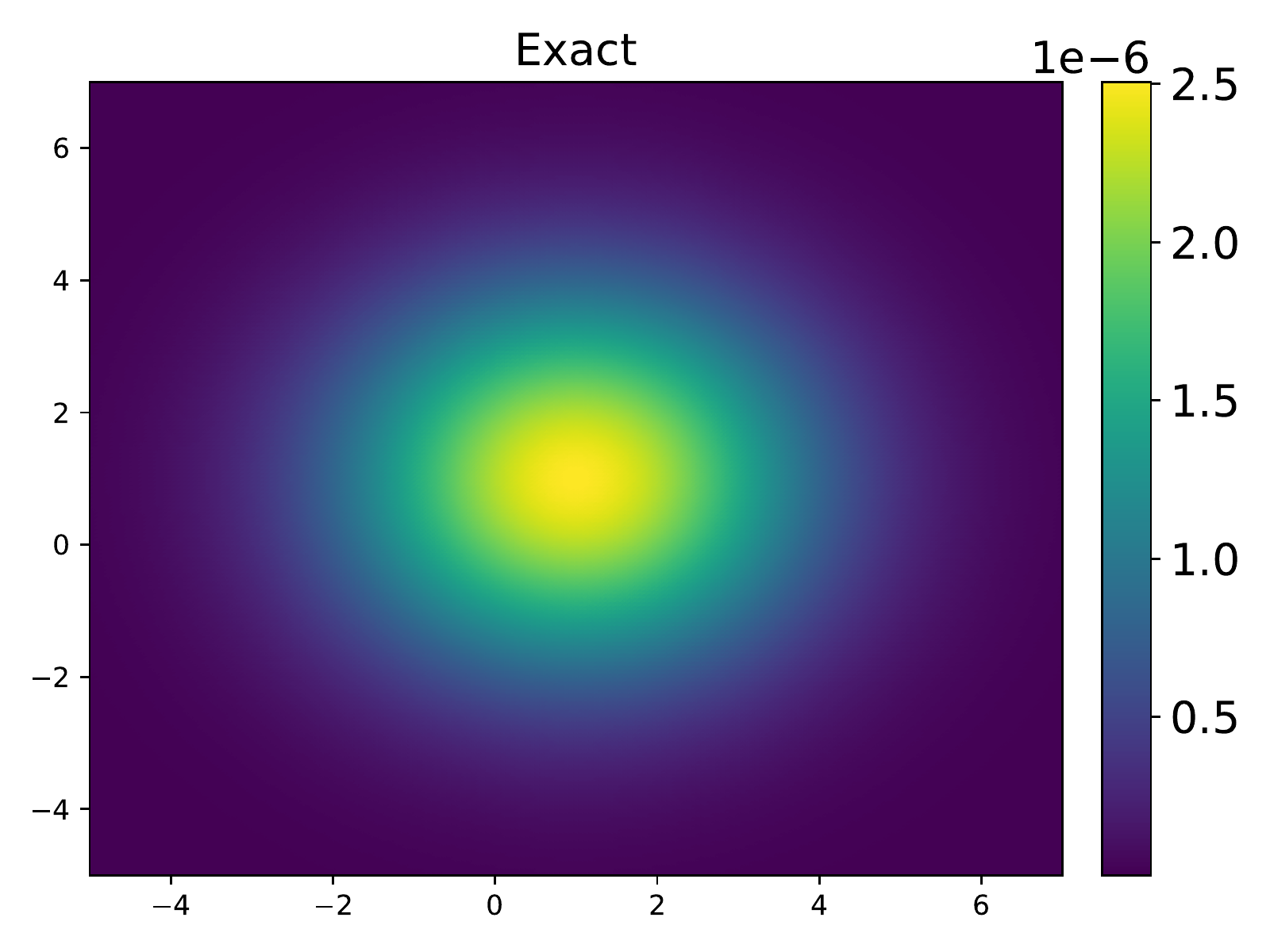}
	\end{minipage}
	\begin{minipage}[t]{0.3\linewidth}	
		\includegraphics[height=4cm, width=4.5cm]{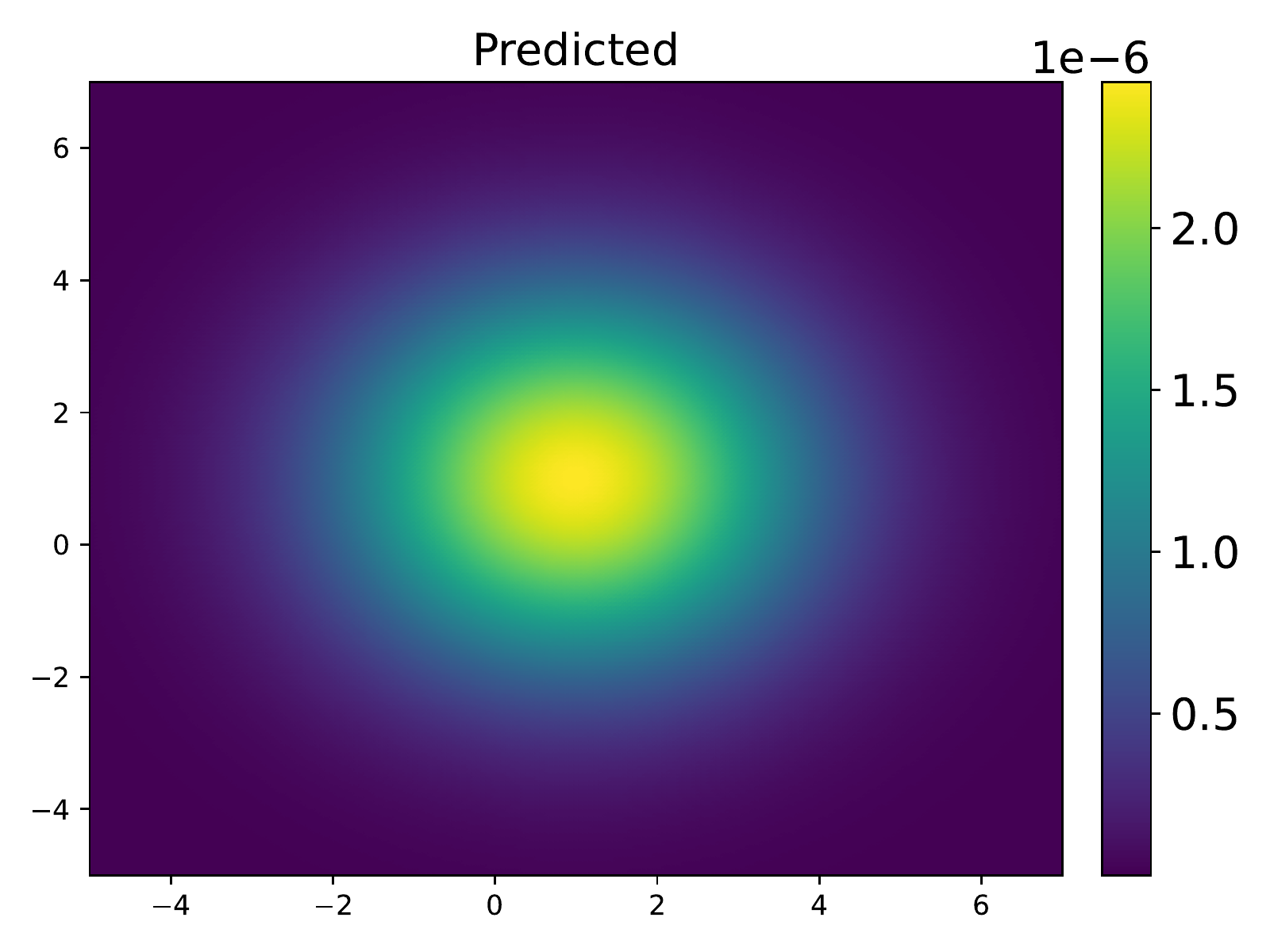}
	\end{minipage}
	\begin{minipage}[t]{0.3\linewidth}	
		\includegraphics[height=4cm, width=4.5cm]{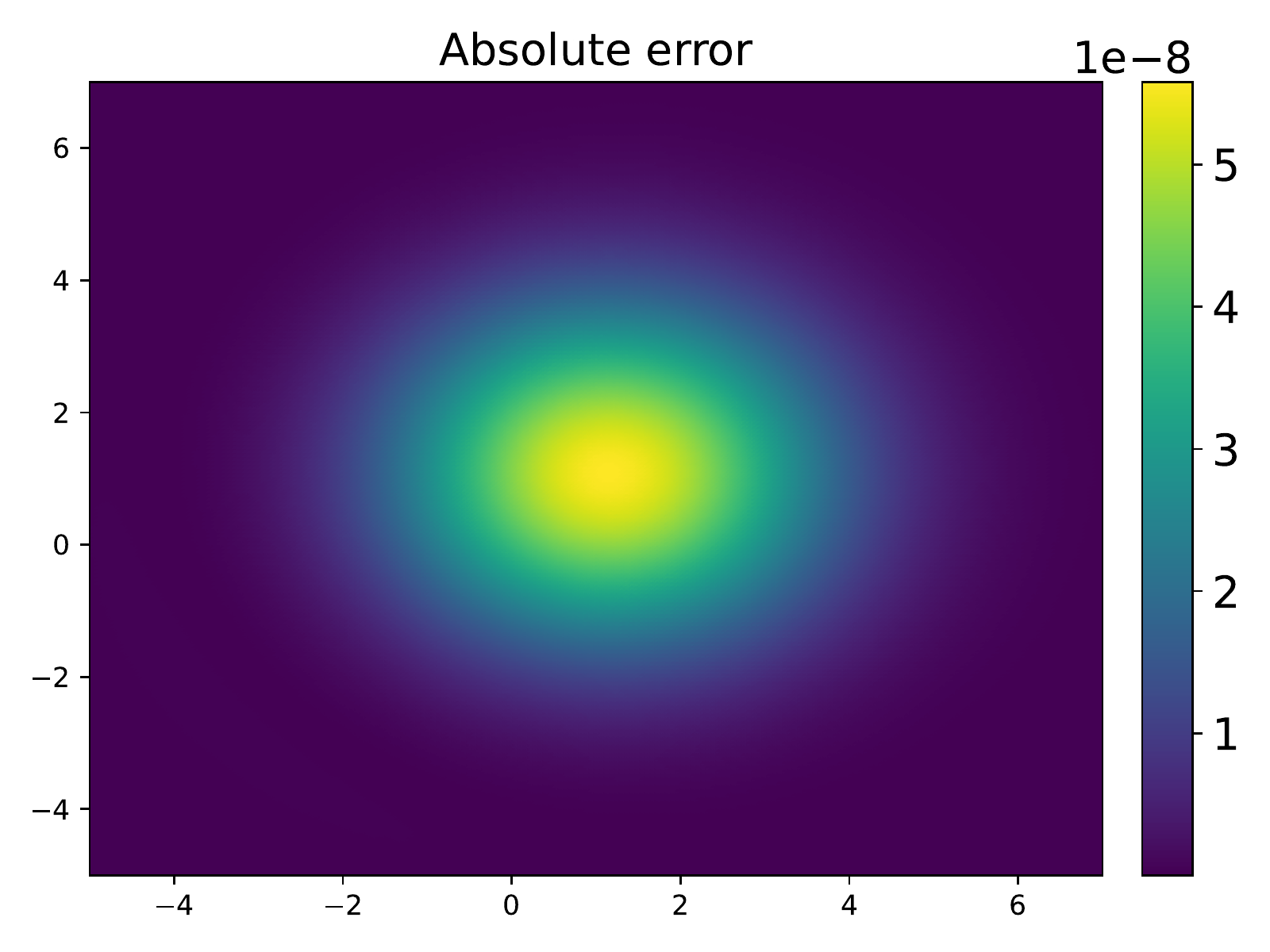}
	\end{minipage}
	\caption{MCNF for the 8-dimensional problem, where the first two dimensions are plotted at $x_3=x_4=\dots=x_8=1$. Predicted solution versus the reference solution. Left: exact solution. Middle: prediction. Right: absolute error.}
	\label{d8}
\end{figure}
\begin{figure}[h]
	\centering
	\subfigure{
		\includegraphics[height=3.5cm, width=5.5cm]{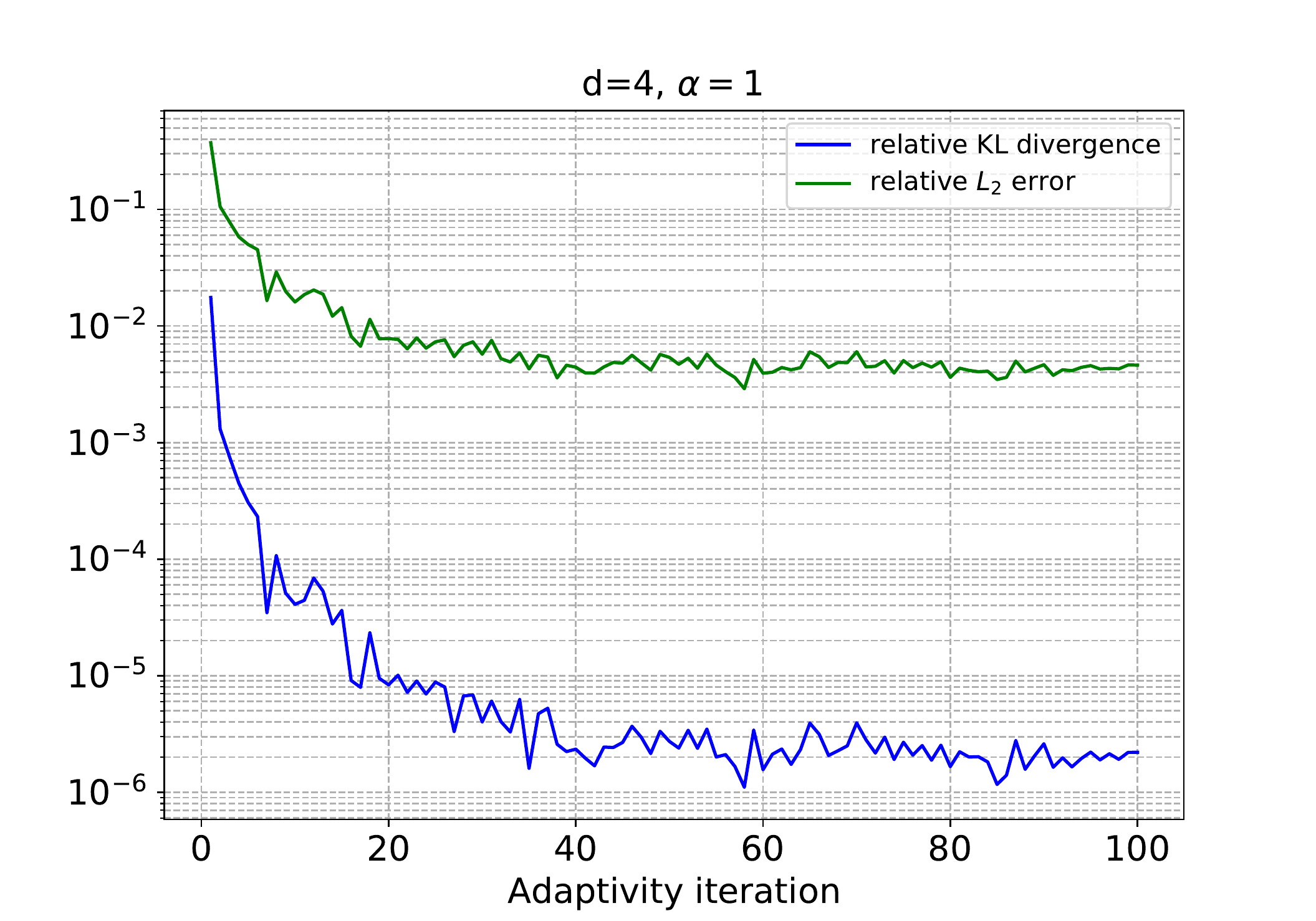}
		\includegraphics[height=3.5cm, width=5.5cm]{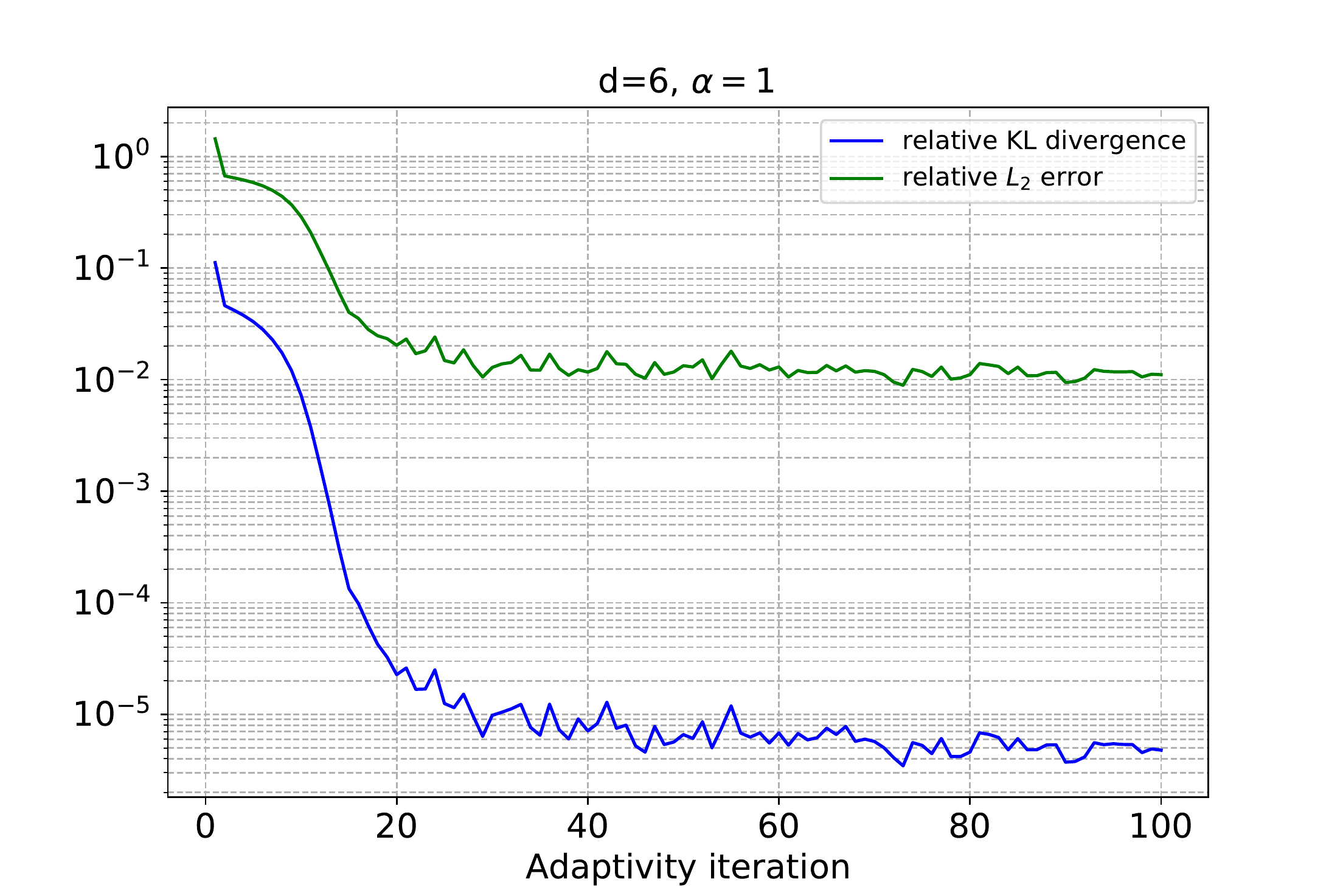}
		\includegraphics[height=3.5cm, width=5.5cm]{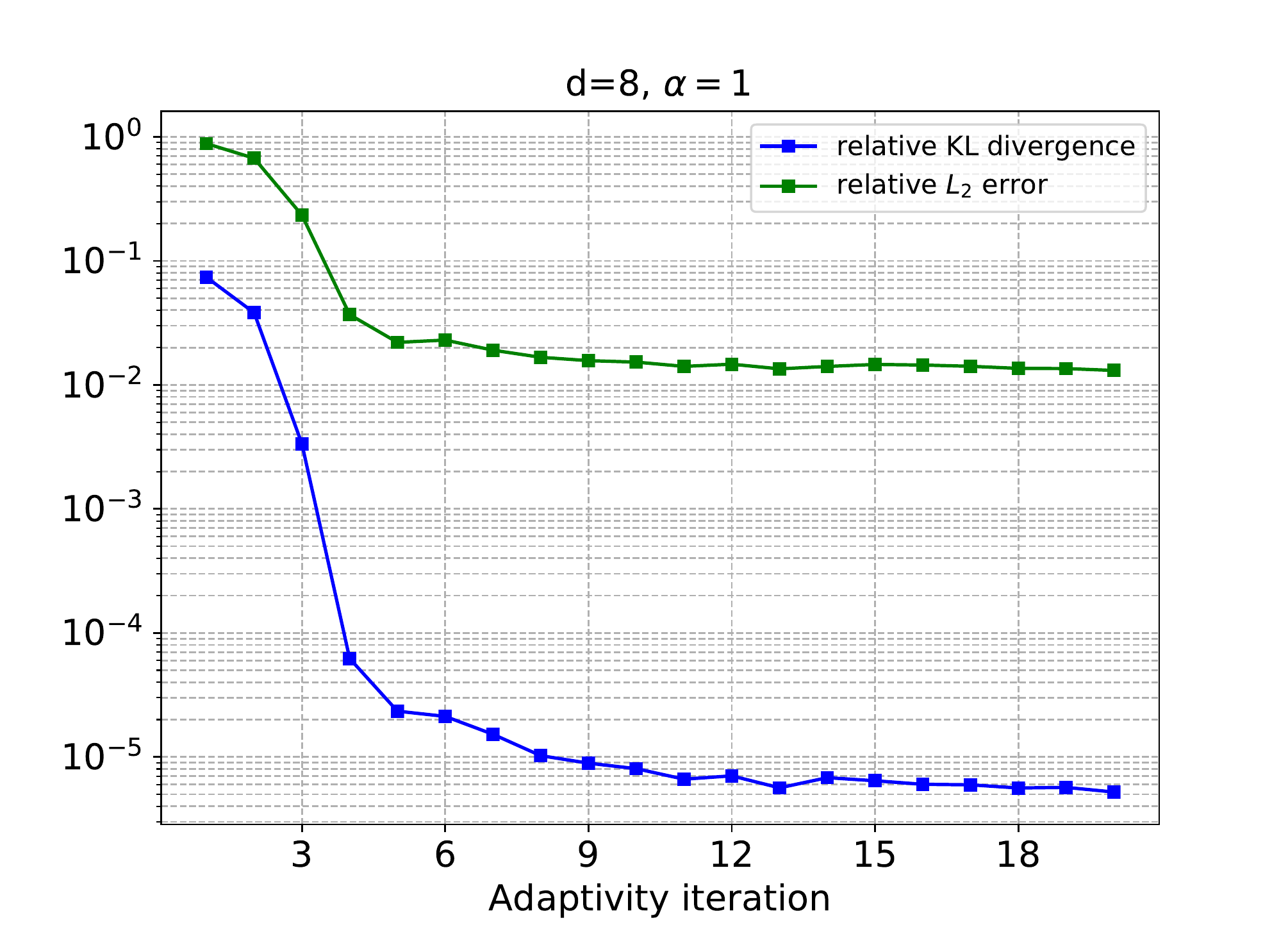}}
	\caption{{Convergence behavior of MCNF for high-dimensional FPEs.} Left: $d=4$. Middle: $d=6$. Right: $d=8$. }
	\label{MCNF_high}
\end{figure}

For NF, we take $L=8$ affine coupling layers with $64$ hidden neurons.  The initial training set is generated via the uniform distributed points in $[-3,5]^d$.  The sample size is $50000$. The batch size is set to be $4096$. We employ MCNF in this problem. GRBFNF is harder to train in high dimensional case since its structure is more complex. The number of samples used to approximate the fractional Laplacian is $200$, $r_0=0.3$,  $\epsilon=0.0001$. We take half-half partition here. For $d=4, 6$,  $100$ adaptivity iterations with 20 epochs for each adaptivity iteration are conducted.  The learning rate is $0.001$ with half decay each $300$ steps.
For $d=8$,  $20$ adaptivity iterations with 200 epochs for each adaptivity iteration are conducted. The learning rate is $0.001$ with half decay each $1000$ steps. The comparisons between the MCNF solutions and the true solutions are presented in Fig. \ref{d4}, Fig. \ref{d6}, Fig. \ref{d8}, which all show great performance of our approach. We also present the relative $L_2$ error and the relative KL divergence in Fig. \ref{MCNF_high}.

\subsection{Time-dependent fractional FPE: Cauchy distribution}
We consider the following stochastic process,
\begin{equation}
\mathrm{d}\mathbf {X} _{t}= \mathrm{d}\mathbf{L}^{\alpha}_t, \quad \alpha=1.
\label{cauchy_sde}
\end{equation}
For $d=2$,  the corresponding fractional Fokker-Planck equation is
\begin{equation}
\begin{aligned}
&\frac{\partial p}{\partial t}=-(-\Delta)^{\alpha/2}p, \quad \alpha=1,\\
&p(\bm{x}, 0)=p_0(\bm{x}).
\end{aligned}
\label{cau_fpe_2d}
\end{equation}
For the initial condition $p_0(\bm{x})=\frac{1}{2\pi(1+\left\|\bm{x}\right\|_2^2)^{3/2}}$, 
the solution of \eqref{cau_fpe_2d} is $p(\bm{x},t)=\frac{t+1}{2\pi\big((t+1)^2+\left\|\bm{x}\right\|_2^2\big)^{3/2}}$, where $\bm{x}\in \mathbb{R}^2$ and $t\in [0,1]$.

\begin{figure}[h]
	\centering
	\subfigure[$t=0$]{
		\includegraphics[height=4cm, width=4.5cm]{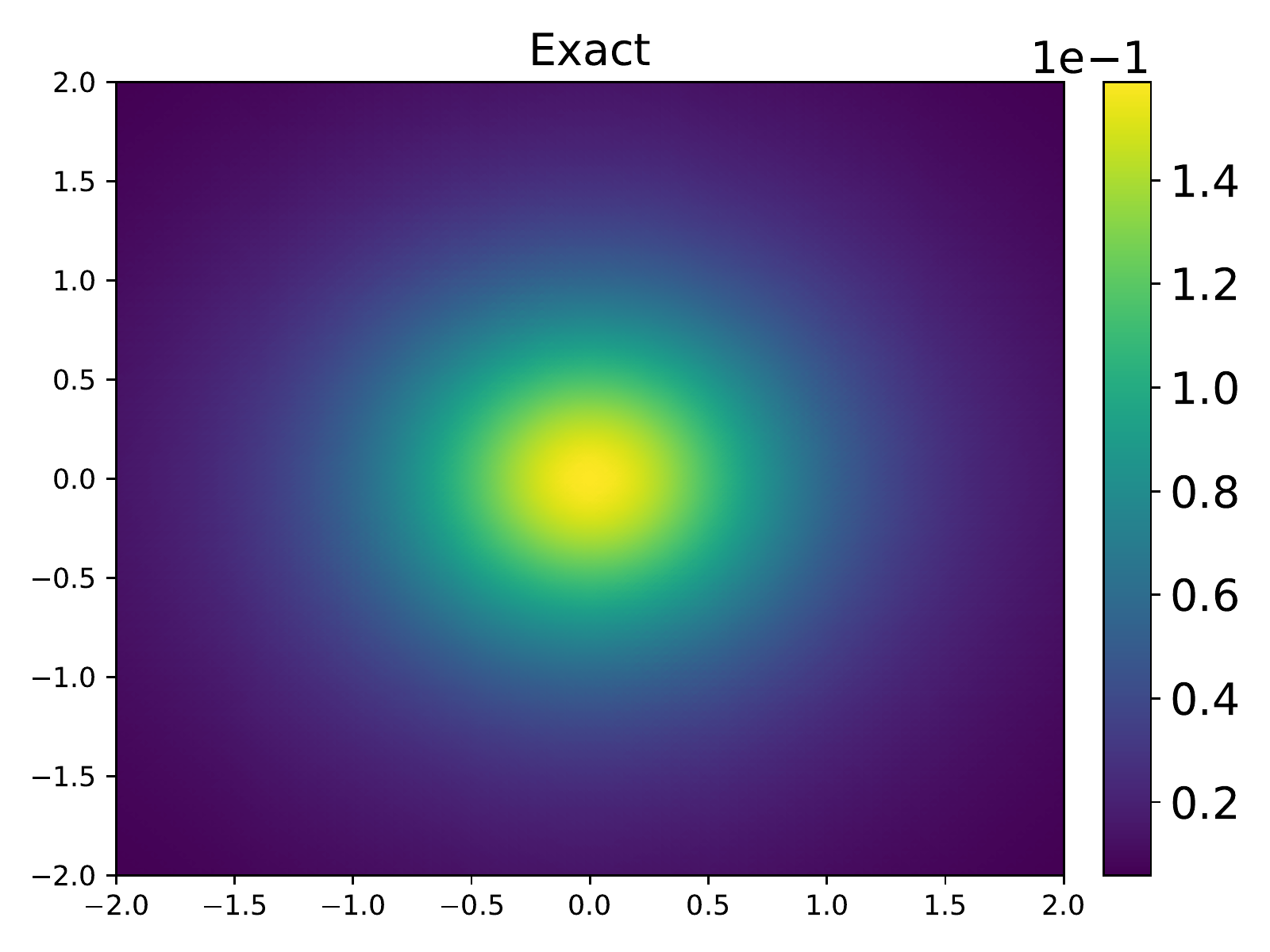}
		\quad
		\includegraphics[height=4cm, width=4.5cm]{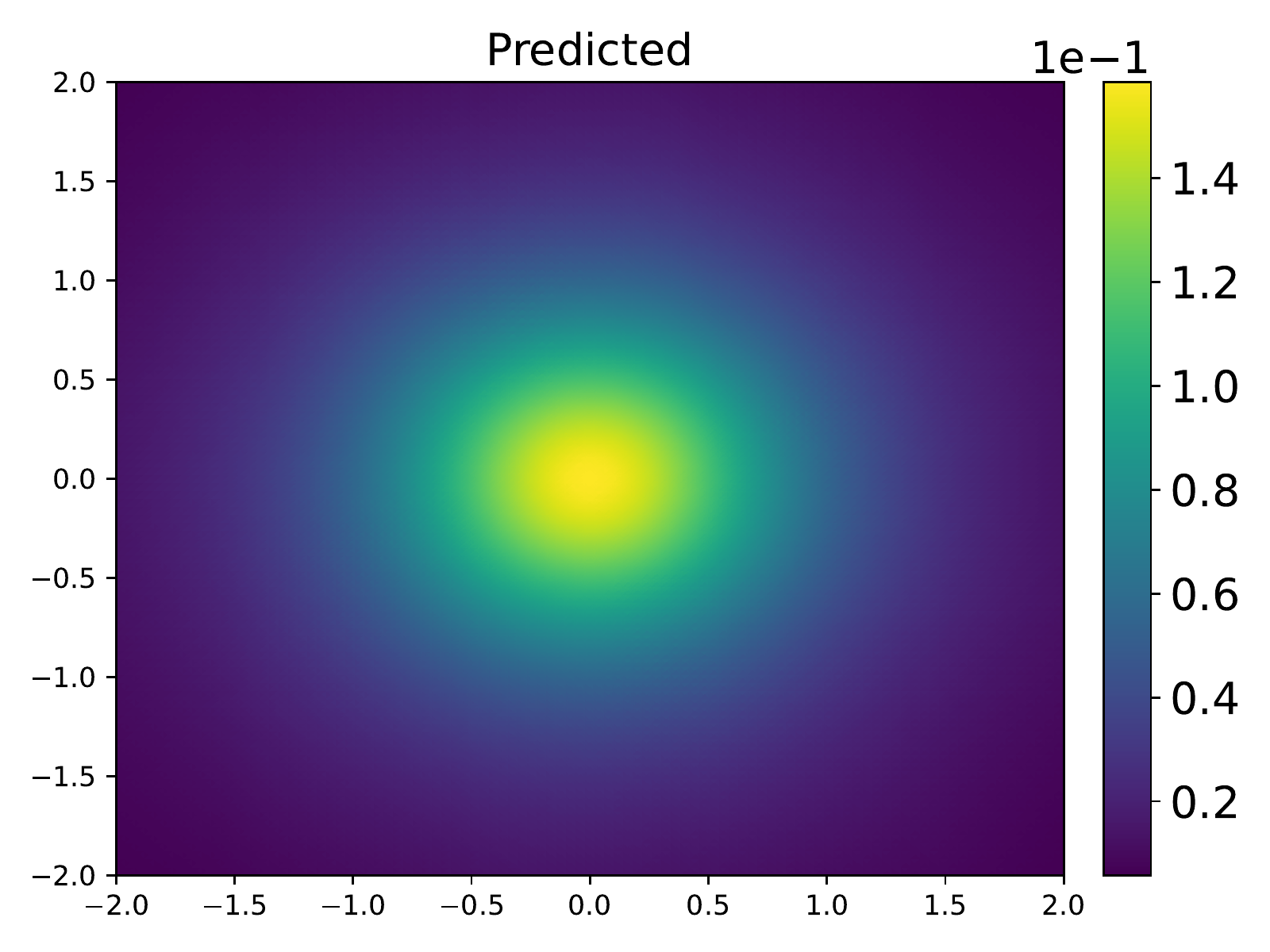}
		\quad
		\includegraphics[height=4cm, width=4.5cm]{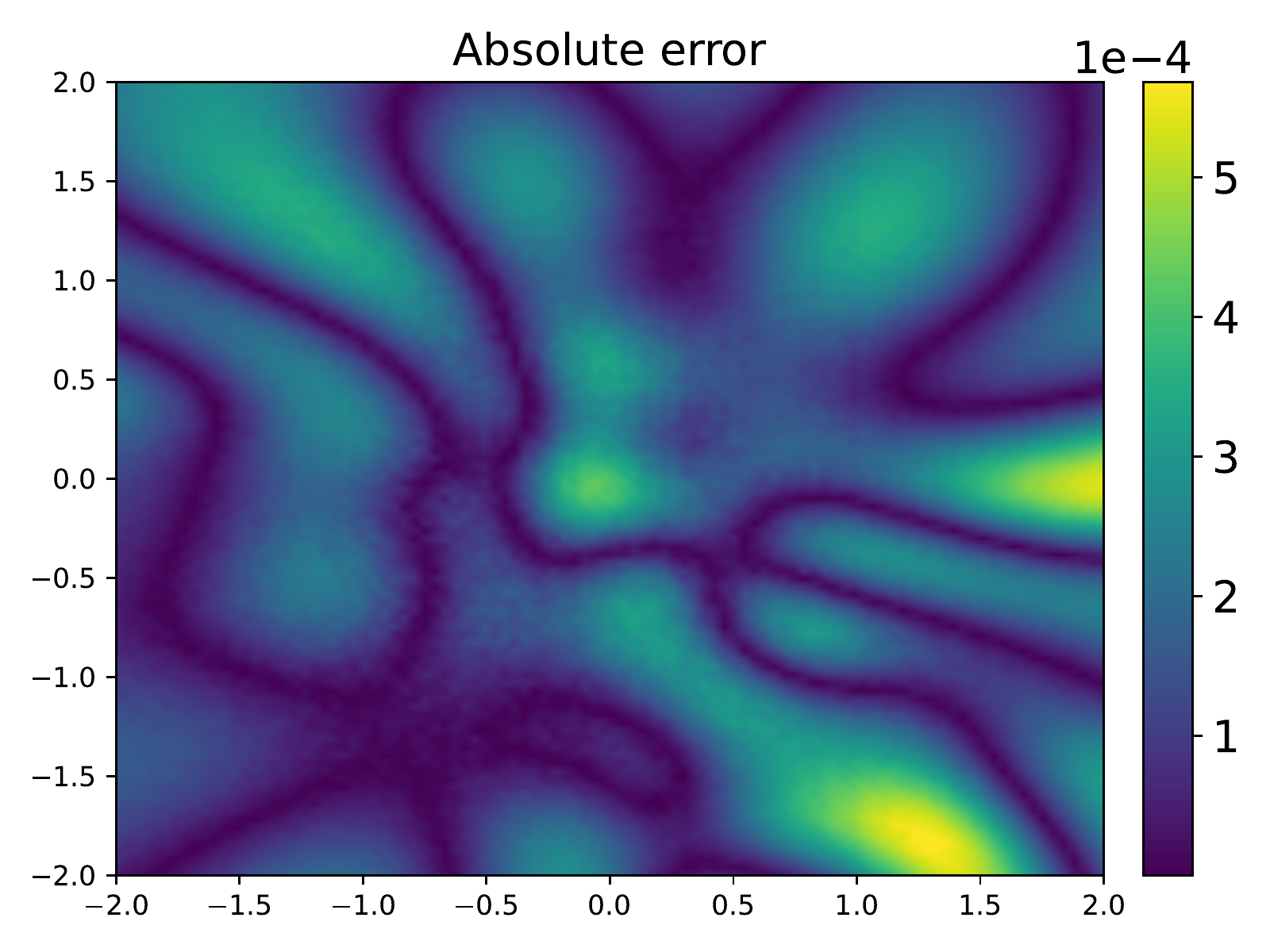}
	}
	\subfigure[$t=0.5$]{
		\includegraphics[height=4cm, width=4.5cm]{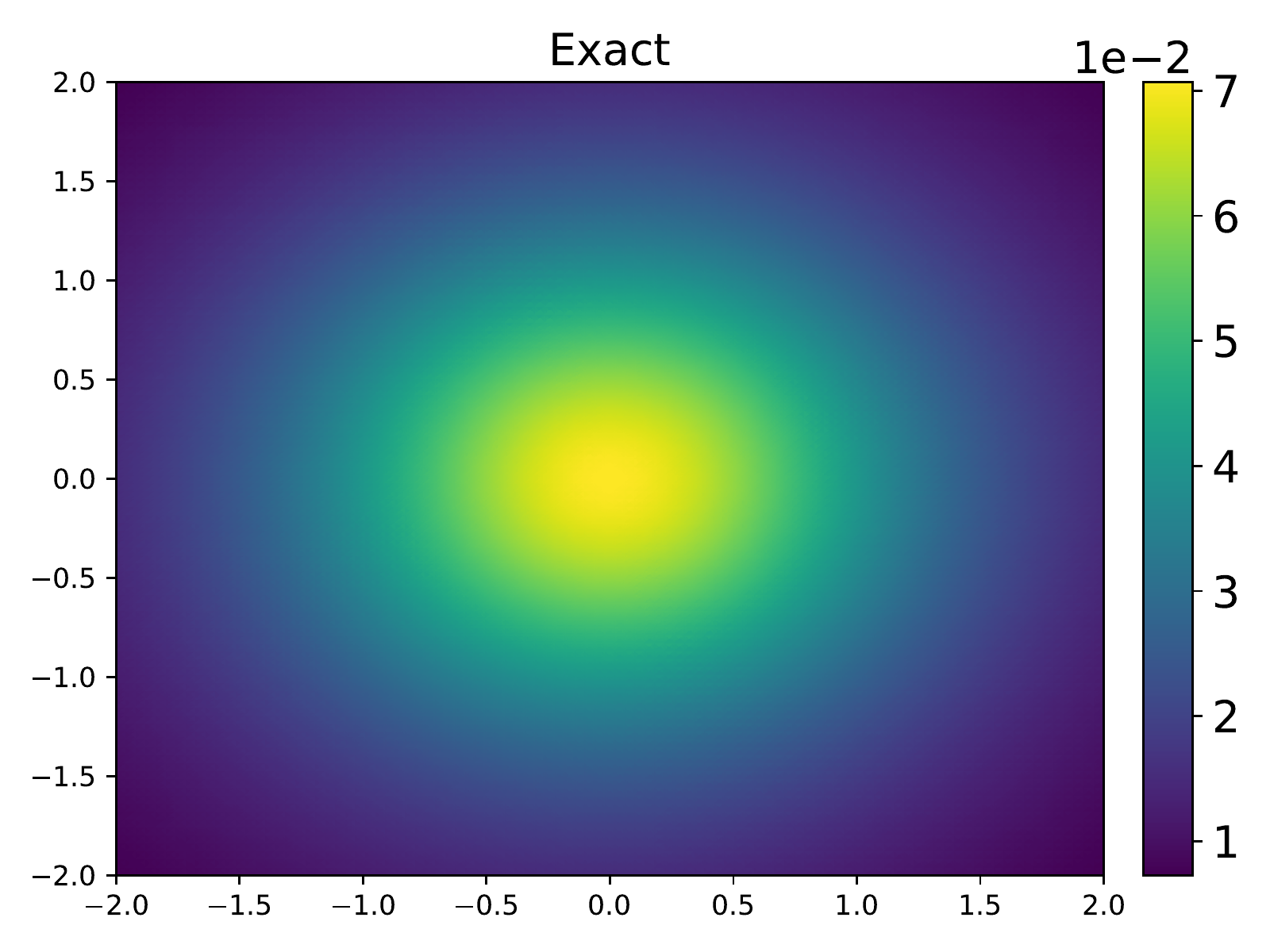}
		\quad
		\includegraphics[height=4cm, width=4.5cm]{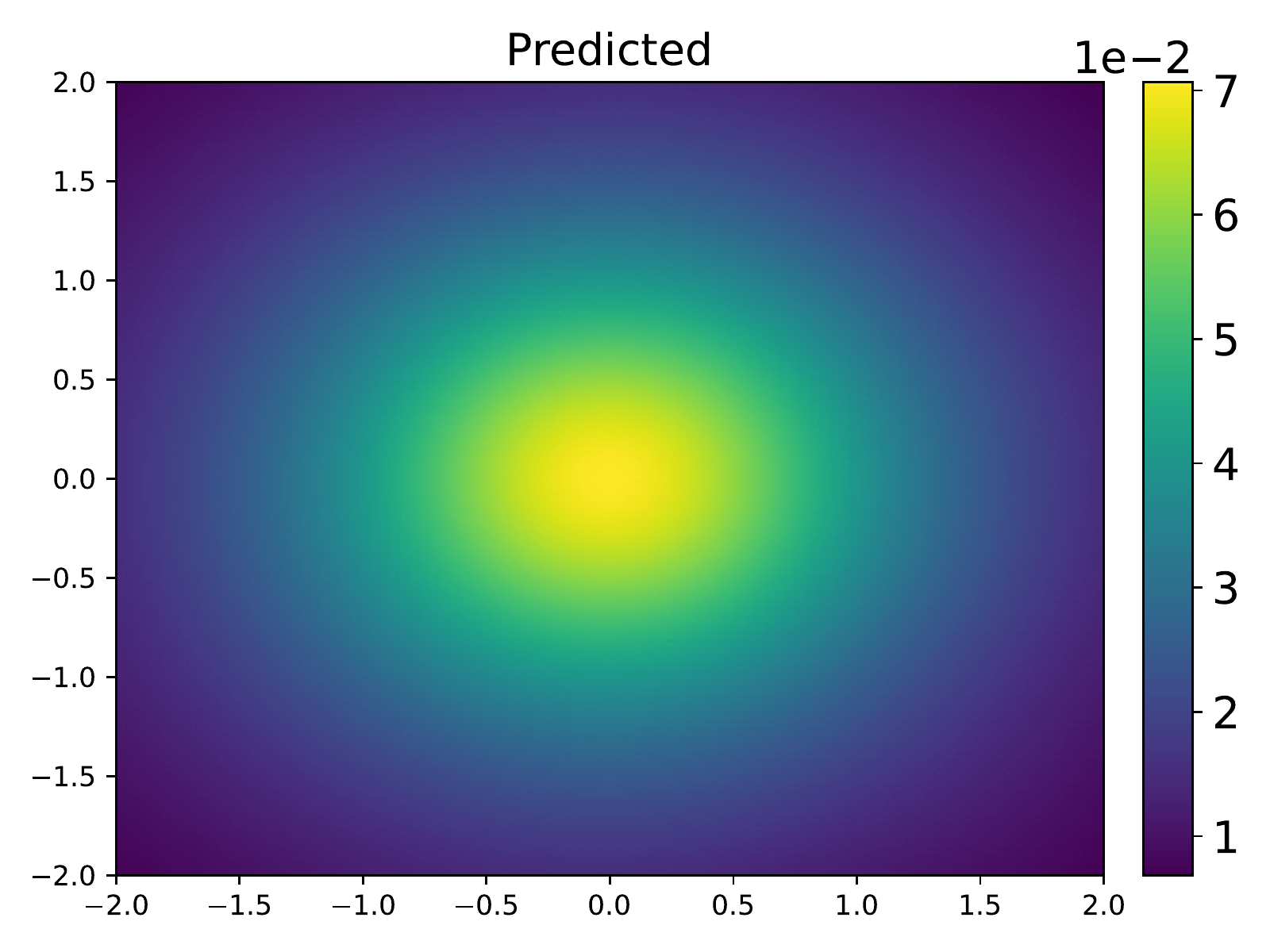}
		\quad
		\includegraphics[height=4cm, width=4.5cm]{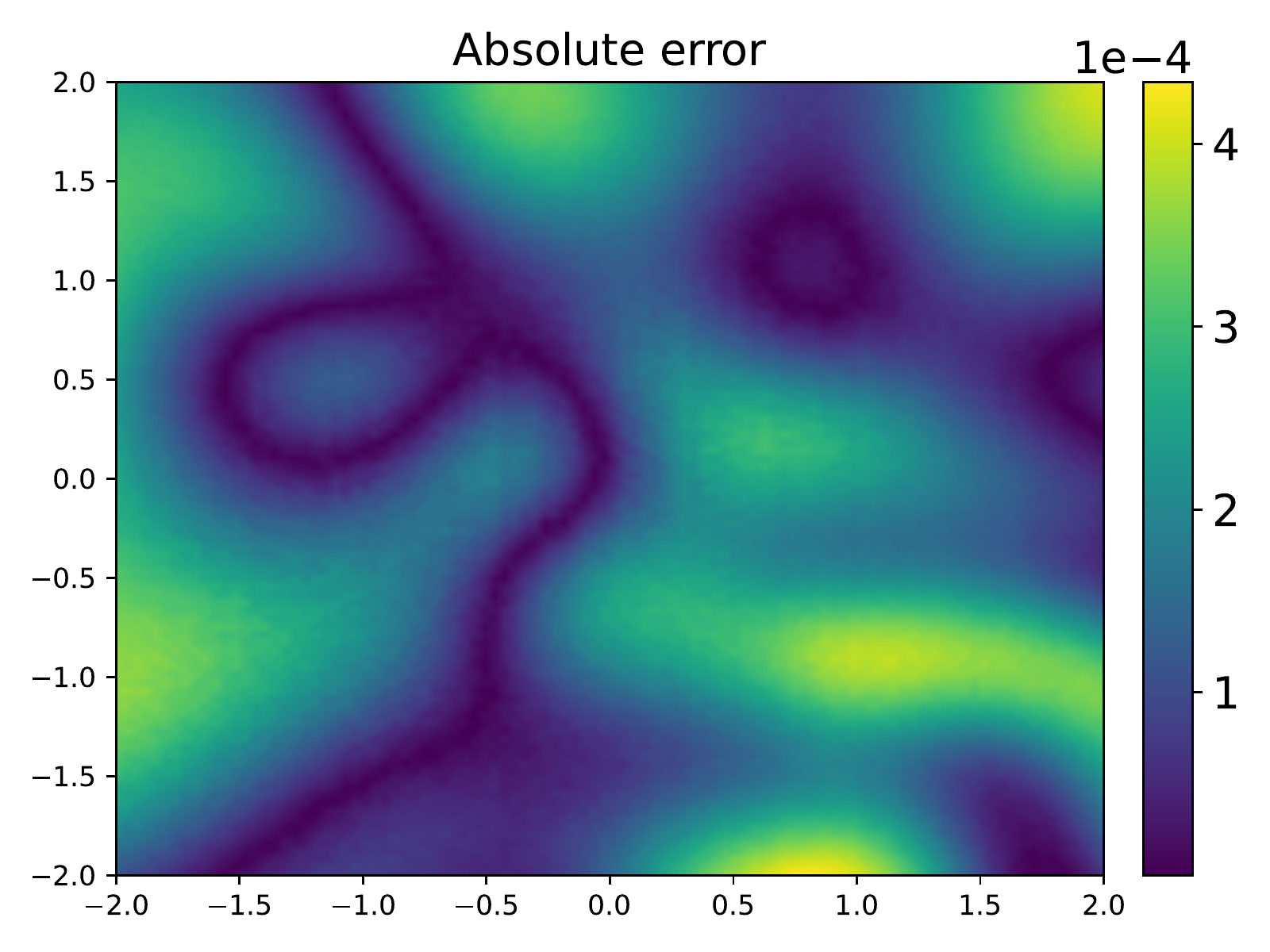}
	}
	\subfigure[$t=1$]{
		\includegraphics[height=4cm, width=4.5cm]{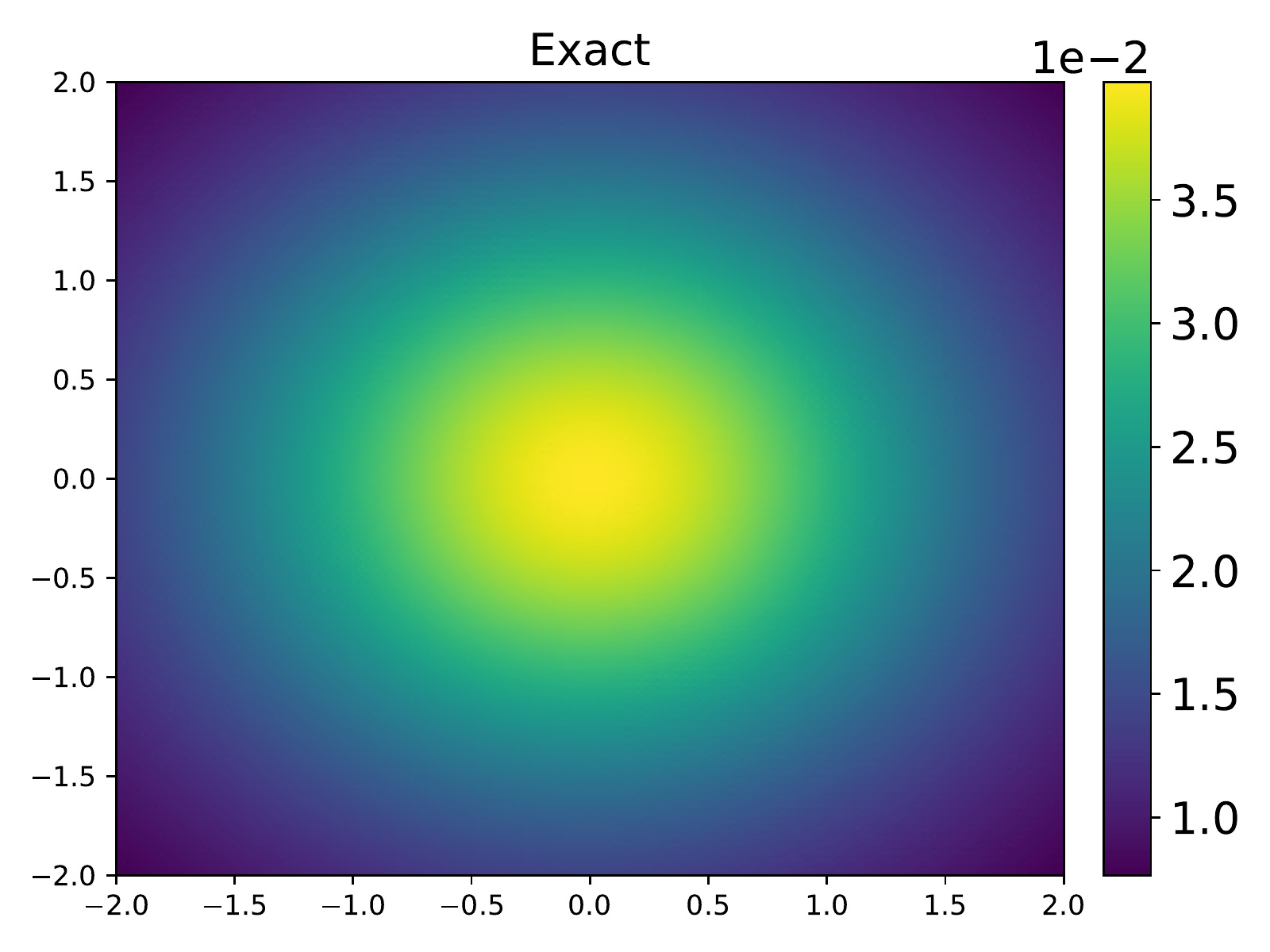}
		\quad
		\includegraphics[height=4cm, width=4.5cm]{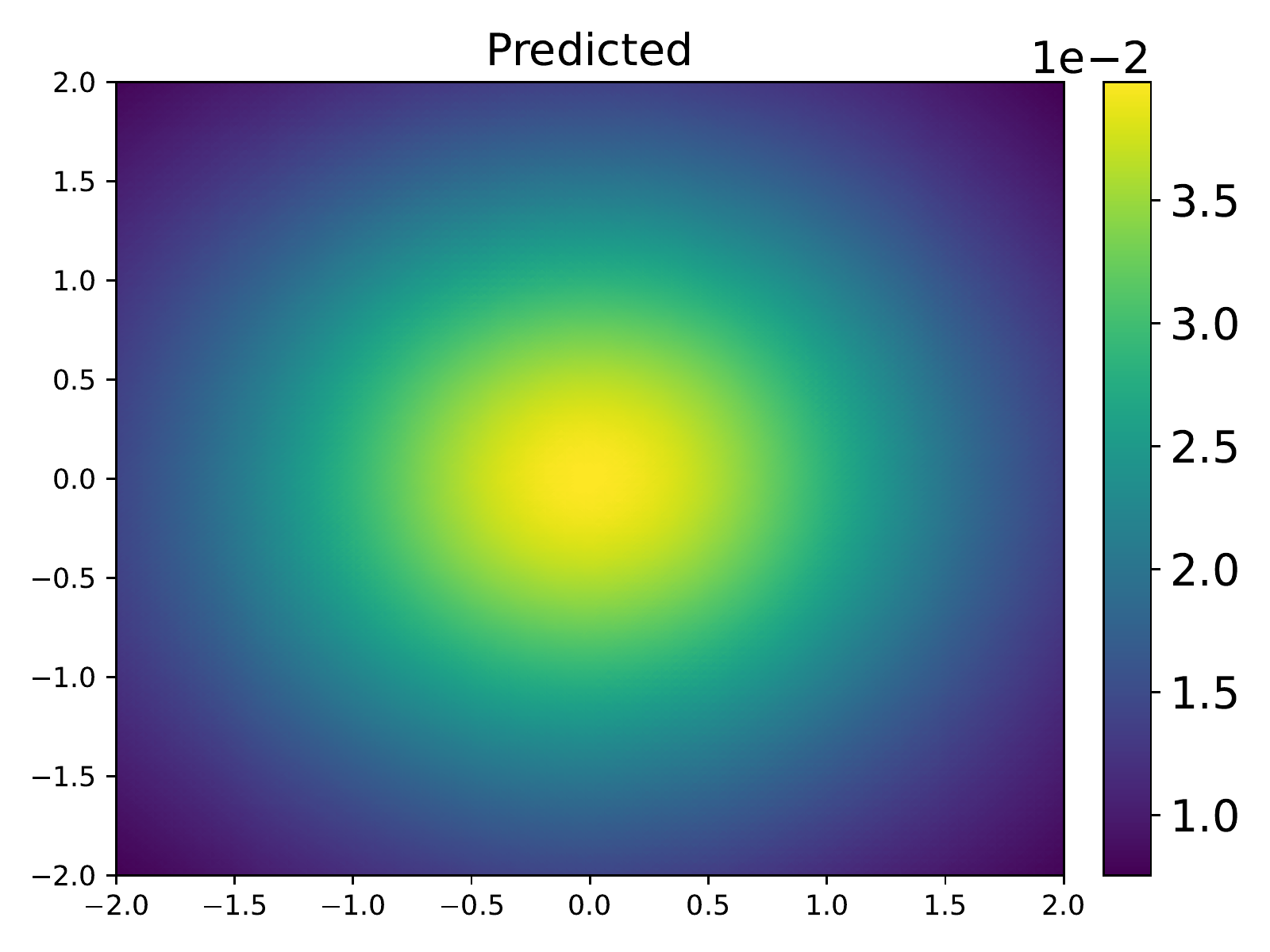}
		\quad
		\includegraphics[height=4cm, width=4.5cm]{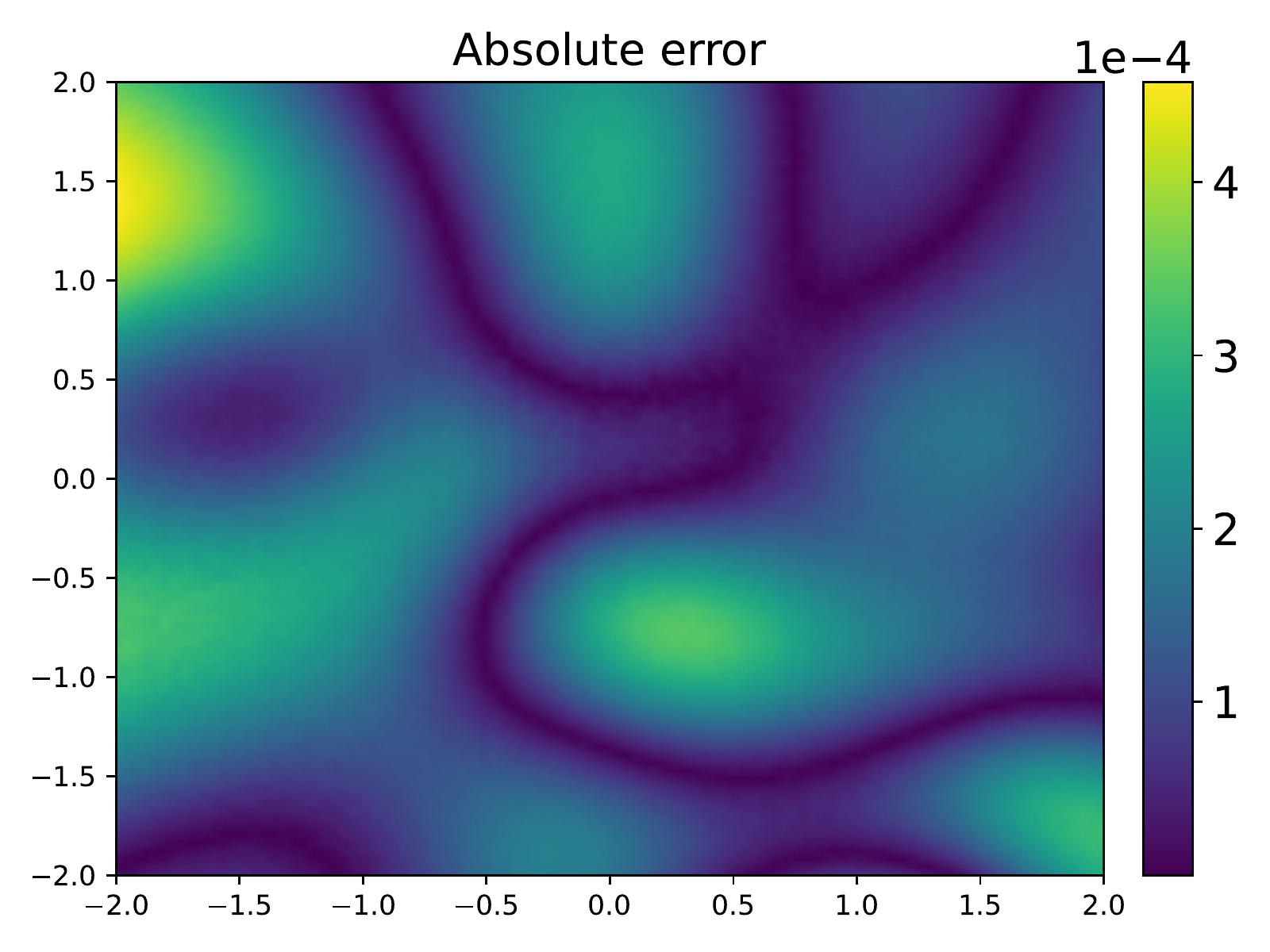}
	}
	\caption{The predicted solutions versus the reference solutions for MCTNF at $t=0, 0.5, 1$. }
	\label{MCTNF_cauchy}
\end{figure}

\begin{figure}[h]
	\centering
	\subfigure{
		\includegraphics[height=5.5cm, width=7cm]{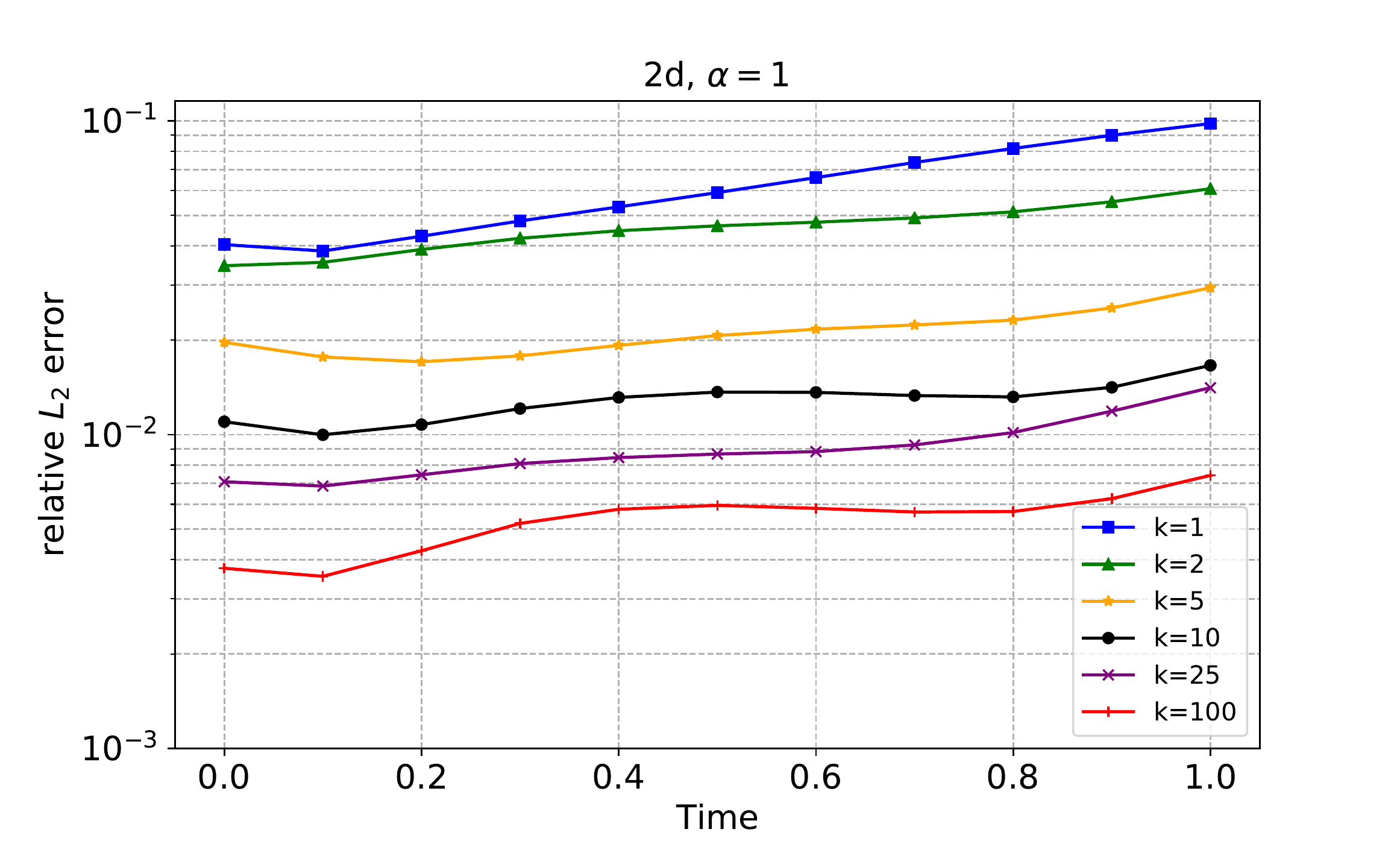}
	    }
	\caption{The relative $L_2$ errors of MCTNF.}
	\label{MCTNF_cauchy_error}
\end{figure}

\begin{figure}[h]
	\centering
	\subfigure{
		\includegraphics[height=3.5cm, width=5.5cm]{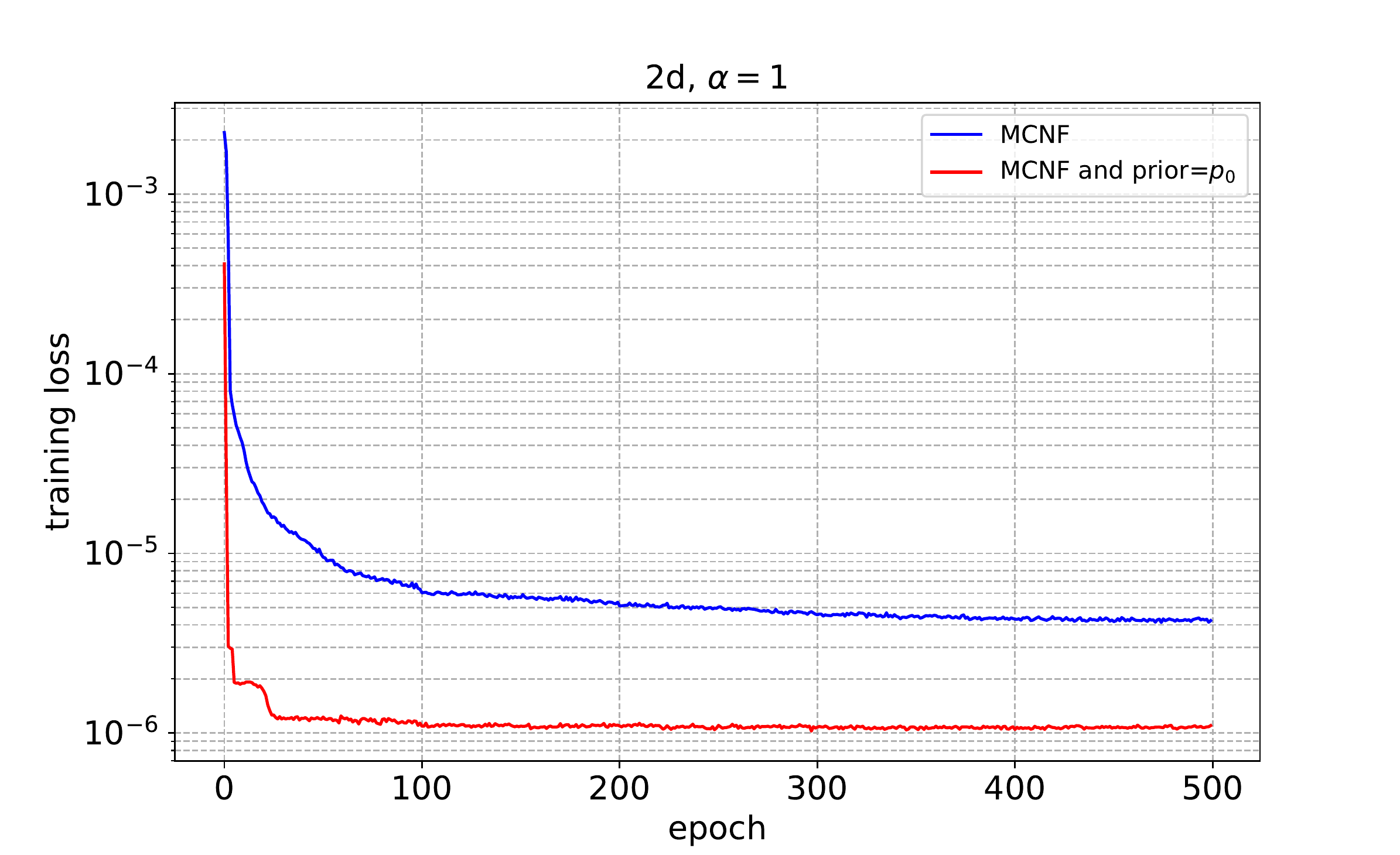}
		\includegraphics[height=3.5cm, width=5.5cm]{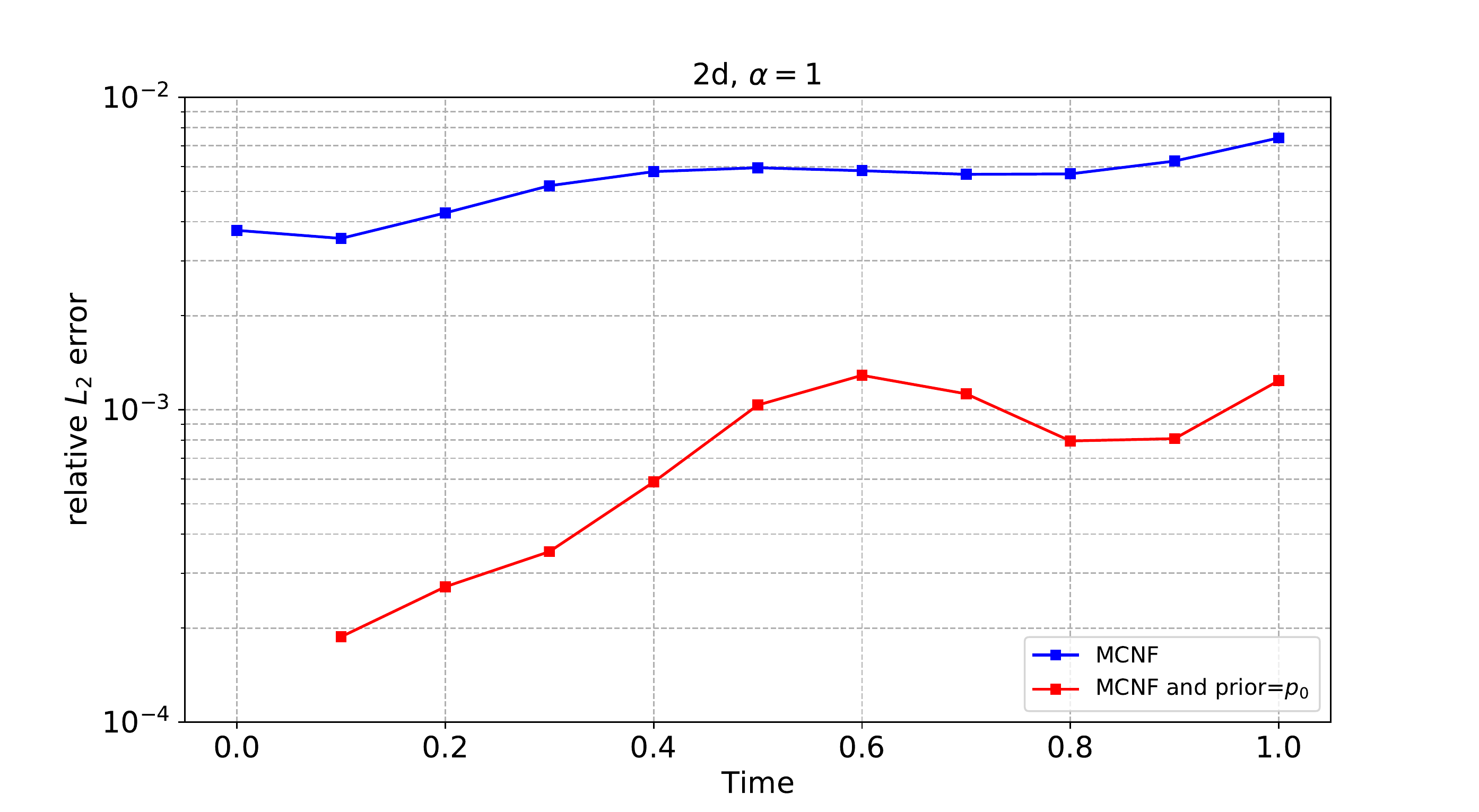}
		\includegraphics[height=3.5cm, width=5.5cm]{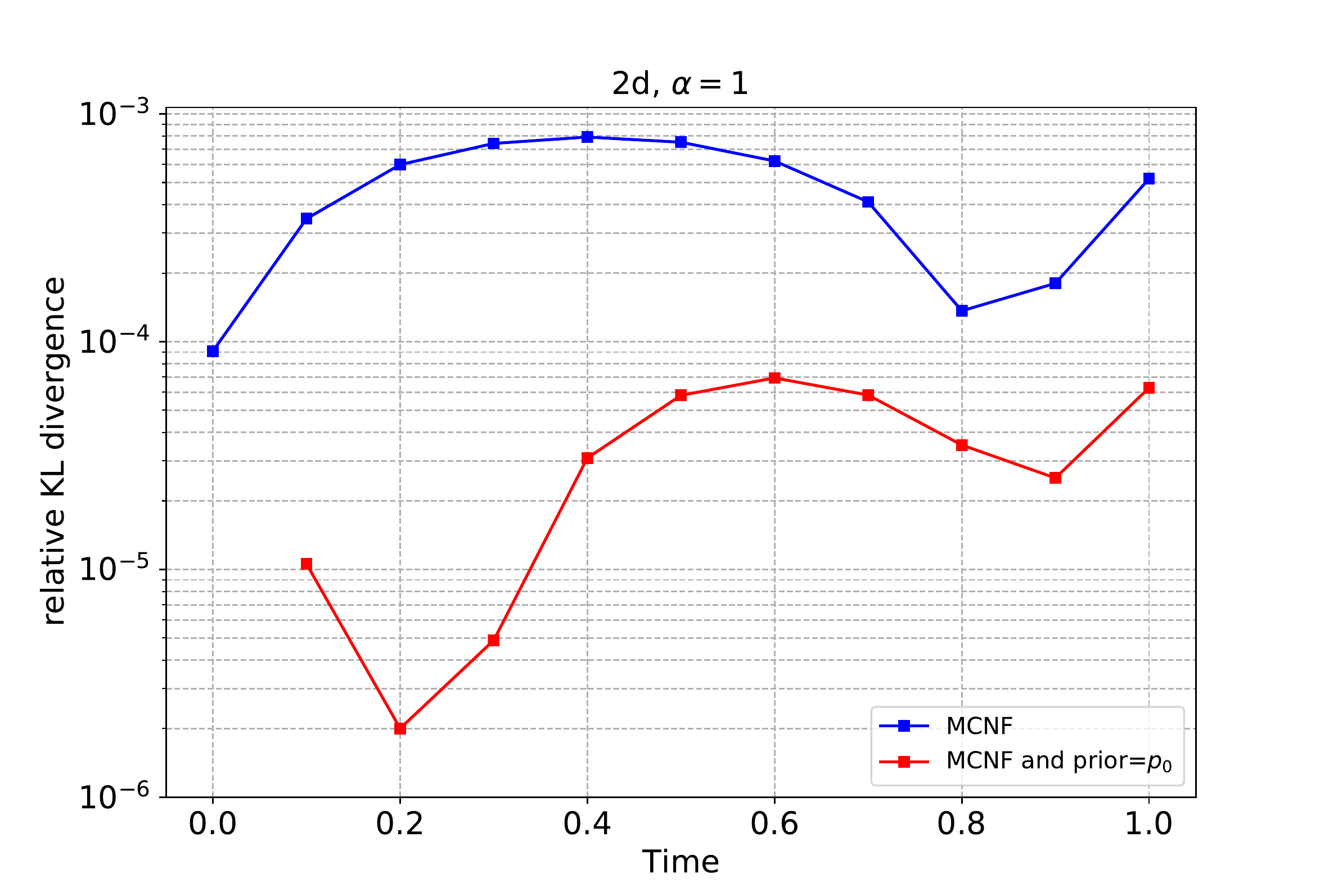}
	}
	\caption{Comparison between the original MCTNF and the modified MCTNF. Left: training loss. Middle: realtive $L_2$ error. Right panel: relative KL divergence.}
	\label{MCTNF_cauchy_com}
\end{figure}

For the NF, we take $L=8$ affine coupling layers with $32$ hidden neurons. The initial spatial samples are drawn from a uniform distribution in $[-3, 3]^2$ and temporal samples are generated from a uniform distribution in $[0,1]$. The sample size is $100000$ and the batch size is set to be $4096$.
For the {MCNF}, the number of samples used to approximate the fractional Laplacian is $100$, $r_0 = 1$, $r_{\epsilon} = 0.01$. $100$ adaptivity iterations with $5$ epochs for each adaptivity iteration are conducted. The initial learning rate is $0.001$ with half decay each $100$ steps. One can observe a good agreement between the predicted solutions and the ground truth from the Fig. \ref{MCTNF_cauchy}.
The relative $L_2$ error and the relative KL divergence against time $t$ for different adaptive iterations are also provided in Fig.\ref{MCTNF_cauchy_error}, which indicates the efficiency of adaptivity. We present the comparison of the relative error between the original MCTNF and modified MCTNF in Fig. \ref{MCTNF_cauchy_com}. The modified MCTNF indeed improve the approximation. It is worth mentioning that the numerical error seems to increase as time evolves. We will explore this issue in the subsequent work.

\section{Conclusions} \label{Conclusion}
We have proposed flow-based adaptive algorithms for solving fractional FPEs. The core idea is to model the unknown PDF by a normalizing flow which yields an explicit PDF model as well as the corresponding exact random samples. 
For stationary FPEs, we proposed two methods: MCNF and GRBFNF. It is usually hard to choose a suitable computational area for unbounded problems. Our methods alleviate this difficulty by adaptively updating the training points. We train the MCNF model or GRBFNF model with current training points, and generate new training points using the current approximate solution. Then the training sets and the solution approximation are updated alternately. 
For time-dependent FPEs, we proposed MCTNF, where we modified the affine coupling layer to satisfy the initial condition exactly to improve the accuracy. 
Our approaches are validated by numerical experiments for both stationary and time-dependent FPEs. 
Compared to non-adaptive methods both MCNF and GRBFNF may improve the accuracy by at least one order of magnitude. From the numerical results, GRBFNF appeals to be more suitable for low-dimensional problems while MCNF demonstrates more flexibility for high-dimensional problems. 
 The main difference between MCNF and GRBFNF is how the fractional Laplacian is approximated. MCNF uses the Monte Carlo approximation while GRBFNF relies on the GRBF approximation of the solution. GRBFNF is more effective for low-dimensional problems since the GRBF approximation is a linear model. MCNF performs better for high-dimensional problems because of the weak dependence of the Monte Carlo method on dimensionality. However, to further reduce the statistical error of the MC  approximation of the fractional Laplacian, we may consider variance reduction techniques, which will be left for future study.

\section*{Acknowledgments}
This work is supported by the National Key R\&D Program of China (2020YFA0712000), the NSF of China (under grant numbers 12288201 and 11731006), and the Strategic Priority Research Program of Chinese Academy of Sciences (Grant No. XDA25010404). The second author is supported by NSF grant DMS-1913163.
\bibliographystyle{plainnat}
\bibliography{references_new}

\end{document}